\documentclass{article}
\usepackage[final]{colm2026_conference}
\usepackage{microtype}
\usepackage{hyperref}
\usepackage{url}
\usepackage{booktabs}
\usepackage{lineno}
\usepackage[utf8]{inputenc}
\usepackage[T1]{fontenc}
\usepackage{amsfonts}
\usepackage{nicefrac}
\usepackage{xcolor}
\usepackage{graphicx}
\usepackage{multirow}
\usepackage{makecell}
\usepackage{amsmath}
\usepackage{amssymb}
\usepackage{empheq}
\usepackage{listings}
\usepackage{subcaption}
\usepackage{tabularx}
\usepackage{array}
\usepackage{threeparttable}
\usepackage{wrapfig}
\usepackage{pifont}
\usepackage[most]{tcolorbox}
\usepackage{enumitem}
\definecolor{promptbg}{rgb}{0.95,0.95,0.95}
\definecolor{promptborder}{rgb}{0.7,0.7,0.7}
\lstdefinestyle{promptstyle}{backgroundcolor=\color{promptbg},frame=single,rulecolor=\color{promptborder},basicstyle=\ttfamily\small,breaklines=true,numbers=none,showspaces=false,showstringspaces=false,tabsize=4}
\newtcolorbox{checklistbox}[2][]{enhanced,arc=3mm,colback=gray!10,colframe=gray!45,boxrule=0.5pt,toprule=0.3pt,colbacktitle=black!75,coltitle=white,fonttitle=\bfseries\normalsize,title={#2},boxsep=0pt,left=4pt,right=4pt,top=4pt,bottom=2pt,lefttitle=4pt,toptitle=4pt,bottomtitle=1pt,titlerule=0pt,#1}
\definecolor{DarkGreen}{RGB}{34,139,34}
\definecolor{DarkRed}{RGB}{200,55,49}
\definecolor{DarkOrange}{RGB}{230,145,56}
\newcommand{\cmark}{\textcolor{DarkGreen}{\ding{51}}}
\newcommand{\xmark}{\textcolor{DarkRed}{\ding{55}}}
\newcommand{\pmark}{\textcolor{DarkOrange}{$\blacktriangle$}}
\graphicspath{{ICLR_2027/}}
\usepackage{tocloft}
\usepackage{titletoc}
\usepackage[title]{appendix}

\title{PsyBer-Agent: A Psychology-Driven User Agent for Recommender Systems}

\author{Chong Zhang$^{1,2}$, Qijie Shen$^{3}$, Haichao Zhang$^{1,2}$, Can Wang$^{1}$, Xiaobo Jin$^{1}$, Jia Wang$^{1}$\thanks{Corresponding author.}\\[0.4em]\textsuperscript{1} Xi'an Jiaotong-Liverpool University,\quad \textsuperscript{2} University of
 Liverpool,\quad \textsuperscript{3} Alibaba Group}
\begin{document}
\ifcolmsubmission
\linenumbers
\fi
\maketitle

\begin{figure}[ht]
    \centering
    \vspace{-10pt}
    \includegraphics[width=0.94\linewidth]{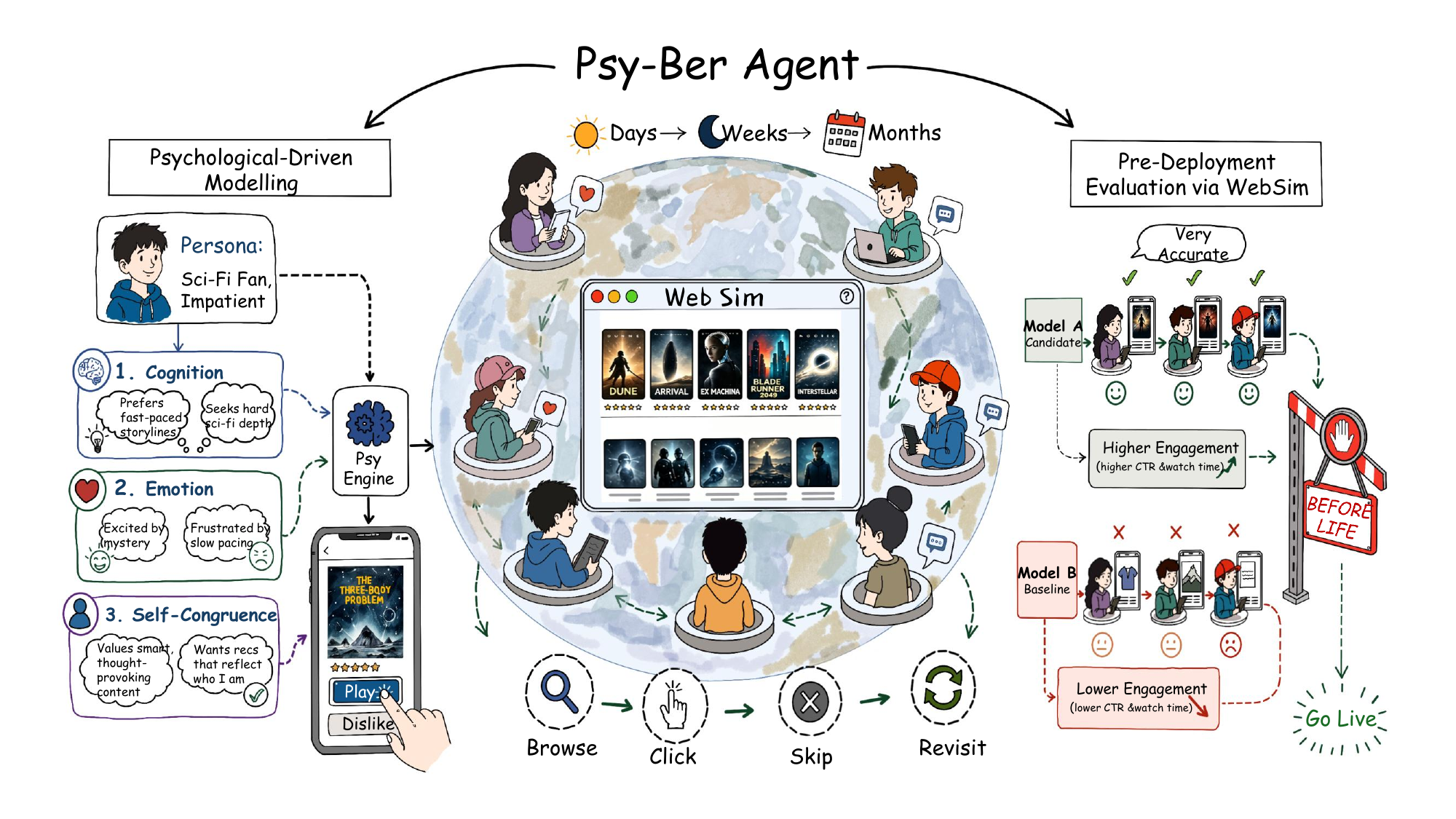}
    \label{fig:Method_workflow}
    
\end{figure}
\begin{abstract}
    Evaluating recommender systems before deployment requires simulators that remain realistic over long, closed-loop interaction cycles. Current LLM-based agents generally function as surface imitators that generate realistic short-term responses but fail to capture the underlying user dynamics such as fatigue, affect, and shifting preferences. We propose \textbf{PsyBer-Agent}, a psychology-inspired simulator that models these dynamics through its \textbf{Psy-Engine}. This engine represents each user with three evolving, model-defined latent states. To calibrate simulated behavior against real interaction logs, we align log-conditioned reference trajectories with simulator-generated latent-state trajectories via Unbalanced Gromov--Wasserstein optimal transport. This approach provides structured supervision for a lightweight policy without the computational burden of fine-tuning the backbone LLM. We also introduce WebSim and the PsyBer Benchmark, which encompass multimodal scenarios and seven recommender backbones. Our framework evaluates behavioral fidelity, latent-state consistency, robustness, and responses to social-cue perturbations. On the tested horizons, PsyBer-Agent shows lower behavioral discrepancy than the prompt-only simulators while maintaining stable performance under large-scale concurrent simulation for up to 5M queued simulation tasks under bounded concurrency. Code is available in the supplementary material and at:\\
    \textbf{PsyBer-Agent:} \url{https://anonymous.4open.science/r/CC66F/}, \\
    \textbf{WebSim Platform:} \url{https://anonymous.4open.science/r/D8EAX/}.
\end{abstract}

\section{Introduction}
\label{sec:intro}

Modern recommender systems rely on rapid iteration, yet collecting realistic user feedback through surveys, user studies, or online A/B tests remains costly, slow, and risky~\cite{grbovic2018real, gomez2015netflix}. Offline log replay is cheaper but misses post-deployment counterfactual interaction dynamics~\cite{ie2019recsim}. Advances in LLMs and agentic systems make pre-deployment user simulation increasingly plausible~\cite{huang2025recommender, gur2023real, zhang2024ai}: LLM-based agents can generate trajectories such as clicks, skips, and continued browsing~\cite{zhang2024generative, wang2025user}, providing a richer testbed for recommender evaluation, A/B testing, and robustness analysis.
\begin{wraptable}{r}{0.68\textwidth}
\centering
\small
\setlength{\tabcolsep}{2.0pt}
\renewcommand{\arraystretch}{1.02}
\caption{Comparison of agent-based user simulators. ``Dyn.'' (dynamic state modeling), ``Psy.'' (psychological modeling), and ``Align.'' (real-log alignment).}
\label{tab:simulated_user_agents}
\small
\begin{tabular}{lccccc}
\toprule
\textbf{Method} & \textbf{Year} & \textbf{Scale} & \textbf{Dyn.} & \textbf{Psy.} & \textbf{Align.} \\
\midrule
RecAgent~\cite{10.1145/3708985} & 2023 & 50 & \cmark & \xmark & \pmark \\
Agent4Rec~\cite{zhang2024generative} & 2024 & 1K & \cmark & \xmark & \pmark \\
UserSim~\cite{zhang2025llm} & 2024 & Data & \pmark & \xmark & \pmark \\
SimUSER~\cite{bougie-watanabe-2025-simuser} & 2025 & 1K & \cmark & \pmark & \pmark \\
RecUserSim~\cite{chen2025recusersim} & 2025 & 3K & \cmark & \pmark & \pmark \\
PUB~\cite{ma2025pub} & 2025 & Data & \pmark & \pmark & \cmark \\
DyTA4Rec~\cite{10.1145/3746252.3760878} & 2025 & 6K & \cmark & \xmark & \cmark \\
RecInter~\cite{jin-etal-2025-beyond} & 2025 & 1K & \cmark & \xmark & \cmark \\
\midrule
\textbf{PsyBer-Agent} & \textbf{2026} & \textbf{5M} & \cmark & \cmark & \cmark \\
\bottomrule
\end{tabular}
\parbox{0.50\textwidth}{\footnotesize\raggedright
\cmark: explicit; \pmark: partial/implicit; \xmark: no explicit.
}
\vspace{-10pt}

\end{wraptable}

However, current LLM-based user simulators remain unreliable as pre-deployment evaluation proxies~\cite{seshadri2026lost, chen2026towards}. Existing methods improve local realism through personas, memory, feedback distillation, or latent reasoning, but many remain primarily prompt-driven: they generate plausible local actions without an explicit, controllable state-transition mechanism connecting history to subsequent behavior. As a result, local errors may accumulate into behavioral drift, making these simulators brittle across environments and long-horizon interactions~\cite{chaney2018algorithmic, nakash2025breaking}. More specifically, existing systems generally do not expose a unified computational mechanism for representing how curiosity-, fatigue-, social satisfaction-, and self-congruence-related signals change under repeated exposure~\cite{knijnenburg2012explaining}. This limits their ability to represent cumulative preference dynamics in a controlled and interpretable state space. Practical recommender evaluation also requires scalable and reproducible multi-user simulation for model comparison, A/B discrimination, and robustness analysis, while existing systems remain mostly restricted to single-agent or small-scale settings~\cite{gilotte2018offline}. \autoref{tab:simulated_user_agents} summarizes these differences.

\begin{figure}[t]
    \centering
    
    \includegraphics[width=1.0 \linewidth]{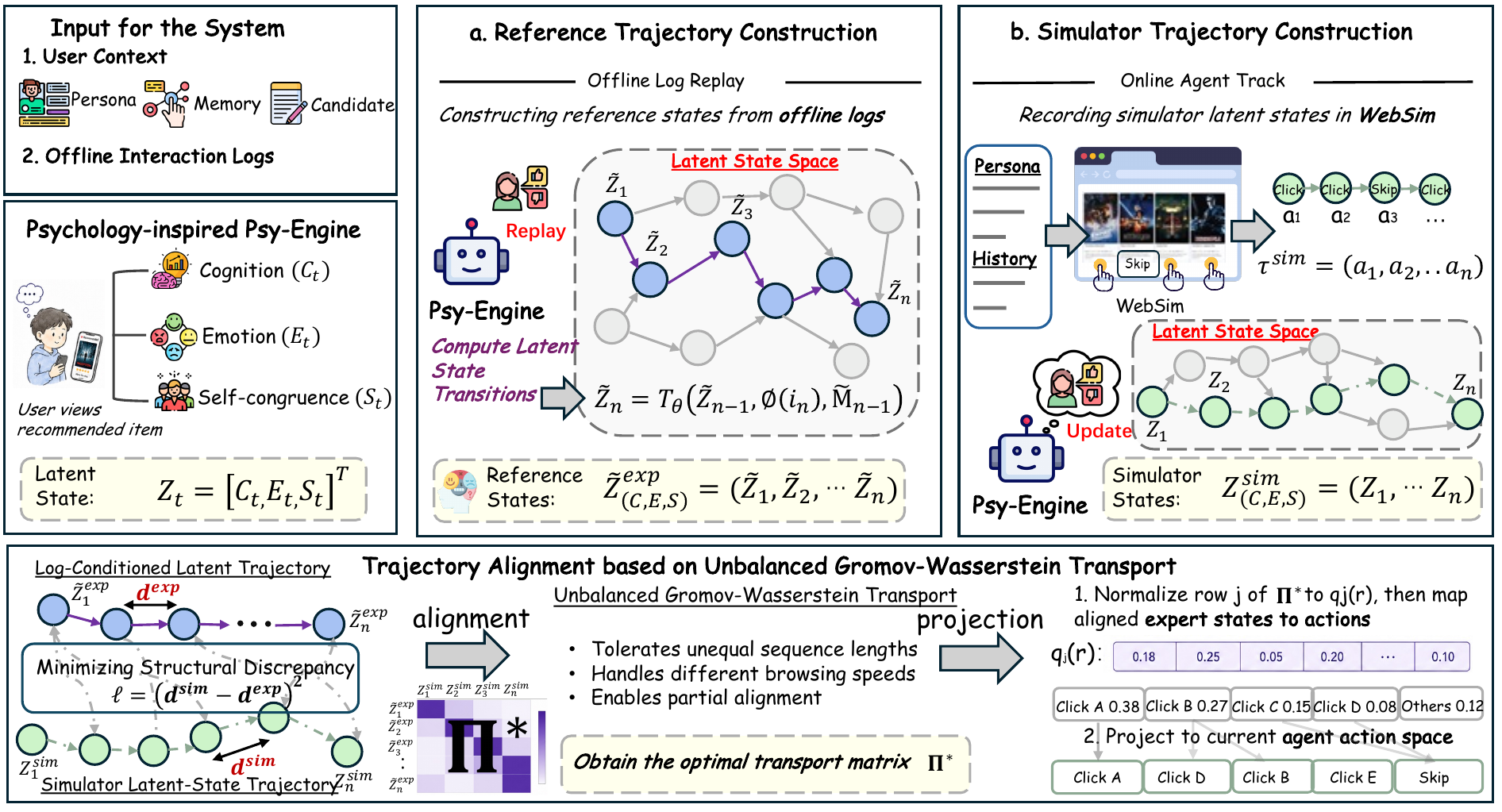}
    
    \caption{The workflow of PsyBer-Agent. PsyBer-Agent is mainly composed of~\textbf{Psy-Engine} and~\textbf{Psycho-behavioral Alignment}.}
    \label{fig:method_main}
    
\end{figure}

To address these limitations, we propose \textbf{PsyBer-Agent}, a psychology-inspired user simulator with an explicit model-defined latent-state engine. PsyBer-Agent builds on \texttt{Psy-Engine} to model three evolving user-state dimensions: cognition, emotional/social satisfaction, and self-congruence. These model-defined states interact with personas, memory, multimodal item representations, so behavior is generated through explicit state transitions rather than prompt-only role-playing. We further introduce a psycho-behavioral alignment framework based on Unbalanced Gromov-Wasserstein (UGW) optimal transport~\cite{sejourne2021unbalanced}, which converts offline expert trajectories into structured supervision for a lightweight online policy without fine-tuning the backbone LLM. The main workflow of PsyBer-Agent can be seen in~\autoref{fig:method_main}.

Built on PsyBer-Agent, we introduce~\texttt{PsyBer}  Benchmark and the~\texttt{WebSim} Platform for simulated-user A/B testing and recommender model evaluation. The benchmark evaluates behavior fidelity, latent-state tracking, decision robustness, and efficiency under controlled closed-loop rollouts. Across WebSim evaluations, PsyBer-Agent more faithfully matches real interaction logs, better matches model-defined latent-state trajectories, and remains stable under perturbations and stress. By distilling expensive backbone reasoning into a lightweight online policy, it further supports industrial-scale concurrent simulation at up to~5M queued simulation tasks under bounded concurrency, with bounded latency growth, high success rates, and explicit cost accounting. Together, these results establish PsyBer-Agent as a practical and scalable proxy for pre-deployment recommender evaluation. Our contributions are:
\begin{itemize}[leftmargin=1.5em, itemsep=0pt, topsep=0pt, parsep=0pt, partopsep=0pt]
    \item We propose \textbf{PsyBer-Agent}, a psychology-inspired user simulator with explicit, model-defined latent-state dynamics for controllable long-horizon recommender simulation.
    \item We introduce a psycho-behavioral alignment framework that uses \textbf{Psy-Engine} and UGW alignment to distill observed offline behavior into efficient online policies without backbone fine-tuning.
    \item We build \textbf{PsyBer Benchmark} and \textbf{WebSim}, a unified testbed for closed-loop A/B testing, behavior-fidelity evaluation, robustness analysis, and million-scale simulation.
\end{itemize}

\section{Problem Formulation and Notation}
\label{sec:problem}

Given implicit interaction logs $\mathcal{D}=\{(u,i,t)\}$, where $u$ denotes a user, $i$ denotes an interacted item, and $t$ denotes the interaction timestamp. We reconstruct each user’s expert sequence $\tau_u^{\mathrm{exp}}=(i_{u,1},\ldots,i_{u,T_u})$. At step $t$, the simulator observes a top-$K$ slate $\mathcal{L}_{u,t}=\{\ell_{u,t}^{(1)},\ldots,\ell_{u,t}^{(K)}\}$ and selects an action $a_{u,t}\in\{1,\ldots,K,\texttt{skip}\}$ from the available action set.

Each item is represented by $\phi(i)=(\mathbf{v}_i,n_i,c_i,p_i)$, where $\mathbf{v}_i$ is the semantic embedding, and $n_i,c_i,p_i$ denote identity gap, cognitive load, and social proof. Cross-domain feature construction is detailed in Appendix~\ref{App:Psy-Engine_init}. The users' psychological state is
$\mathbf{z}_{u,t}=[C_{u,t},E_{u,t},S_{u,t}]^\top\in[0,1]^3$, with persona $\psi_u$ and memory $\mathcal{M}_{u,t}$, and evolves as
\begin{equation}
    \mathbf{z}_{u,t+1} = \mathcal{T}_\Theta(\mathbf{z}_{u,t},\mathcal{L}_{u,t},a_{u,t},\mathcal{M}_{u,t},\psi_u).
\end{equation}
To align $\mathbf{z}_{u,t}$with $\tau_u^{\mathrm{exp}}$, we train the fast policy $\pi_\theta$, we align simulated and expert trajectories via UGW, and obtain soft action targets $\hat{\mathbf{p}}_{u,t}$:
\begin{equation}
\mathcal{L}_{\mathrm{policy}} = \sum_u\sum_t \mathrm{CE}\!\left(
\hat{\mathbf{p}}_{u,t},
\pi_\theta(\cdot\mid\mathbf{z}_{u,t},\mathcal{L}_{u,t})
\right).
\end{equation}

\begin{checklistbox}{Problem Statement}
    Given implicit logs $\mathcal{D}$, can aligning observed user behaviors with psychology-inspired, model-defined state transitions yield a reliable behavioral simulator for recommendation systems?
\end{checklistbox}

\section{PsyBer-Agent Framework}
\label{sec:PsyBer_method}

\textbf{Overview.} PsyBer-Agent couples psychology-inspired latent-state modeling with psycho-behavioral alignment: \emph{Psy-Engine} defines user state, transition dynamics, and action evaluation, while \emph{PsyBer-Agent} constructs multimodal item features, reconstructs log-conditioned latent reference trajectories, and learns a lightweight, fast policy via UGW-based alignment. 

The C/E/S states are computational constructs motivated by psychological principles, so CES-MSE measures internal consistency within this model-defined state space rather than validated correspondence to independently measured human psychological states. PsyBer-Agent is implemented based on Camel-AI~\cite{li2023camel}.

\subsection{Simulation Experiment Platform Construction}
\label{sec:websim}

We build \textbf{WebSim}, an end-to-end platform for multimodal asset preparation, offline alignment/policy training, and browser-level interactive simulation (details in Appendix~\ref{App:websim}). For each dataset, WebSim creates a multimodal semantic view and a psycho-behavioral view for Psy-Engine: each item $i$ is encoded by $\mathbf{v}_i$ with features $(n_i,c_i,p_i)$, and each user is initialized with persona anchors $(\mathbf{u}_{\text{base}}, \mathbf{u}_{\text{ideal}}, \mathbf{u}_{\text{real}})$. During simulation, a web service and browser controller execute click/pagination actions, use the trained fast policy when available or a constrained vision-language fallback otherwise, and log traces, psychological states, and run summaries for reproducible evaluation.

\paragraph{Persona Anchor Construction.}
For each user $u$, WebSim constructs three pre-window persona anchors: $\mathbf{u}_{\mathrm{base}}$ encodes available profile and metadata, $\mathbf{u}_{\mathrm{ideal}}$ summarizes long-term preferences, and $\mathbf{u}_{\mathrm{real}}$ summarizes the most recent $K$ temporally ordered interactions. Item representations are pooled with a masked mean and projected into a shared persona space; missing fields and empty histories use fixed cold-start vectors with an availability mask. These anchors condition the simulator and are not interpreted as direct measurements of psychological traits. Information from the simulation window and later periods is excluded.

\subsection{Psy-Engine Modeling}
\label{sec:psy-engine}

We design Psy-Engine as a psychology-inspired latent-state engine for history-dependent user behavior. At each step $t$, the simulator maintains $\mathbf{z}_t=[C_t,E_t,S_t]^\top\in[0,1]^3$, with coordinates motivated by cognitive reserve $C_t$, emotional satisfaction $E_t$, and self-congruence $S_t$, respectively. These variables serve as model-defined latent states for behavior generation, with their psychological interpretations providing modeling priors.

\paragraph{State Transition Model.} Given the consumed item $i_t$ with features $\phi(i_t)=(\mathbf{v}_t,n_t,c_t,p_t)$ and memory bank $\mathcal{M}_t$, here, $n_t$, $c_t$, and $p_t$ represent incongruence, a cognitive-load proxy, and social proof. Psy-Engine computes novelty $\nu_t$, semantic saturation $\sigma_t$, semantic fit $\beta_t$, serendipitous interest $\rho_t$, and self-conflict $\gamma_t$. Detailed definitions are provided in Appendix~\ref{App:Psy-Engine}. We use $\operatorname{Proj}_{[0,1]}(x)=\min(1,\max(0,x))$ to keep the state coordinates bounded.

Psy-Engine organizes these signals into an attractiveness proxy $A_t$, a fatigue proxy $F_t$, and two interaction-dependent targets, emotional satisfaction $E_t$ and
self-congruence $S_t$, here, $E_t^\star$ and $S_t^\star$ denote target states computed from the current item.:
\begin{equation}
\begin{aligned}
    A_t &= f_A(\nu_t,n_t), &
    F_t &= f_F(c_t,\sigma_t), &
    E_t^\star &= f_E(\beta_t,\rho_t,p_t), &
    S_t^\star &= f_S(\beta_t,\gamma_t).
\end{aligned}
\end{equation}
The inputs to these functions are motivated by exploratory interest \cite{loewenstein1994psychology,silvia2008interest}, cognitive load and
satiation \cite{sweller1988cognitive,10.1086/521898}, affective satisfaction
and social influence \cite{KOTKOV2016180,cialdini2004social}, and
self-congruence \cite{higgins1987self,sirgy1982self}. We instantiate these
functions as fixed, bounded aggregation rules; their exact definitions are
provided in Appendix~\ref{App:Psy-Engine}. Their specific forms are
computational choices used to operationalize the corresponding modeling
motivations, rather than uniquely determined psychological laws.

The resulting state is updated through a bounded relaxation operator:
\begin{equation}
\mathbf{z}_{t+1} = \operatorname{Proj}_{[0,1]^3} \left( \mathbf{z}_t+\alpha\mathbf{d}_t \right),
\quad
\mathbf{d}_t=
\begin{bmatrix}
A_t-F_t\\
E_t^\star-E_t\\
S_t^\star-S_t
\end{bmatrix},
\label{eq:psy_state_update}
\end{equation}
where $\mathbf{z}_t=[C_t,E_t,S_t]^\top$,
$\operatorname{Proj}_{[0,1]^3}$ keeps each coordinate in $[0,1]$, and $\alpha\in[0,1]$ controls the relaxation rate. Thus, the cognitive coordinate responds to the net attractiveness--fatigue signal, whereas the affective and self-congruence coordinates move toward interaction-dependent targets. We use a shared update rate to keep the parameterization compact, rather than to assume identical psychological time constants. We set $\alpha=0.35$ following the sensitivity analysis in Appendix~\ref{App:Exp_Sens}. This bounded transition is a parsimonious computational hypothesis for history-dependent behavior; its coordinate-wise expansion is provided in Appendix~\ref{App:cross_domain}. For subsequent trajectory construction, we denote the complete Psy-Engine transition by
\begin{equation}
\mathbf{z}_{t+1} = \mathcal{T}_{\Theta}\left(\mathbf{z}_t, \phi(i_t), \mathcal{M}_t \right),
\label{eq:psy_transition_operator}
\end{equation}
where $\mathcal{T}_{\Theta}$ summarizes the bounded state update
defined above and $\Theta$ denotes the fixed Psy-Engine parameters.
Implementation details are provided in Appendix~\ref{App:Align_con}.

\subsection{Psycho-behavioral Alignment}
\label{sec:psycho_behavioral_alignment}

Offline logs lack annotations of users' latent decision states. PsyBer-Agent replays logged actions in Psy-Engine to construct model-defined latent-state trajectories and derive state-conditioned soft supervision, enabling policy imitation under comparable computational states without treating them as direct psychological measurements. Details are provided in Appendix~\ref{App:matrix_init}.

\paragraph{Model-Defined State Sequence Construction.}
Offline browsing logs record clicks but do not provide annotations of the latent states underlying those decisions. We therefore use the logged clicks as forced inputs to the deterministic Psy-Engine transition and reconstruct a model-defined reference trajectory for subsequent alignment.
Let $\tau^{\mathrm{exp}}=(i_1,\ldots,i_n)$ denote an expert click sequence, where $i_r$ is the $r$-th clicked item. For comparison, let $\mathbf{Z}^{\mathrm{sim}}=(\mathbf{z}^{\mathrm{sim}}_1,\ldots,\mathbf{z}^{\mathrm{sim}}_m)$ denote the simulator-generated trajectory, with $\mathbf{z}^{\mathrm{sim}}_j\in[0,1]^3$ encoding cognitive reserve ($C$), emotional satisfaction ($E$), and self-congruence ($S$). Starting from the persona-conditioned initial state $\tilde{\mathbf{z}}_0$ and replay memory $\tilde{\mathcal{M}}_0$, we apply teacher forcing by treating each logged expert item as the consumed item and recursively obtain the model-defined pseudo-expert trajectory $\tilde{\mathbf{Z}}^{\mathrm{exp}}=(\tilde{\mathbf{z}}_1,\ldots,\tilde{\mathbf{z}}_n)$:
\begin{equation}
\tilde{\mathbf{z}}_r
=
\mathcal{T}_{\Theta}\!\left(
\tilde{\mathbf{z}}_{r-1},
\phi(i_r),
\tilde{\mathcal{M}}_{r-1}
\right),
\quad r=1,\ldots,n.
\end{equation}
The resulting state $\tilde{\mathbf{z}}_r$ serves as the model-defined reference state at the $r$-th expert interaction for trajectory alignment and state-conditioned supervision. Here, $\mathcal{T}_{\Theta}$ denotes the item-level Psy-Engine transition defined above, $\phi(i_r)=(\mathbf{v}_r,n_r,c_r,p_r)$ is the psycho-behavioral feature tuple of item $i_r$, and $\tilde{\mathcal{M}}_{r-1}$ stores the normalized embeddings of previously replayed expert items. Detailed construction procedures are provided in Appendix~\ref{App:Align_con}.

\paragraph{Prior-Guided Latent-State Trajectory Alignment.}
We align the model-defined expert trajectory and the simulator trajectory using an Unbalanced Gromov--Wasserstein (UGW) transport plan~\cite{sejourne2021unbalanced}. Because the expert trajectory contains clicks but no latent-state annotations, and may differ from the simulator trajectory in length and browsing rhythm, we first construct a soft state prior from expert prefixes:
\begin{equation}
\hat{\mathbf{z}}_r = g_{\varphi}\!\left(
\psi,\tau^{\mathrm{exp}}_{1:r},\phi(i_r),\frac{r}{n}
\right),
\end{equation}
where $\psi$ denotes the user persona and $\tau^{\mathrm{exp}}_{1:r}=(i_1,\ldots,i_r)$ is the expert prefix. The regressor $g_{\varphi}$ maps each prefix into the simulator state space, detailed descriptions are provided in Appendix~\ref{App:matrix_init}. For simulator step $j$ and expert position $r$, we define an initialization cost $c^{\mathrm{init}}_{j,r}$ that combines state and local-dynamics discrepancies:
\begin{equation}
c^{\mathrm{init}}_{j,r} = \lambda_z
\left\| \mathbf{z}^{\mathrm{sim}}_j-\hat{\mathbf{z}}_r \right\|_2^2
+
\lambda_{\Delta} \left\| \Delta\mathbf{z}^{\mathrm{sim}}_j-\Delta\hat{\mathbf{z}}_r \right\|_2^2 ,
\end{equation}
where $\mathbf{z}^{\mathrm{sim}}_j$ denotes the model-defined latent state generated by the simulator at step $j$, while $\hat{\mathbf{z}}_r$ denotes the latent state predicted by the regressor from the expert trajectory prefix up to position $r$, $\Delta\mathbf{z}^{\mathrm{sim}}_j=\mathbf{z}^{\mathrm{sim}}_j-\mathbf{z}^{\mathrm{sim}}_{j-1}$ and $\Delta\hat{\mathbf{z}}_r=\hat{\mathbf{z}}_r-\hat{\mathbf{z}}_{r-1}$, with $\Delta\mathbf{z}^{\mathrm{sim}}_1=\Delta\hat{\mathbf{z}}_1=\mathbf{0}$. The nonnegative coefficients $\lambda_z$ and $\lambda_{\Delta}$ control the contributions of the two discrepancies; further details are provided in Appendix~\ref{App:matrix_init}

Let $\Pi=(\Pi_{j,r})$ denote the transport plan, where $\Pi_{j,r}$ is the soft matching weight between simulator step $j$ and expert position $r$. We initialize $\Pi$ from the cost $c^{\mathrm{init}}_{j,r}$ as
\begin{equation}
\Pi^{(0)}_{j,r}
\propto
\exp\!\left( -\frac{c^{\mathrm{init}}_{j,r}}{\tau_0} \right),
\qquad
\Pi^\star = \arg\min_{\Pi\ge 0} \mathcal{L}_{\mathrm{UGW}}(\Pi),
\end{equation}
where $\tau_0>0$ controls the sharpness of the soft matching. Starting from $\Pi^{(0)}$, UGW minimizes $\mathcal{L}_{\mathrm{UGW}}(\Pi)$ over nonnegative transport plans to obtain the optimized plan $\Pi^\star$. The detailed UGW alignment process can be found in Appendix~\ref{App:matrix_init}.

\paragraph{Alignment-guided Fast Policy Learning and Inference.}
Because trajectory alignment may leave some simulator steps without direct expert actions, we distill $\Pi^\star$ into soft action targets for a lightweight fast policy used as a fallback during online simulation.
The optimized transport plan $\Pi^\star$ is row-normalized into step-wise soft supervision, where $q_j(r)$ denotes the alignment weight between simulator step $j$ and expert position $r$. The aligned expert-item weights are projected onto the executable action set $\mathcal{A}_j=\mathcal{L}_j\cup\{\texttt{skip}\}$, where matched items contribute to their corresponding slate actions and residual unmatched mass is assigned to $\texttt{skip}$, yielding the soft action label $\hat{p}_j$ over $\mathcal{A}_j$. The fast policy is trained by
\begin{equation}
q_j(r)
=
\frac{\Pi^\star_{j,r}}
{\sum_{r'=1}^{n}\Pi^\star_{j,r'}+\delta},
\ \delta>0,
\quad
\mathcal{L}_{\mathrm{policy}}
=
\sum_j
\mathrm{CE}\!\left(
\hat{p}_j,
\pi_\theta(\cdot\mid\mathbf{z}^{\mathrm{sim}}_j,\mathcal{A}_j)
\right).
\end{equation}

At inference, the fast policy selects an action from the same executable action set:
\begin{equation}
a_j^{\mathrm{fast}}
=
\arg\max_{a\in\mathcal{A}_j}
\pi_\theta
\left(
a\mid\mathbf{z}^{\mathrm{sim}}_j,\mathcal{A}_j
\right),
\quad
\mathcal{A}_j=\mathcal{L}_j\cup\{\texttt{skip}\}.
\end{equation}
The selected action is used directly or as a top-$k$ hint for PsyBer-Agent, with the backbone LLM invoked only when needed. The fast policy uses latent-state, persona, memory, and candidate-item features. It falls back to the LLM/VLM when its confidence is insufficient or the action space is unsupported, and uses random selection only if that route also fails. This enables efficient online inference from aligned psycho-behavioral supervision.

\section{PsyBer Benchmark}
\label{sec:benchmark}

\subsection{Benchmark Coverage}

\begin{figure}[!ht]
\centering
\includegraphics[width=\linewidth]{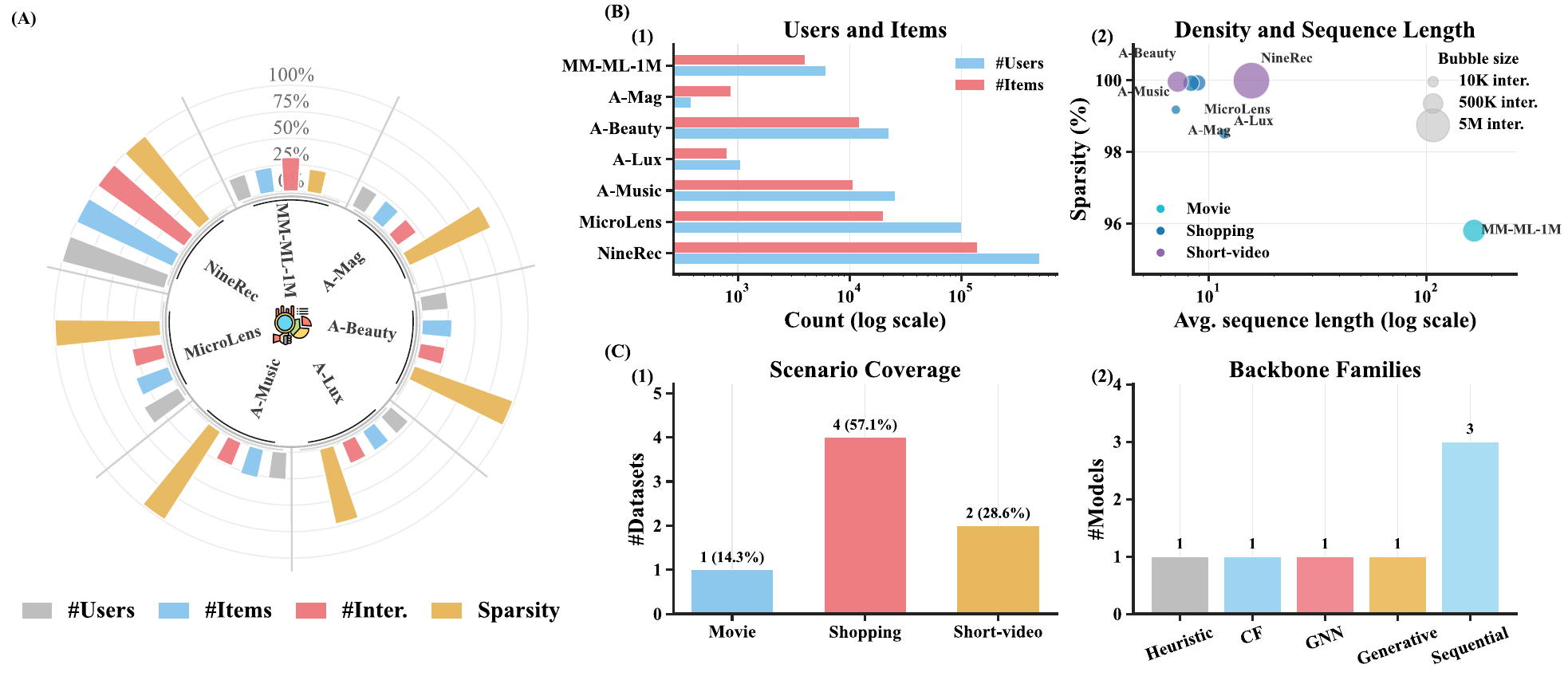}
\caption{Data Statistics for the PsyBer Benchmark. \textbf{(A.)} Dataset footprint; \textbf{(B.)} scale and density; \textbf{(C.)} scenario and backbone coverage.}
\label{fig:data_stat}
\end{figure}

The PsyBer Benchmark consists of two core tasks: \textbf{(1.)} Multi-scenario user online simulation. \textbf{(2.)} Evaluate the performance of the recommendation model. The ~\autoref{fig:data_stat} shows detailed data statistics. Detailed descriptions of the task definitions are provided in Appendix~\ref{sec:task_def}.

\subsection{Task Evaluation}

\textbf{WebSim Experiment Platform.} We introduce \textbf{WebSim}, a Flask-based closed-loop simulator that decouples user interaction from recommender inference and uses random fallback for unavailable or empty model outputs, enabling consistent cross-model and cross-dataset evaluation. Additional implementation details are provided in Appendix~\ref{App:websim}.

\textbf{Recommender Models and Datasets.} To support recommendation simulation across diverse scenarios, WebSim is built on a set of representative recommendation backbones, including PopRec~\cite{cremonesi2010performance}, BPR-MF~\cite{rendle2012bpr}, LightGCN~\cite{he2020lightgcn}, Multi-VAE~\cite{liang2018variational}, GRU4Rec~\cite{hidasi2015session}, SASRec~\cite{kang2018self}, and BERT4Rec~\cite{sun2019bert4rec}. To construct PsyBer-Test, a multi-scene and multi-modal benchmark, we integrate recommendation datasets from several representative application scenarios, including MM-ML-1M~\cite{zhang2026exploring} for movie viewing, the Amazon Review ~\cite{hou2024bridging} for shopping, and MicroLens~\cite{ni2025content} and NineRec~\cite{zhang2025ninerec} for image-text information flow and short-video browsing scenarios. Detailed data statistics are provided in Appendix~\ref{App:data} and~\ref{App:Ben_cov_eva}.

\textbf{Evaluation Scope.} We evaluate PsyBer-Agent from four aspects: Online A/B test evaluation, behavior fidelity, baseline simulator comparison, and scalability and cost. Detailed metrics design is provided in the Appendix~\ref{App:metrics_eval}.

\section{Experiments}
\label{sec:exp}

\subsection{Experimental Settings}

\textbf{Backbone Recommender Models.} We evaluate the proposed PsyBer-Agent on representative baseline models from the PsyBer Benchmark. Detailed descriptions and core mechanisms of these models are provided in Appendix~\ref{App:tested_recsys}.

\textbf{User Simulator Baselines.} We compare representative agent-based simulators covering social, conversational, RL-based, and platform-oriented recommendation settings. Details and the protocol-compatibility and reproducibility criteria for quantitative inclusion are provided in
Appendix~\ref{App:sim_baseline}.

\textbf{Dataset \& Hardware Setting.} We split each user's interactions chronologically into 70/10/20 train/validation/test sets and evaluate on held-out users. Hardware details are reported in Appendix~\ref{App:exp_hard}.

\subsection{Evaluation and Analysis}

We evaluate PsyBer-Agent through three main experiments and three diagnostic analyses. The main experiments examine: (\textbf{I.}) behavior fidelity at both population and individual levels, (\textbf{II.}) baseline simulator comparison, and (\textbf{III.}) scalability and token cost under high-concurrency simulation. The diagnostic analyses include (\textbf{IV.}) component ablation, (\textbf{V.}) information-channel ablation, and (\textbf{VI.}) an offline backbone sanity check. Additional analyses of group behavior, parameter sensitivity, robustness, unit cost, and fallback usage are reported in Appendices~\ref{App:Group_be}, \ref{App:Exp_Sens}, \ref{App:Exp_Rob}, and \ref{App:exp_hard}.
\begin{figure}[!ht]
\centering
\includegraphics[width=0.9\linewidth]{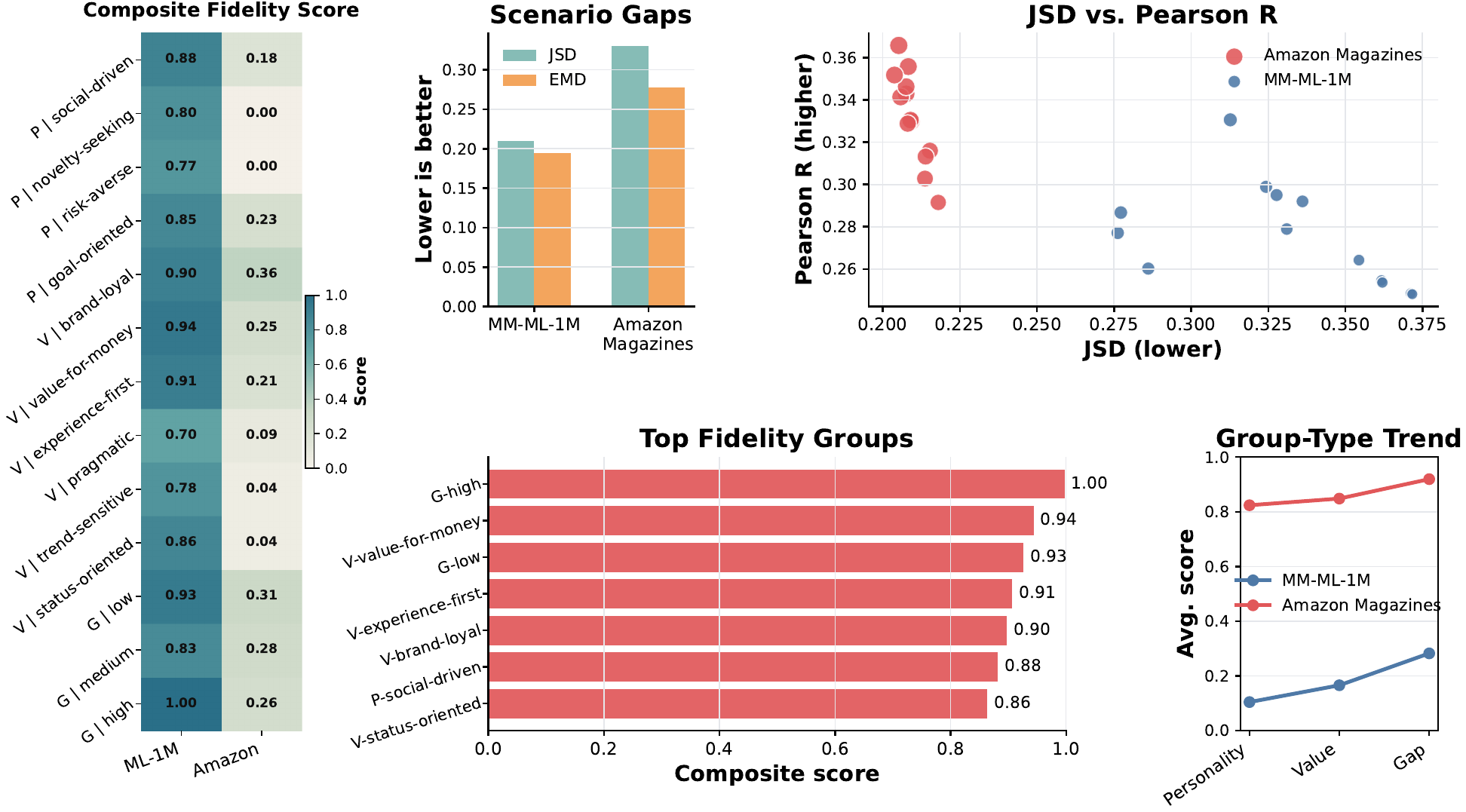}
\caption{PsyBer-Agent simulation experiments and profile-conditioned population analysis.}
\label{fig:group_analysis}
\end{figure}

\textbf{PsyBer-Agent Online A/B Test Evaluation.}
We compare 500 human users with 500 simulated users under the same recommendation interface and fixed backend, with the decision-maker as the only varying factor. HR\@10, NDCG\@10, and CTR are computed using the same protocol, held-out relevance targets, and replayed candidate slates for both groups. The full PsyBer-Agent with Qwen3-VL-235B closely matches human behavior, with absolute metric differences of 0.0030, 0.0016, and 0.0040, respectively. Removing memory increases these differences to 0.0160, 0.0106, and 0.0220, while additionally removing the user profile increases them to 0.0400, 0.0246, and 0.0500. These results indicate that memory and user-profile modeling are important for behavioral agreement under a fixed backend. They do not establish user-satisfaction prediction or the ranking of alternative recommenders in a live A/B test. The fixed-backend agreement protocol is detailed in Appendix~\ref{App:ab_test}.
\begin{table}[ht]
\centering

\caption{Performance comparison of PsyBer-Agent and Human Users across different configurations.}
\label{tab:psyber_results}

\scriptsize
\begin{tabular}{@{}llcccccc@{}}
\toprule
\multirow{2}{*}{\textbf{Perspective}} & \multirow{2}{*}{\textbf{Method}} & \multicolumn{3}{c}{\textbf{PsyBer-Agent}} & \multicolumn{3}{c}{\textbf{Human Users}} \\
\cmidrule(lr){3-5} \cmidrule(lr){6-8}
 & & HR@10 $\uparrow$ & NDCG@10 $\uparrow$ & CTR $\uparrow$ & HR@10 $\uparrow$ & NDCG@10 $\uparrow$ & CTR $\uparrow$ \\
\midrule
Backbone Scaling & \makecell{PsyBer-Agent + \\ Qwen3-VL-235B} & 0.2350 & 0.1150 & 0.2960 & 0.2380 & 0.1166 & 0.3000 \\
\midrule
\multirow{3}{*}{Mechanism Validation} & PsyBer-Agent & 0.2350 & 0.1150 & 0.2960 & 0.2380 & 0.1166 & 0.3000 \\
& w/o Memory & 0.2220 & 0.1060 & 0.2780 & 0.2380 & 0.1166 & 0.3000 \\
& w/o User Profile & 0.1980 & 0.0920 & 0.2500 & 0.2380 & 0.1166 & 0.3000 \\
\bottomrule
\end{tabular}
\end{table}

\begin{checklistbox}{Findings 1}
PsyBer-Agent's predictions show low discrepancy with human behavior under the tested fixed-backend protocol, and both memory and user-profile modeling are key to preserving this agreement.
\end{checklistbox}

\newpage

\textbf{Agent Behavior Fidelity Evaluation.}

\begin{wraptable}{r}{0.42\textwidth}
\centering
\small
\setlength{\tabcolsep}{2.4pt}
\renewcommand{\arraystretch}{1.02}
\vspace{-10pt}
\caption{Behavioral realism results across scenarios and recommender backbones.}
\label{tab:behavioral_realism}
\resizebox{1.0\linewidth}{!}{%
\begin{tabular}{@{}llccc@{}}
\toprule
\textbf{Scenarios} & \textbf{Rec Models} & \textbf{JSD$\downarrow$} & \textbf{Pearson R$\uparrow$} & \textbf{EMD$\downarrow$} \\
\midrule
\multirow{6}{*}{\textbf{MM-ML-1M}}
& BPR-MF    & 0.3304 & 0.1615 & 0.3711 \\
& LightGCN  & 0.5603 & 0.1867 & 0.1649 \\
& Multi-VAE & 0.3249 & 0.1217 & 0.3673 \\
& GRU4Rec   & 0.3551 & 0.1755 & 0.3643 \\
& SASRec    & 0.6515 & 0.1803 & 0.4602 \\
& BERT4Rec  & 0.3292 & 0.1796 & 0.3683 \\
\midrule
\multirow{6}{*}{\makecell[l]{\textbf{Amazon}\\\textbf{Magazine Subscriptions}}}
& BPR-MF    & 0.2085 & 0.5164 & 0.2315 \\
& LightGCN  & 0.2126 & 0.5466 & 0.2245 \\
& Multi-VAE & 0.2094 & 0.5436 & 0.2333 \\
& GRU4Rec   & 0.2105 & 0.5455 & 0.2386 \\
& SASRec    & 0.2106 & 0.5437 & 0.2357 \\
& BERT4Rec  & 0.2173 & 0.5500 & 0.2491 \\
\bottomrule
\end{tabular}}

\end{wraptable}

We assess PsyBer-Agent's behavioral realism across scenarios and recommender backbones using three complementary metrics: JSD for population-level interaction-distribution mismatch, Pearson correlation ($R$) for item-level agreement with real interaction frequencies, and EMD for downstream behavioral discrepancy after matched prefixes. As shown in~\autoref{tab:behavioral_realism} and~\autoref{fig:group_analysis}, PsyBer-Agent preserves realistic behavioral patterns across scenarios and backbones, with stronger alignment on Amazon Magazine Subscriptions than on MM-ML-1M. Amazon Magazine Subscriptions forms compact low-JSD/high-$R$/low-EMD clusters, while MM-ML-1M is more dispersed, suggesting that narrower e-commerce preferences are easier to simulate than semantically diverse movie-consumption dynamics.

To further summarize profile-conditioned fidelity, we compute a composite fidelity score in \autoref{fig:group_analysis} by combining normalized JSD, Pearson $R$, and EMD into a single diagnostic ranking score. The resulting group-level patterns show that identity-gap and value-oriented groups achieve the highest fidelity, indicating that PsyBer-Agent preserves stable behavioral signatures across heterogeneous profile groups. Overall, the remaining mismatch appears more related to scenario and recommender effects than to unstable persona conditioning. Detailed group-level analysis results corresponding to \autoref{fig:group_analysis} are provided in Appendix~\ref{App:Group_Fid}.

\begin{checklistbox}{Findings 2}
PsyBer-Agent preserves group-level behavioral signatures across scenarios, and the remaining fidelity gap appears to be driven more by scenario complexity and recommender exposure than by unstable persona conditioning.
\end{checklistbox}

\textbf{Baseline Simulator Comparison.}

\begin{wraptable}{r}{0.64\linewidth}
\centering
\setlength{\tabcolsep}{2.1pt}

\caption{Basic simulator fidelity comparison.}

\label{tab:baseline_simulator_basic}
\scriptsize
\begin{tabular}{lccccc}
\toprule
\textbf{Simulator} & \textbf{Acc.} $\uparrow$ & \textbf{CTRGap} $\downarrow$ & \textbf{JSD} $\downarrow$ & \textbf{Pearson R} $\uparrow$ & \textbf{Latency} $\downarrow$ \\
\midrule
RecAgent~\cite{10.1145/3708985} & 0.212 & 0.482 & 0.420 & 0.264 & 5.46s \\
Agent4Rec~\cite{zhang2024generative} & 0.193 & 0.489 & 0.447 & 0.264 & 5.74s \\
UserSim~\cite{zhang2025llm} & 0.128 & 0.500 & 0.605 & 0.142 & 3.31s \\
DyTA4Rec~\cite{10.1145/3746252.3760878} & 0.254 & 0.455 & 0.335 & 0.281 & 5.91s \\
\midrule
\textbf{PsyBer-Agent} & \textbf{0.275} & \textbf{0.406} & \textbf{0.286} & \textbf{0.305} & \textbf{2.53s} \\
\bottomrule
\end{tabular}

\end{wraptable}

To compare PsyBer-Agent with existing user-simulation baselines under a controlled setting, we conduct a fixed-candidate replay experiment on MM-ML-1M with 500 users, 10 decision steps per user, and five random seeds. At each step, all simulators observe the same user histories, item metadata, and candidate pages generated under LightGCN and SASRec, and are evaluated on decision accuracy, CTR calibration, distributional fidelity, item-level correlation with real logs, and latency. As shown in ~\autoref{tab:baseline_simulator_basic}, PsyBer-Agent achieves the best overall performance in this controlled comparison, with the highest accuracy and correlation as well as the lowest CTRGap, JSD, and inference latency among all methods. These results suggest that explicit psychological state modeling yields a more faithful user-behavior prior for replay-based recommendation simulation.

\textbf{PsyBer-Agent Scalability \& Cost Evaluation.} We evaluate whether PsyBer-Agent remains efficient under large-scale workloads by running $N\in\{500,1\mathrm{K},100\mathrm{K},1\mathrm{M},5\mathrm{M}\}$ simulation tasks over 3 runs and reporting throughput, p95 latency/decision, wall-clock time, success rate, and token cost. Detailed metric definitions are provided in Appendix~\ref{app:metric_sca}. As shown in~\autoref{tab:scalability_cost}, \textsc{PsyBer-Agent} scales to $5\mathrm{M}$ tasks with bounded p95 latency growth (2.52s to 3.80s), success rates above 91\%, aggregate throughput of approximately 4.1 tasks/s after queue warm-up for workloads of at least $N\geq100\mathrm{K}$, and explicit wall-clock/token-cost accounting, indicating stable million-scale simulation under external model-call overhead. Hardware settings, unit cost, and fallback statistics are provided in Appendix~\ref{App:exp_hard}.

\begin{table}[ht]

\caption{Scalability and cost of PsyBer-Agent under large-scale workloads.}
\label{tab:scalability_cost}

\centering
\scriptsize
\setlength{\tabcolsep}{2.6pt}
\renewcommand{\arraystretch}{1.04}
\begin{tabularx}{\columnwidth}{@{}
    >{\centering\arraybackslash}p{1.9cm}
    >{\centering\arraybackslash}p{2.4cm}
    >{\centering\arraybackslash}p{2.4cm}
    >{\centering\arraybackslash}p{2.1cm}
    >{\centering\arraybackslash}p{1.8cm}
    >{\centering\arraybackslash}p{1.8cm}
@{}}
\toprule
\shortstack[c]{\textbf{Workload} \\ \textbf{Size} $N$}
& \shortstack[c]{\textbf{Throughput}\\\textbf{(tasks/s)} $\uparrow$}
& \shortstack[c]{\textbf{p95 Lat.}\\\textbf{/Decision (s)} $\downarrow$}
& \shortstack[c]{\textbf{Wall-clock}\\\textbf{Time} $\downarrow$}
& \shortstack[c]{\textbf{Success}\\\textbf{Rate} $\uparrow$}
& \shortstack[c]{\textbf{Token}\\\textbf{Cost} $\downarrow$} \\
\midrule
\textbf{500}  & 1.00 & 2.52s & 8.3 mins    & 100.00\% & 4262.7K   \\
\textbf{1K}   & 0.94 & 2.65s & 17.8 mins   & 100.00\% & 10265.3K  \\
\textbf{100K} & 4.18 & 2.74s & 398.4 mins  & 92.75\%  & 983.31M   \\
\textbf{1M}   & 4.12 & 3.27s & 4058.8 mins & 91.34\%  & 10277.83M \\
\textbf{5M}   & 4.09 & 3.80s & 20358.9 mins & 95.18\% & 52262.77M \\
\bottomrule
\end{tabularx}

\end{table}
\begin{checklistbox}{Findings 3}
PsyBer-Agent maintains bounded latency and high completion rates up to $5\mathrm{M}$ tasks. Scaling behavior is dominated by predictable growth in wall-clock time and token cost.
\end{checklistbox}

\subsection{Ablation Study}
\label{sec:abl}

\subsubsection{Component Ablation}
\label{sec:abl_comp}

\begin{wrapfigure}{r}{0.54\textwidth}
    \centering
    \captionsetup{font=footnotesize}
    \captionof{table}{Ablation study of PsyBer-Agent components.}
    \label{tab:abl_comp}
    \begingroup
    \scriptsize
    \setlength{\tabcolsep}{3.2pt}
    \renewcommand{\arraystretch}{0.9}
    \begin{tabular}{lccc}
        \toprule
        \multirow{2}{*}{\textbf{Architecture / Variants}}
        & \multicolumn{3}{c}{\textbf{Simulation Fidelity}} \\
        \cmidrule(l){2-4}
        & \textbf{CTRGap ($\downarrow$)} & \textbf{Exp-JS ($\downarrow$)} & \textbf{CES-MSE ($\downarrow$)} \\
        \midrule
        \textit{Direct Agent w/ Prompt} & 0.8926 & 0.7852 & 0.1654 \\
        \midrule
        \textit{w/o} Preprocessing Agent & 0.8652 & 0.7613 & 0.1290 \\
        \textit{w/o} Psy-Engine & 0.7469 & 0.7382 & 0.1154 \\
        \textit{w/o} Optimal Transport & 0.6906 & 0.6817 & 0.1053 \\
        \textit{w/o} Fusion Memory & 0.2985 & 0.6764 & 0.1254 \\
        \midrule
        \textbf{PsyBer-Agent (Ours Full)} & \textbf{0.2608} & \textbf{0.4297} & \textbf{0.0636} \\
        \bottomrule
    \end{tabular}
    \endgroup
    
\end{wrapfigure}
We first examine the constituent components of PsyBer-Agent by ablating the preprocessing agent, Psy-Engine, unbalanced Gromov--Wasserstein alignment, and fusion memory. As shown in Table~\ref{tab:abl_comp}, the full configuration achieves the lowest CTRGap, Exp-JS, and CES-MSE among the evaluated variants. Removing the preprocessing agent, Psy-Engine, or transport alignment increases one or more discrepancies, suggesting that structured input conditioning, model-defined state transitions, and distribution-level trajectory alignment provide complementary benefits under the tested protocol. Removing fusion memory has a smaller effect on CTRGap than removing several other components, but increases CES-MSE, indicating that memory remains useful for maintaining internal state consistency. These results support the contribution of the individual modules without implying psychological validity of the latent states.

\subsubsection{Information Ablation}
\label{sec:abl_info}
\begin{wrapfigure}{r}{0.54\textwidth}
    \centering
    
    \captionsetup{font=footnotesize}
    \captionof{table}{Ablation study of information channels in PsyBer-Agent across simulation-fidelity metrics.}
    \label{tab:abl_info}
    
    \begingroup
    \scriptsize
    \setlength{\tabcolsep}{3.2pt}
    \renewcommand{\arraystretch}{0.9}
    \begin{tabular}{lccc}
        \toprule
        \multirow{2}{*}{\textbf{Architecture / Variants}}
        & \multicolumn{3}{c}{\textbf{Simulation Fidelity}} \\
        \cmidrule(l){2-4}
        & \textbf{CTRGap ($\downarrow$)} & \textbf{Exp-JS ($\downarrow$)} & \textbf{CES-MSE ($\downarrow$)} \\
        \midrule
        \textit{Direct Agent w/ Prompt} & 0.8926 & 0.7852 & 0.1654 \\
        \midrule
        \textit{w/o} Item Information & 0.7824 & 0.6770 & 0.1282 \\
        \textit{w/o} Psychological Changes & 0.8816 & 0.6804 & 0.1254 \\
        \textit{w/o} User Identity Profile & 0.6738 & 0.6761 & 0.1086 \\
        \textit{w/o} Scene Information & 0.5243 & 0.6783 & 0.0855 \\
        \midrule
        \textbf{PsyBer-Agent (Ours Full)} & \textbf{0.2608} & \textbf{0.4297} & \textbf{0.0636} \\
        \bottomrule
    \end{tabular}
    \endgroup
    
\end{wrapfigure}
Beyond component-level ablations, we evaluate the information channels used by PsyBer-Agent, including item textual and visual information, psychological-change signals, user identity profiles, and scene information. As shown in Table~\ref{tab:abl_info}, the full configuration obtains the best overall results under the rerun protocol, while removing individual channels increases at least one of the reported discrepancies. The observed changes indicate that item content, dynamic state signals, user information, and scene context provide complementary conditioning information. These ablations measure sensitivity to the available input channels and do not constitute independent evidence that the latent variables correspond to psychologically validated traits. All ablation variants use the same data split, candidate-replay protocol, and random-seed configuration; variants that include the fast policy are retrained under their corresponding ablated state or information setting.

\section{Related work}
\label{sec:related_work}

\paragraph{User Simulation for Recommender Evaluation.} Recommender evaluation is moving from static log replay toward LLM-driven user simulation. Recent methods improve realism through persona and working-memory architectures~\cite{chen2025recusersim, liu2025recoworld, bernard2025usersimcrs}, align simulated behavior with human preferences via feedback distillation~\cite{wei2025mirroring}, leverage latent reasoning signals for recommendation interactions~\cite{zhang2025llm}, and generate human-camouflaged adversarial users~\cite{gu2025llm}. MMRC also reveals memory degradation and error accumulation in sustained multimodal conversations~\cite{xue-etal-2025-mmrc}. Long-horizon agent benchmarks also isolate memory performance across tool-use, research, coding, and web tasks~\cite{xu2026memgym}. However, most simulators remain prompt-dependent and model plausible local actions rather than explicit internal dynamics such as fatigue, curiosity, and self-consistency over long horizons. PsyBer-Agent addresses this gap with an explicit psychological engine for motivation-aligned behavior beyond prompt-only imitation.

\paragraph{Psychological User Modeling and Behavioral Calibration.} Psychological and calibration-oriented modeling has been used to improve user and agent realism through Big Five profiles~\cite{ma2025pub, zhao2025percrs}, contextual cognitive alignment~\cite{feng2025contextualizing}, broader value and motivation modeling~\cite{peng2025survey, gebreegziabher2026behavioral}, and co-evolving user-system feedback~\cite{kim2026selfevolverec}. Beyond recommendation, generative-agent calibration emphasizes behavior-level alignment, including implicit alignment to target behavioral distributions~\cite{wang2025implicit}, uncertainty-aware decision calibration~\cite{solopova2026strategic}, reproducible social simulation~\cite{taillandier2025integrating}, and multi-level, context-sensitive alignment criteria~\cite{zeng2025multilevel} across mobility~\cite{vu2025modeling}, economic-expectation~\cite{gusella2025expectations}, and public-health simulations~\cite{hou2025society}. StockAgent simulates investor behavior~\cite{zhang2024stockagent}, while PRISM models affective dynamics in online discourse~\cite{lu2026prism}. PsyBer-Agent couples psychological dynamics with log-based trajectory alignment for recommender evaluation.

\paragraph{Robustness, Safety, and Large-scale Simulation.} LLM-centered recommender systems and agent-based simulation introduce new robustness and safety risks, including targeted poisoning~\cite{guan2026venomrec}, persistent backdoors~\cite{ning2025backdoor}, fragile explanation modules~\cite{zhang2026robustexplain}, and membership-inference leakage~\cite{he2025membership}. Web-based agents also face hidden prompt injection through malicious fonts in external content~\cite{xiong-etal-2025-invisible, Zhang2024GoalGuided, Zhang2024TargetDriven,Zhang2026SemanticPreservingPromptHijacking}. Agentic attacks further increase realism through LLM-optimized manipulation trajectories~\cite{ning2024cheatagent} and simulator-leveraged shilling~\cite{gu2025simulation}, while LLM-as-recommender systems remain susceptible to preference-oriented biases and exploits~\cite{tang2026trustable}. These risks motivate user simulators that support behavior fidelity together with scalable stress testing, robustness analysis, privacy-aware evaluation, and controlled pre-deployment diagnosis.

\section{Conclusion}
\label{Sec:conclusion}

We presented PsyBer-Agent, a psychology-inspired user simulator that combines model-defined latent-state dynamics, offline UGW alignment, and a fast policy with LLM/VLM fallback for closed-loop recommender evaluation. Across 200 runs and 3,981 decisions, it achieved 3.11\,s average latency, full fast-hint coverage, and a 99.9\% batch success rate. On the PsyBer Benchmark, it reduced behavioral discrepancy and improved model-defined state consistency over prompt-based baselines under fixed replay. These results support its use as a pre-deployment simulation proxy, but do not establish psychological validity or predict live A/B rankings. Future work will add independently validated supervision and broaden evaluation across domains, horizons, and robustness settings.

\newpage

\section*{Statements}
\subsection*{AI use statement}
In this work, generative AI tools were used for language polishing and minor editorial assistance during manuscript preparation, as well as for coding assistance in the engineering of the WebSim platform and experimental scripts. All LLM-generated text and code were reviewed, verified, and tested by the authors. The research ideas, methodology, experimental design, and all reported results were produced and validated by the authors, who take full responsibility for the final content of this work, including any text or artifacts produced with the aid of generative AI.

\subsection*{Ethics Statement}

This work uses publicly available recommendation datasets and simulated user interactions. In addition, the fixed-backend human--simulator agreement study in Section~\ref{sec:exp} involved 500 human participants, who were genuine anonymous users of the website. All participants provided informed consent before the experiment and were informed of the purpose, procedure, and duration of the study; participation was voluntary and could be withdrawn at any time without penalty. The experiment consisted solely of browsing recommendation pages and making click/skip decisions, involved no sensitive personal information, and posed minimal risk. The study was approved by the ethics committee of the authors' institution. Collected interaction logs (clicks and skips) are stored in anonymized form, are used only for research evaluation, and will not be linked to personally identifiable information. PsyBer-Agent is intended as a diagnostic proxy for pre-deployment evaluation, not as the sole basis for deployment decisions. Potential risks—including manipulative recommendation strategies, engagement-driven feedback loops, privacy concerns from interaction logs, and fairness issues across user groups—are discussed in the broader-impact statement (Appendix~\ref{App:broad}) together with mitigation guidelines such as responsible data handling, fairness checks, human oversight, and staged online validation.

\subsection*{Reproducibility statement}

We provide detailed descriptions of the Psy-Engine transition rules and hyperparameters (Appendix~\ref{App:Psy-Engine}), the UGW alignment objective and its initialization (Appendix~\ref{App:Psy-be_alignment}), the WebSim platform implementation (Appendix~\ref{App:websim}), all evaluation metrics (Appendix~\ref{App:metrics_eval}), the hardware and concurrency setup (Appendix~\ref{App:exp_hard}), and the fixed-backend human--simulator agreement protocol, including participant recruitment and the unified metric computation. Source code for PsyBer-Agent and the WebSim platform is available at \url{https://anonymous.4open.science/r/CC66F/} and \url{https://anonymous.4open.science/r/D8EAX/}, respectively.

\newpage

\bibliography{conference}
\bibliographystyle{colm2026_conference}

\newpage

\appendix
\begin{appendices}
    \startcontents[app]
    
    \begin{center}
        \textbf{\Large Content of Appendix}
    \end{center}
    
    \printcontents[app]{}{1}{}
    \newpage
\newpage


\section{Design of Psy-Engine: A Psychological State Engine for the PsyBer-Agent}
\label{App:Psy-Engine}

\subsection{Initial Settings: Core State, Input Features, and Persona}
\label{App:Psy-Engine_init}

In this part, we pre-process the information observable to the agent to reduce the burden of online computation. At interaction step $t$, Psy-Engine maintains a three-dimensional state
\begin{equation}
    \mathbf{z}_t=[C_t,E_t,S_t]^\top\in[0,1]^3,
\end{equation}
where $C_t$, $E_t$, and $S_t$ denote cognitive reserve, emotional satisfaction, and self-congruence at time $t$, respectively. Each candidate item $i_t$ is represented by
\begin{equation}
    \phi(i_t)=\left(\mathbf{v}_t,n_t,c_t,p_t\right),
\end{equation}
where $\mathbf{v}_t\in\mathbb{R}^d$ is the item embedding, $n_t\in[0,1]$ measures item-level incongruence or negative-preference strength, $c_t\in[0,1]$ denotes cognitive load, and $p_t\in[0,1]$ is the social-proof score.

\subsection{Cross-domain feature construction.} 
\label{App:cross_domain}
For cross-domain recommendation scenarios, we map heterogeneous item metadata into a unified psycho-behavioral feature tuple. Textual fields and visual assets are encoded into normalized embeddings and fused to obtain $v_t$. The incongruence feature $n_t$ measures the mismatch between the item and the user's current preference or persona anchor, and may incorporate text-image inconsistency as an item-level prior. The cognitive-load feature $c_t$ is estimated from content complexity, including text length, metadata density, and visual complexity. The social-proof feature $p_t$ is computed from observable signals such as ratings, review counts, click/like rates, or platform hotness indicators.

All signals are computed from pre-window item metadata and normalized over the global item set. Social-proof features use only ratings, review counts, popularity, or logged click/like statistics available before the evaluated interaction. All scalar features are normalized to $[0,1]$ over the global item set, ensuring that each item retains stable psycho-behavioral values across interaction steps.

The user persona contains three anchors, $\mathbf{u}_{\mathrm{base}},\ \mathbf{u}_{\mathrm{ideal}},\ \mathbf{u}_{\mathrm{real}}\in\mathbb{R}^d$, and we denote the persona by
\begin{equation}
    \psi=\left(\mathbf{u}_{\mathrm{base}},\mathbf{u}_{\mathrm{ideal}},\mathbf{u}_{\mathrm{real}}\right).
\end{equation}
Here, $\mathbf{u}_{\mathrm{base}}$ summarizes the user's static profile and metadata, $\mathbf{u}_{\mathrm{ideal}}$ captures long-term preferences from historical interactions, and $\mathbf{u}_{\mathrm{real}}$ captures recent preferences from the latest valid interactions.
For completeness, the three anchors are computed as follows. Let $\mathbf{e}(x)$ denote the encoder for a profile field, metadata field, or item, and let $\mathbf{P}_{\mathrm{base}}$, $\mathbf{P}_{\mathrm{ideal}}$, and $\mathbf{P}_{\mathrm{real}}$ project the corresponding representations into a shared $d$-dimensional persona space. 

If $m^{\mathrm{base}}_{u,j}\in\{0,1\}$ indicates availability of field $j$, then

\begin{equation}\label{eq:u_base}
\mathbf{u}_{\mathrm{base}} = \mathbf{P}_{\mathrm{base}}\left(\frac{\sum_{j=1}^{J}m^{\mathrm{base}}_{u,j}\mathbf{e}(x_{u,j})}{\max\left(1,\sum_{j=1}^{J}m^{\mathrm{base}}_{u,j}\right)}\right).
\end{equation}

Let $\mathcal{H}^{\mathrm{long}}_u$ denote the valid pre-window historical interactions and let $\mathbf{v}_i$ be the representation of item $i$. With interaction mask $m_{u,i}$, the long-term anchor is

\begin{equation}\label{eq:u_ideal}
\mathbf{u}_{\mathrm{ideal}} = \mathbf{P}_{\mathrm{ideal}}\left(\frac{\sum_{i\in\mathcal{H}^{\mathrm{long}}_u}m_{u,i}\mathbf{v}_i}{\max\left(1,\sum_{i\in\mathcal{H}^{\mathrm{long}}_u}m_{u,i}\right)}\right).
\end{equation}

Let $\mathcal{H}^{K}_u=(i_{u,1},\ldots,i_{u,K})$ be the most recent $K$ valid interactions in temporal order, with mask $m^K_{u,k}$. The recent-interaction anchor is

\begin{equation}\label{eq:u_real}
\mathbf{u}_{\mathrm{real}} = \mathbf{P}_{\mathrm{real}}\left(\frac{\sum_{k=1}^{K}m^K_{u,k}\mathbf{v}_{i_{u,k}}}{\max\left(1,\sum_{k=1}^{K}m^K_{u,k}\right)}\right).
\end{equation}

If no valid field or interaction is available, the corresponding numerator is replaced by a fixed cold-start vector and the availability mask is set to zero. The anchors are used as conditioning representations and are not interpreted as direct measurements of psychological traits.The engine also maintains a bounded memory sequence

\begin{equation}
    \mathcal{M}_t=[\mathbf{m}_1,\ldots,\mathbf{m}_{|\mathcal{M}_t|}],\qquad |\mathcal{M}_t|\le M_{\max},
\end{equation}

where $M_{\max}$ is the memory capacity and each memory entry $\mathbf{m}$ is a normalized embedding of a previously consumed item. When the memory reaches $M_{\max}$, newly consumed items are inserted using a first-in-first-out update rule. We further denote by $\mathcal{M}_t^{(h)}$ the subsequence consisting of the most recent $\min(h,|\mathcal{M}_t|)$ entries of $\mathcal{M}_t$, where $h\in\mathbb{N}_+$ is the short-term memory horizon. The engine uses a global update rate $\alpha\in[0,1]$ to control the sensitivity of state transitions.

For each item $i_t$, we pre-compute item-level features $(\mathbf{v}_t,c_t,p_t)$ during offline preprocessing, while $n_t$ is computed with respect to the current user's preference anchor. The item embedding $\mathbf{v}_t$ is obtained by fusing the item's textual and visual features, and is normalized as
\begin{equation}
\hat{\mathbf{v}}_t=\frac{\mathbf{v}_t}{\max(\|\mathbf{v}_t\|_2,\epsilon)},
\end{equation}
where $\epsilon>0$ is a small smoothing constant for numerical stability. The incongruence score $n_t$ is computed as the semantic distance between the item and the user's current preference anchor:
\begin{equation}
n_t = 1-\max(0,\cos(\hat{\mathbf{v}}_t,\mathbf{u}^{\mathrm{pref}}_t)),
\end{equation}
where $\mathbf{u}^{\mathrm{pref}}_t$ denotes the current preference anchor, instantiated by $\mathbf{u}_{\mathrm{base}}$ unless otherwise specified.

The cognitive-load score $c_t$ is estimated from the intrinsic complexity of item content and normalized to $[0,1]$:
\begin{equation}
    c_t = \operatorname{Norm}_{\mathcal{I}}\big(\mathrm{Comp}(i_t)\big),
\end{equation}
where $\mathrm{Comp}(i_t)$ denotes the raw content-complexity score, computed from signals such as text length, metadata density, and visual complexity. The social-proof score $p_t$ is computed from observable social feedback and is also normalized to $[0,1]$:
\begin{equation}
    p_t = \operatorname{Norm}_{\mathcal{I}}\big(\mathrm{Social}(i_t)\big),
\end{equation}
where $\mathrm{Social}(i_t)$ is instantiated by aggregate feedback signals such as item rating, popularity, review count, or, in WebSim, other agents' like and click rates. Here, $\operatorname{Norm}_{\mathcal{I}}(\cdot)$ denotes min--max normalization over the global item set $\mathcal{I}$, rather than over the current candidate set, so that the same item has a stable normalized score across different interaction steps.

\subsection{Similarity, Projection, and Indicator Operators}
\label{App:Psy-Engine_sim}

We use cosine similarity
\begin{equation}
\cos(\mathbf{a},\mathbf{b})=
\begin{cases}
\dfrac{\mathbf{a}^{\top}\mathbf{b}}{\|\mathbf{a}\|_2\|\mathbf{b}\|_2},
& \|\mathbf{a}\|_2\|\mathbf{b}\|_2>0,\\
0, & \text{otherwise},
\end{cases}
\end{equation}
and the projection operator
\begin{equation}
\operatorname{Proj}_{[0,1]}(x)
=
\min(1,\max(0,x)).
\end{equation}

We also use the indicator function
\begin{equation}
\mathbf{1}[\mathcal{E}]
=
\begin{cases}
1, & \text{if the event or condition } \mathcal{E} \text{ is true},\\
0, & \text{otherwise}.
\end{cases}
\end{equation}
For example, $\mathbf{1}[i_r=y]$ equals $1$ when the $r$-th
expert item is $y$ and $0$ otherwise. It is used to aggregate the
alignment weights associated with the same item identity.

\subsection{Intermediate Drives}
\label{App:Psy-Engine_factors}

Given item $i_t$, novelty $\nu_t$ and semantic saturation $\sigma_t$ are defined as
\begin{equation}
    \nu_t=
    \begin{cases}
        \operatorname{Proj}_{[0,1]}\!\left(1-\max_{\mathbf{m}\in\mathcal{M}_t}\cos(\hat{\mathbf{v}}_t,\mathbf{m})\right), & |\mathcal{M}_t|>0,\\
        1, & |\mathcal{M}_t|=0,
    \end{cases}
\end{equation}
and
\begin{equation}
    \sigma_t=
    \begin{cases}
        \operatorname{Proj}_{[0,1]}\!\left(
        \dfrac{1}{|\mathcal{M}_t^{(h)}|}
        \sum_{\mathbf{m}\in\mathcal{M}_t^{(h)}} \max\!\big(0,\cos(\hat{\mathbf{v}}_t,\mathbf{m})\big)
        \right), & |\mathcal{M}_t^{(h)}|>0,\\
        0, & |\mathcal{M}_t^{(h)}|=0.
    \end{cases}
\end{equation}

The resulting drive factors are
\begin{equation}
    A_t=\frac{\nu_t+n_t}{2},\qquad
    F_t=\frac{c_t+\sigma_t}{2},
    \label{eq:attractiveness_fatigue}
\end{equation}

These factors provide a compact decomposition of the psychological drives underlying user behavior. 

In \eqref{eq:attractiveness_fatigue}, $A_t$ and $F_t$ denote attentional attractiveness and cognitive fatigue, whose difference forms the net cognitive flux. The design of $A_t$ follows the intuition that novelty and preference inconsistency stimulate exploratory interest~\cite{loewenstein1994psychology, silvia2008interest}, while $F_t$ captures the dual depletion caused by intrinsic cognitive load and repeated exposure to semantically similar items~\cite{sweller1988cognitive, 10.1086/521898}. 

We further compute the semantic-fit and serendipitous-interest signals as $\beta_t=\max(0,\cos(\hat{\mathbf{v}}_t,\mathbf{u}_{\mathrm{base}}))$ and $\rho_t=\nu_t\beta_t$, respectively. The affective target is then given by $E_t^\star=(\beta_t+\rho_t+p_t)/3$, combining semantic fit, novelty-driven serendipity, and social proof~\cite{KOTKOV2016180, cialdini2004social}. Finally, the self-conflict signal is defined as $\gamma_t=\operatorname{Proj}_{[0,1]}(\max(0,\cos(\hat{\mathbf{v}}_t,\mathbf{u}_{\mathrm{real}})-\cos(\hat{\mathbf{v}}_t,\mathbf{u}_{\mathrm{ideal}})))$, and the self-alignment target is $S_t^\star=\operatorname{Proj}_{[0,1]}(\beta_t-\gamma_t)$, following the rationale that self-fit enhances self-congruence while self-discrepancy or guilt weakens it~\cite{higgins1987self, sirgy1982self}. Together, these quantities replace multi-term hand-crafted penalties with a small set of interpretable drives that remain stable across scenarios.

\subsection{State Transition Equations}
\label{App:Psy-Engine_state_trans}

The three state variables are updated from the current state $\mathbf{z}_t$ using the projection operator $\operatorname{Proj}_{[0,1]}$, consistent with Eq.~(\ref{eq:psy_state_update}) in the main text:
\begin{equation}
    C_{t+1} = \operatorname{Proj}_{[0,1]}\!\left(C_t+\alpha(A_t-F_t)\right),
    \label{eq_app:t1}
\end{equation}
\begin{equation}
    E_{t+1} = \operatorname{Proj}_{[0,1]}\!\left((1-\alpha)E_t+\alpha E_t^\star\right),
    \label{eq_app:t2}
\end{equation}
\begin{equation}
    S_{t+1} = \operatorname{Proj}_{[0,1]}\!\left((1-\alpha)S_t+\alpha S_t^\star\right).
    \label{eq_app:t3}
\end{equation}

The cognitive state responds to the net cognitive flux $A_t-F_t$, while emotional satisfaction and self-congruence move toward their interaction-dependent targets $E_t^\star$ and $S_t^\star$. This transition is a model-defined computational hypothesis; it does not imply independent psychological validity. The fixed Psy-Engine parameters are collectively denoted as $\Theta$, including $\alpha$, $h$, $M_{\max}$, the feature weights used in the drive functions, and the fixed feature-normalization statistics.

\section{Design of PsyBer-Agent Psycho-behavioral Alignment}
\label{App:Psy-be_alignment}

In PsyBer-Agent psycho-behavioral alignment, we describe how PsyBer-Agent aligns offline expert behavior with the simulator-side psychological state sequence. For clarity, we omit the user subscript and consider one expert sequence and one simulator sequence at a time.

Let the expert sequence be
\begin{equation}
    \tau^{\mathrm{exp}} = (i_1, \ldots, i_n),
\end{equation}
where $i_r$ denotes the clicked item at expert position $r$. Let the simulator-side psychological sequence be
\begin{equation}
    \mathcal{Z}^{\mathrm{sim}}=(\mathbf{z}_1,\ldots,\mathbf{z}_m),
\end{equation}
where
\begin{equation}
    \mathbf{z}_j=[C_j,E_j,S_j]^\top \in [0,1]^3
\end{equation}
is the three-dimensional psychological state defined by Psy-Engine, with $C_j$, $E_j$, and $S_j$ denoting cognition, emotional satisfaction, and self-congruence, respectively.

\subsection{Model-Defined State Sequence Construction}
\label{App:Align_con}

The offline log records expert item clicks but does not contain annotations of the latent states underlying those decisions. Because the observed clicks alone cannot identify a unique psychological trajectory, we use them as externally supplied actions and reconstruct a model-defined reference trajectory under the same persona-conditioned Psy-Engine used by the simulator. This reconstruction is used for trajectory alignment and policy supervision, rather than treated as a direct measurement of human psychology.

Starting from the persona-conditioned initial state $\tilde{\mathbf{z}}_0$ and replay memory $\tilde{\mathcal{M}}_0$, we replay the expert sequence $\tau^{\mathrm{exp}}=(i_1,\ldots,i_n)$ using teacher forcing, where each logged item is supplied as the consumed item at its corresponding step. Consistent with the transition operator in the main text, we recursively compute
\begin{equation}
\tilde{\mathbf{z}}_r =
\mathcal{T}_{\Theta}\!\left(
\tilde{\mathbf{z}}_{r-1},
\phi(i_r),
\tilde{\mathcal{M}}_{r-1}
\right),
\qquad r=1,\ldots,n.
\end{equation}
where $\mathcal{T}_{\Theta}$ denotes the item-level Psy-Engine transition, $\phi(i_r)=(\mathbf{v}_r,n_r,c_r,p_r)$ is the psycho-behavioral feature tuple of item $i_r$, and $\Theta$ collects the fixed Psy-Engine parameters. The term $\tilde{\mathcal{M}}_{r-1}$ denotes the replay memory before step $r$, storing normalized embeddings of previously replayed expert items. Thus, $\tilde{\mathbf{z}}_r$ is the model-defined latent reference state associated with the $r$-th logged interaction; it is not an independently observed psychological label.

After replaying $i_r$, the memory is updated by appending its normalized embedding and truncating the sequence to the most recent $M_{\max}$ entries. This update preserves the same bounded-memory convention used during online simulation, so that the reconstructed expert-side dynamics and simulator-side dynamics are evaluated under comparable state-transition conditions. This yields the model-defined reference sequence
\begin{equation}
\tilde{\mathbf{Z}}^{\mathrm{exp}} =
(\tilde{\mathbf{z}}_1,\ldots,\tilde{\mathbf{z}}_n),
\qquad
\tilde{\mathbf{z}}_r \in [0,1]^3.
\end{equation}

The resulting sequence is subsequently aligned with the simulator trajectory to transfer information from observed expert actions into state-conditioned soft supervision. In this process, the expert clicks determine the replay inputs, while the Psy-Engine determines the corresponding latent reference states.

For any state sequence $\mathbf{x}_{1:T}=(\mathbf{x}_1,\ldots,\mathbf{x}_T)$, we define its backward increments as
\begin{equation}
\Delta \mathbf{x}_k =
\begin{cases}
\mathbf{x}_k-\mathbf{x}_{k-1}, & 2\le k\le T,\\
\mathbf{0}, & k=1.
\end{cases}
\label{eq:state_increment}
\end{equation}
This convention is applied to the simulator, reconstructed expert-reference, and regressed expert pseudo-state sequences.

\subsection{Transport Initialization and UGW Alignment}
\label{App:matrix_init}

A direct uniform initialization of the transport matrix is too weak, because the expert item sequence and the simulator psychological sequence are heterogeneous and may have different lengths. To improve alignment stability, we discard sequences shorter than a minimum length threshold and truncate overly long sequences to a fixed maximum horizon during preprocessing. 

We therefore first learn an offline regression prior that maps expert-side prefixes into the same three-dimensional state space as the simulator. Specifically, we train a regression function $g_{\varphi}$ to predict the model-defined state prior:
\begin{equation}
    \hat{\mathbf{z}}_r = g_{\varphi}\!\left(\psi,\,\tau^{\mathrm{exp}}_{1:r},\,\phi(i_r),\,\frac{r}{n}\right)
\end{equation}

where $\psi$ denotes the user persona, $\tau^{\mathrm{exp}}_{1:r} = (i_1, \ldots, i_r)$ is the expert prefix up to position $r$, $\varphi$ denotes the regression parameters, $r/n$ represents the relative position of the current interaction $r$ within the full log sequence of length $n$, and $\hat{\mathbf{z}}_r$ is the regressed pseudo state aligned to the simulator state space. Let $\mathcal{D}_{\mathrm{exp}}$ denote the offline collection of expert browsing sequences. Here, $\tilde{\mathbf{z}}_r$ denotes the model-defined reference state reconstructed by replaying the expert sequence through Psy-Engine. The regression parameters are learned offline by minimizing
\begin{equation}
    \varphi^\star
    =
    \arg\min_{\varphi}
    \sum_{\tau^{\mathrm{exp}}\in\mathcal{D}_{\mathrm{exp}}}
    \sum_{r=1}^{n}
    \left\|
    g_{\varphi}\!\left(
        \psi,
        \tau^{\mathrm{exp}}_{1:r},
        \phi(i_r),
        \frac{r}{n}
    \right)
    -
    \tilde{\mathbf{z}}_r
    \right\|_2^2 .
\end{equation}

Here, $\varphi^\star$ denotes the parameter value that minimizes the regression objective; the superscript $\star$ marks the solution obtained by optimization. Using the increment rule in Eq.~(\ref{eq:state_increment}), we define the pseudo-state increments $\Delta\hat{\mathbf{z}}_r$ accordingly.

Given the simulator state sequence $\mathcal{Z}^{\mathrm{sim}}=(\mathbf{z}_1,\ldots,\mathbf{z}_m)$ and the regressed expert pseudo-state sequence $(\hat{\mathbf{z}}_1,\ldots,\hat{\mathbf{z}}_n)$, we define the initialization cost and use it to induce the initial transport weights as follows:
\begin{equation}\begin{array}{c@{\qquad}c}c^{\mathrm{init}}_{j,r} = \lambda_z\|\mathbf{z}_j-\hat{\mathbf{z}}_r\|_2^2
    + \lambda_{\Delta} \|\Delta \mathbf{z}_j-\Delta \hat{\mathbf{z}}_r\|_2^2
& \Pi_{j,r}^{(0)}\propto \exp\!\left(-\frac{c^{\mathrm{init}}_{j,r}}{\tau_0}\right),
\end{array}\end{equation}
Here, $c^{\mathrm{init}}_{j,r}$ is the prior matching cost, $\lambda_z\ge 0$ and $\lambda_{\Delta}\ge 0$ weight the state and local-change discrepancies, respectively, and $\tau_0>0$ controls the sharpness of the soft matching distribution. We then rescale $\Pi^{(0)}$ using the prescribed marginal distributions $a\in\mathbb{R}_+^m$ and $b\in\mathbb{R}_+^n$; unless otherwise specified, both marginals are uniform.

After initialization, we optimize $\Pi$ by Unbalanced Gromov-Wasserstein (UGW) alignment. On the expert side, let $\mathbf{v}_r$ denote the fused semantic embedding contained in $\phi(i_r)$. We define the expert-side internal distance $d_{\mathrm{exp}}$ as
\begin{equation}
    d_{\mathrm{exp}}(r,r') =
    \lambda_v \bigl(1-\cos(\mathbf{v}_r,\mathbf{v}_{r'})\bigr),     
\end{equation}
where $\lambda_v\ge 0$ controls semantic discrepancy. On the simulator side, the internal distance between two psychological states is defined as
\begin{equation}
    d_{\mathrm{sim}}(j,j') = \lambda_s\|\mathbf{z}_j-\mathbf{z}_{j'}\|_2
    + \lambda_{\Delta}^{\mathrm{sim}} \|\Delta \mathbf{z}_j-\Delta \mathbf{z}_{j'}\|_2,     
\end{equation}

where $\lambda_s \ge 0$ controls state discrepancy and $\lambda_{\Delta}^{\mathrm{sim}} \ge 0$ controls increment discrepancy. We assume a positive structural correspondence between the two spaces: simulator-state pairs that are farther apart should, on average, correspond to expert-item pairs that are more semantically dissimilar. Accordingly, we penalize the difference between their weighted distances. The structural discrepancy between relations $(j,j')$ and $(r,r')$ is
\begin{equation}
    \ell\bigl((j,j'),(r,r')\bigr)
    = \left(d_{\mathrm{sim}}(j,j') - d_{\mathrm{exp}}(r,r')\right)^2.
\end{equation}

Accordingly, the UGW objective is
\begin{equation}
    \mathcal{L}_{\mathrm{UGW}}(\Pi) =
    \sum_{j,j',r,r'}
    \ell\bigl((j,j'),(r,r')\bigr)
    \Pi_{j,r}\Pi_{j',r'} - \varepsilon H(\Pi)
    +\tau_{\mathrm{ugw}} \mathrm{KL}(\Pi\mathbf{1}_n \,\|\, a)
    +\tau_{\mathrm{ugw}} \mathrm{KL}(\Pi^\top \mathbf{1}_m \,\|\, b),
\end{equation}
\label{eq:ugw}
Here, $\Pi_{j,r}$ and $\Pi_{j',r'}$ denote the transport masses assigned to the two simulator--expert correspondences $(j,r)$ and $(j',r')$, respectively. Their product $\Pi_{j,r}\Pi_{j',r'}$, an entry of $\Pi\otimes\Pi$, weights the structural comparison between these correspondences. where
\begin{equation}
    H(\Pi) = -\sum_{j,r} \Pi_{j,r} \bigl(\log\Pi_{j,r}-1\bigr)
\end{equation}
is the entropy regularizer, and $\varepsilon>0$ is its weight. With this sign convention, the term $-\varepsilon H(\Pi)$ encourages a smoother transport plan during optimization. 

The coefficient $\tau_{\mathrm{ugw}}>0$ controls marginal relaxation, $\mathbf{1}_n$ and $\mathbf{1}_m$ denote all-ones vectors of dimensions $n$ and $m$, respectively, and $\mathrm{KL}(\cdot\|\cdot)$ denotes the Kullback--Leibler divergence. The term $\tau_{\mathrm{ugw}}\mathrm{KL}(\Pi\mathbf{1}_n \,\|\, a)$ penalizes deviations of the simulator-side marginal from $a$, while $\tau_{\mathrm{ugw}}\mathrm{KL}(\Pi^\top\mathbf{1}_m \,\|\, b)$ penalizes deviations of the expert-side marginal from $b$. Starting from the initialization $\Pi^{(0)}$, we obtain the final transport matrix by solving
\begin{equation}
    \Pi^\star = \arg\min_{\Pi\ge 0} \mathcal{L}_{\mathrm{UGW}}(\Pi).
\end{equation}

Because the optimization is unbalanced, the learned transport plan naturally tolerates different sequence lengths, browsing speeds, and locally unmatched segments.

\subsection{Alignment-induced Soft Supervision for Policy Learning}
\label{app:policy}

The offline expert logs contain clicked items but do not specify which action should be taken on the simulator's current candidate slate. After reconstructing the model-defined expert trajectory and aligning it with the simulator trajectory in Appendix~\ref{App:matrix_init}, the optimized transport matrix $\Pi^\star$ provides a step-wise correspondence between the two trajectories. We use this correspondence to construct state-conditioned soft action supervision for the fast policy.

The optimized transport matrix $\Pi^\star$ specifies which expert positions are most compatible with each simulator step. We first convert it into normalized matching weights:
\begin{equation}
    q_j(r) = \frac{\Pi^\star_{j,r}}{\sum_{r'=1}^{n}\Pi^\star_{j,r'}+\delta},
\end{equation}
where $\delta>0$ is a small constant for numerical stability. Based on $q_j(r)$, we define the item-level soft target distribution
\begin{equation}
    \hat{q}_j(y)
    =
    \sum_{r=1}^{n}
    q_j(r)\mathbf{1}[i_r=y].
\end{equation}
Here, $\hat{q}_j(y)$ represents the probability mass assigned to item identity $y$ under the alignment-induced supervision at simulator step $j$. This distribution reflects which expert items are most behaviorally compatible with the current simulator psychological state $\mathbf{z}_j$.

At simulator step $j$, let the current candidate slate be $\mathcal{L}_j=(\ell_{j,1},\ldots,\ell_{j,K_s})$, where $\ell_{j,k}$ is the $k$-th candidate item and $K_s$ is the slate size. The corresponding action set is $\mathcal{A}_j = \{\ell_{j,1},\ldots,\ell_{j,K_s},\mathrm{skip}\}$.

Here, $\mathrm{skip}$ denotes the no-click or continue-browsing action, implemented as page continuation in WebSim. Since an expert item may be absent from the current candidate slate, we project the item-level target onto $\mathcal{A}_j$:
\begin{equation}
\hat{p}_j(a) =
    \begin{cases}
    \hat{q}_j(\ell_{j,k}),
    & a=\ell_{j,k},\quad k=1,\ldots,K_s,\\[4pt]
    1-\displaystyle\sum_{k=1}^{K_s}
    \hat{q}_j(\ell_{j,k}),
    & a=\mathrm{skip}.
    \end{cases}
\end{equation}


The residual mass assigned to $\mathrm{skip}$ accounts for aligned expert items that are not available in the current slate, as well as unmatched trajectory mass. Therefore, the projection converts expert-side alignment information into a valid probability distribution over the actions currently executable by the simulator.

The resulting $\hat{p}_j$ provides dense, state-conditioned supervision even though the offline logs contain only sparse click observations. We use it to define the imitation objective for the online fast policy:

\begin{equation}
    \mathcal{L}_{\mathrm{imit}} = \sum_{j}\mathrm{CE}\!\left(\hat{p}_j, \pi_{\theta}(\cdot \mid \mathbf{z}^{\mathrm{sim}}_j, \mathcal{A}_j) \right),
\end{equation}

where $\pi_{\theta}(\cdot\mid\mathbf{z}^{\mathrm{sim}}_j,\mathcal{A}_j)$ is the policy distribution over the current simulator state and executable action set, parameterized by $\theta$, and $\mathrm{CE}(\cdot,\cdot)$ denotes cross-entropy. This objective transfers psycho-behaviorally aligned supervision into the lightweight online policy, rather than relying on sparse click replay alone.


\section{PsyBer Benchmark}
\label{App:benchmark}

\subsection{Benchmark Overview}

\textbf{PsyBer Benchmark} is a closed-loop benchmark for simulated-user evaluation in recommender systems. It is designed to evaluate whether a user simulator can (i) produce realistic and scenario-consistent interaction trajectories under online recommendation exposure, and (ii) support reliable evaluation of target recommendation models in controlled simulation environments. To this end, the benchmark standardizes four components: datasets, recommendation scenarios, target recommender backbones, and evaluation protocols. All experiments are conducted on the WebSim platform, which decouples user-interaction serving from recommender inference and enables consistent cross-model and cross-dataset comparisons. Detailed platform implementation and metric definitions are provided in Appendices~\ref {App:websim} and~\ref {App:metrics_eval}, respectively.

\subsection{Data Construction}
\label{App:data}

We construct the benchmark from seven public datasets spanning movie, e-commerce, short-video, and cross-domain recommendation. Specifically, we include \textbf{MM-ML-1M}~\cite{zhang2026exploring}, a multimodal extension of MovieLens-1M, to evaluate recommendation settings with rich item-side modalities. To test cross-domain robustness in shopping scenarios, we further adopt four re-edited subsets from the 2018 Amazon Reviews collection~\cite{ni2019justifying}: \textbf{Musical\_Instruments}, \textbf{Magazine\_Subscriptions}, \textbf{Beauty}, and \textbf{Luxury\_Beauty}. In addition, we include \textbf{MicroLens-100K}~\cite{ni2025content}, a short-video recommendation dataset, and \textbf{NineRec-Bili500K}~\cite{zhang2025ninerec}, a large-scale cross-domain recommendation dataset with multimodal item metadata.

To ensure fair comparisons across domains, all datasets are processed under a unified protocol for model training and evaluation, including standardized interaction formatting, scenario-specific item metadata construction, and chronological splitting for training, validation/test evaluation. This unified preprocessing allows the same simulator and recommender evaluation pipeline to be applied across heterogeneous recommendation settings.

\subsection{Task Definition}
\label{sec:task_def}

The PsyBer Benchmark consists of two core tasks. \textbf{(1)} Multi-scenario user online simulation. \textbf{(2)} Recommendation model evaluation.

\textbf{Task 1: Multi-scenario user online simulation.}
This task studies whether a simulator can produce realistic user behavior under closed-loop recommendation exposure. Given a recommendation environment, PsyBer-Agent sequentially interacts with the served slates and generates user feedback, including clicks, skips, and continued browsing. The resulting trajectories are used to evaluate whether the simulator preserves realistic interaction patterns, stable long-horizon behavior, and scenario-consistent responses across different recommendation domains.

\textbf{Task 2: Recommendation model evaluation.}
This task evaluates target recommendation models using PsyBer-Agent as a simulated user proxy. On the WebSim platform, under a fixed scenario setting, PsyBer-Agent interacts with each target recommender and generates closed-loop trajectories, which are then used to assess recommendation quality and system behavior. We report online recommendation metrics such as CTR and NDCG@$K$, along with calibration and efficiency metrics such as CTRGap and latency per decision, where $K$ is specified by the evaluation setting for each experiment.

\subsection{Benchmark Coverage and Evaluation Protocol}
\label{App:Ben_cov_eva}
To support broad and reproducible evaluation, the PsyBer Benchmark covers multiple recommendation backbones and application scenarios. The target recommender set includes representative methods from popularity-based, matrix-factorization-based, graph-based, variational, sequential, and transformer-based recommendation, including PopRec~\cite{cremonesi2010performance}, BPR-MF~\cite{rendle2012bpr}, LightGCN~\cite{he2020lightgcn}, Multi-VAE~\cite{liang2018variational}, GRU4Rec~\cite{hidasi2015session}, SASRec~\cite{kang2018self}, and BERT4Rec~\cite{sun2019bert4rec}. The covered recommendation scenarios include movie viewing, shopping, image-text information flow, and short-video browsing. Dataset statistics and coverage of application scenarios and recommender backbones are summarized in \autoref{fig:data_stat}.

All benchmark evaluations follow a unified closed-loop protocol. For each dataset and recommender setting, WebSim serves recommendation slates to the simulator, collects the simulator's feedback, and updates the interaction history stepwise. This design allows benchmark results to reflect both the quality of local actions and long-horizon interaction dynamics. At a high level, the benchmark evaluates PsyBer-Agent from four complementary perspectives: behavior fidelity, baseline simulator comparison, scalability and cost, and group behavior. Detailed metric definitions and evaluation procedures are provided in Appendix~\ref{App:metrics_eval}.

\subsection{WebSim Experiment Platform}
\label{App:websim}

\paragraph{Design goals.} We design WebSim as a controllable and reproducible platform for interactive recommendation simulation. In its current version, WebSim adopts a grid-based interaction protocol and separates three concerns: interaction serving, model-agnostic recommendation inference, and dataset-aware metadata and poster delivery. This modularization allows datasets and recommender backbones to vary without modifying the core serving pipeline.

\paragraph{Interaction state management.} WebSim maintains a lightweight per-session state in Flask sessions together with an auxiliary in-memory table for interaction statistics. The session state stores recent interaction history, cached recommendation IDs, the current page index, the selected model, the dataset identifier, and a simulation identifier, while the statistics table records item-level impressions and clicks. At initialization, the service resets the session and returns four seed-controlled sampled items. Subsequent clicks update the interaction history, refresh the recommendation cache, and return the next batch of candidates, while pagination traverses the cached list deterministically. In the current implementation, the history length is capped at 20, the page size is fixed to 4, and the recommendation cache is constructed with top-$K=200$.

\paragraph{Recommendation layer and model abstraction.} A unified \texttt{MovieRecommender} wrapper dispatches inference to up to seven backbones: PopRec, BPR-MF, GRU4Rec, BERT4Rec, SASRec, LightGCN, and Multi-VAE. For each dataset, only backbones with available checkpoints are instantiated. All models share a normalized item-ID interface and return ranked item IDs; the wrapper then filters unavailable IDs, removes duplicates, and maps the remaining candidates to display cards through \texttt{MovieCatalog}. If a requested model is unavailable, the system falls back to a default model; random candidate sampling is used only when no model can be loaded or when inference returns no valid items.

\paragraph{Data and asset handling.}~\texttt{MovieCatalog} unifies metadata loading for MovieLens-style and Amazon-style datasets in the current implementation and resolves image resources from dataset-local posters and images, as well as legacy fallback paths. When recommendation models are available, the catalog is restricted to items covered by the loaded artifacts, preventing dangling item IDs during serving. If a poster file is missing, the platform generates an SVG placeholder on the fly, ensuring stable rendering across datasets with heterogeneous media completeness.

\paragraph{Reproducibility and deployment.} WebSim supports script-based launch and shutdown, service readiness probing, and environment-configurable dataset and checkpoint paths, enabling reproducible local and batch evaluation. The service port is configurable, and the current Flask implementation binds to \texttt{127.0.0.1} by default. Because dataset paths and model artifacts are configured outside the serving logic, the same codebase can be reused across domains and model variants with minimal engineering overhead.

\noindent\textbf{System architecture}
\par
\smallskip
\noindent
\begin{minipage}[t]{0.32\linewidth}
\strut\par
WebSim adopts a lightweight client--server architecture with a browser-based client that supports click and pagination feedback on a grid page. The UML class diagram is shown in \autoref{fig:app_uml}. The backend is implemented in Flask and provides a compact serving layer for session initialization, item selection, recommendation pagination, session cleanup, poster rendering, and health monitoring. By centralizing session management and recommendation dispatch, the platform keeps the frontend lightweight while simplifying deployment and experiment orchestration.

\end{minipage}\hfill
\begin{minipage}[t]{0.65\linewidth}
\strut\par

\centering
\includegraphics[width=\linewidth]{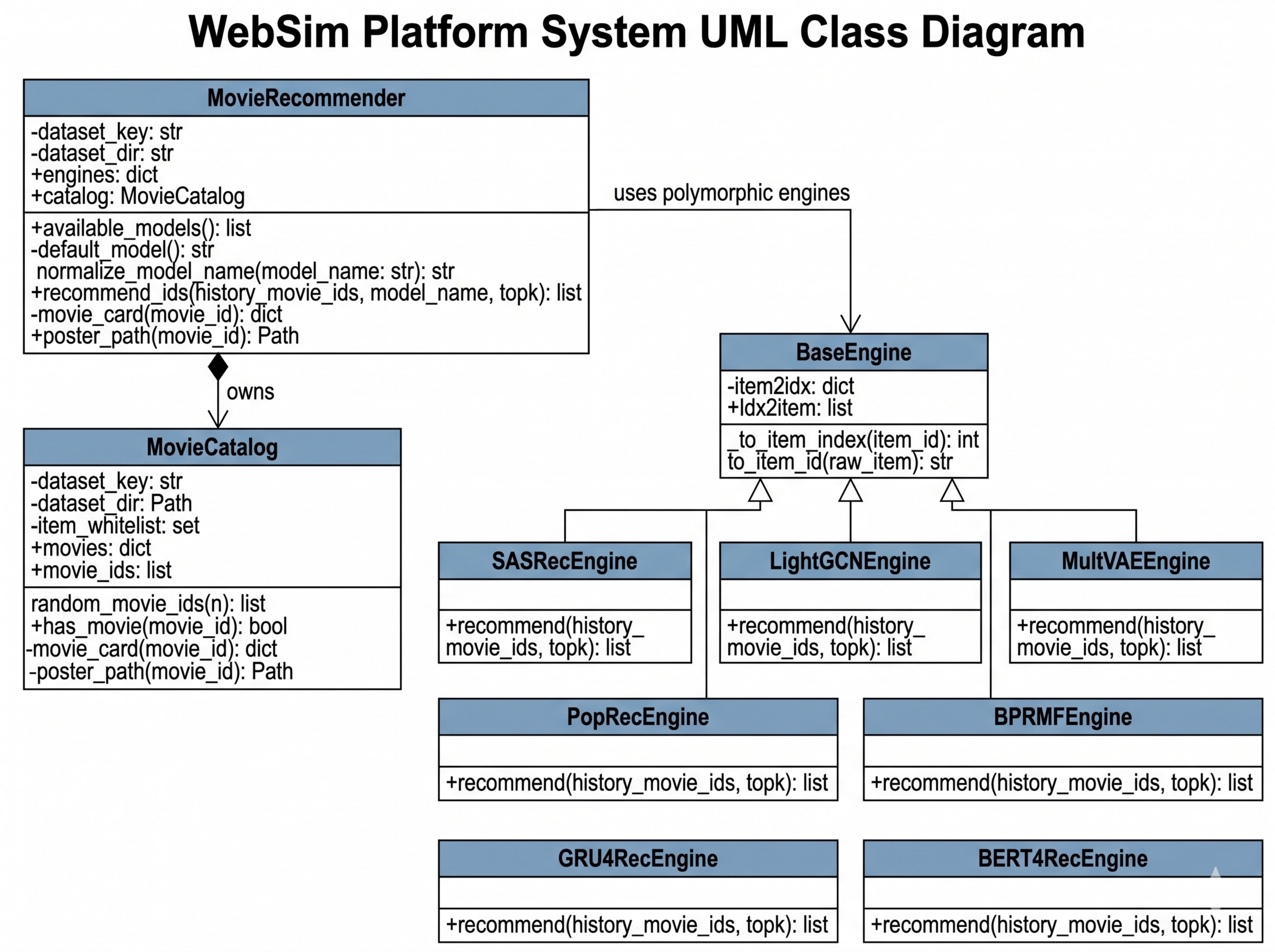}
\captionsetup{hypcap=false}\captionof{figure}{UML class diagram of the WebSim platform.}
\label{fig:app_uml}
\end{minipage}

\subsection{Metrics and Evaluation}
\label{App:metrics_eval}

We use common notation for the metrics reported across scenarios. Let $\mathcal{U}_{\mathrm{eval}}$ denote the evaluated users, $\mathcal{I}$ the item set, and $K$ the ranking cutoff used for top-$K$ evaluation. For each user $u$, $\mathcal{R}_{u,K}$ denotes the top-$K$ recommended list and $\mathcal{G}_u$ denotes the held-out relevant item set. Let $P_{\mathrm{sim}}$ and $P_{\mathrm{exp}}$ be the simulated and expert-reference item interaction distributions, respectively. For matched simulator--expert sequence pairs, let $\mathcal{M}$ denote the matching set. For each matched pair $(u,u')\in\mathcal{M}$, let $T_{u,u'}$ denote the matched sequence length, and let $a^{\mathrm{sim}}_{u,t}$ and $a^{\mathrm{exp}}_{u',t}$ denote the simulated and expert-reference actions at step $t$, respectively. We denote the model-defined C/E/S latent state at step $t$ as
$$
\mathbf{z}_{u,t}=[C_{u,t},E_{u,t},S_{u,t}]^\top\in[0,1]^3.
$$
When distinguishing simulated and expert-reference trajectories, we use $\mathbf{z}^{\mathrm{sim}}_{u,t}$ and $\mathbf{z}^{\mathrm{exp}}_{u',t}$, respectively, where $\mathbf{z}^{\mathrm{exp}}_{u',t}$ is reconstructed from offline interaction logs through Psy-Engine teacher forcing.

\paragraph{Behavior Fidelity.}
We measure whether simulated agents reproduce expert-reference behavior at both population and trajectory levels.

\begin{itemize}
    \item \textbf{Interaction Jensen--Shannon Divergence (Interaction JSD)} ($\downarrow$):
    population-level mismatch between simulated and expert-reference item interaction distributions,
    $$
    \mathrm{JSD}
    =
    \frac{1}{2}\mathrm{KL}(P_{\mathrm{sim}}\|P_{\mathrm{mix}})
    +
    \frac{1}{2}\mathrm{KL}(P_{\mathrm{exp}}\|P_{\mathrm{mix}}),
    \quad
    P_{\mathrm{mix}}=\frac{1}{2}(P_{\mathrm{sim}}+P_{\mathrm{exp}}).
    $$
    Lower Interaction JSD indicates that the simulated population better matches the expert-reference population-level interaction patterns.

    \item \textbf{Earth Mover's Distance (EMD)} ($\downarrow$):
    distributional distance between simulated and expert-reference downstream behavior trajectories after matched prefixes, where trajectories are represented by future interaction sequences or their embedded representations. Lower EMD indicates that simulated agents produce future behavioral trajectories closer to those observed from expert-reference logs.
\end{itemize}

\paragraph{Baseline Simulator Comparison.}
We compare PsyBer-Agent against baseline user simulators under a controlled fixed-candidate replay setting. At each step, all simulators observe the same user histories, item metadata, and candidate pages, and are evaluated on decision accuracy, CTR calibration, distributional fidelity, item-level correlation with real logs, and latency.

\begin{itemize}
    \item \textbf{Decision Accuracy (Acc.)} ($\uparrow$):
    step-wise action agreement between matched simulator and expert-reference trajectories,
    $$
    \mathrm{Acc}
    =
    \frac{1}{|\mathcal{M}|}
    \sum_{(u,u')\in\mathcal{M}}
    \frac{1}{T_{u,u'}}
    \sum_{t=1}^{T_{u,u'}}
    \mathbb{I}\!\left(
    a^{\mathrm{sim}}_{u,t}=a^{\mathrm{exp}}_{u',t}
    \right).
    $$

    \item \textbf{CTRGap} ($\downarrow$):absolute CTR deviation for the evaluated dataset--recommender cell, 
    $$
    \mathrm{CTRGap} = \left| \mathrm{CTR}_{\mathrm{sim}}-\mathrm{CTR}_{\mathrm{exp}} \right|.
    $$

    \item \textbf{Pearson Correlation (R)} ($\uparrow$):
    item-level correlation between simulated and expert-reference interaction frequencies,
    $$
    R = \mathrm{Corr}\!\left(
    \bigl(f_{\mathrm{sim}}(i)\bigr)_{i\in\mathcal{I}},
    \bigl(f_{\mathrm{exp}}(i)\bigr)_{i\in\mathcal{I}}
    \right),
    $$
    where $f_{\mathrm{sim}}(i)$ and $f_{\mathrm{exp}}(i)$ denote the simulated and expert-reference interaction frequencies of item $i$, respectively.

    \item \textbf{Latency/Decision} ($\downarrow$):
    average wall-clock latency per simulator decision.
\end{itemize}

\paragraph{Recommendation Model Evaluation.}
For the benchmark task of evaluating target recommendation models with PsyBer-Agent, we report standard online recommendation metrics derived from simulated interaction trajectories.

\begin{itemize}
    \item \textbf{Hit Rate@10 (HR@10)} ($\uparrow$):
    $$
    \mathrm{HR@10}=
    \frac{1}{|\mathcal{U}_{\mathrm{eval}}|}
    \sum_{u\in\mathcal{U}_{\mathrm{eval}}}
    \mathbb{I}\!\left(
    \mathcal{R}_{u,10}\cap \mathcal{G}_u \ne 
    \right).
    $$

    \item \textbf{NDCG@10} ($\uparrow$):
    $$
    \mathrm{NDCG@10} = \frac{1}{|\mathcal{U}_{\mathrm{eval}}|}
    \sum_{u\in\mathcal{U}_{\mathrm{eval}}}
    \frac{\mathrm{DCG}_u@10}{\mathrm{IDCG}_u@10},
    \quad
    \mathrm{DCG}_u@10 = \sum_{r=1}^{10}
    \frac{2^{\mathrm{rel}_{u,r}}-1}{\log_2(r+1)}.
    $$
    If $\mathrm{IDCG}_u@10=0$, we set the corresponding user-level NDCG to zero.

    \item \textbf{CTR} ($\uparrow$):
    click-through rate induced by the simulator under the evaluated recommender setting,
    $$
    \mathrm{CTR} = \frac{N_{\mathrm{click}}}{N_{\mathrm{impression}}},
    $$
    where $N_{\mathrm{click}}$ and $N_{\mathrm{impression}}$ denote the total numbers of clicks and impressions, respectively.
\end{itemize}

\paragraph{Scalability and Cost.}
\label{app:metric_sca}
We evaluate whether PsyBer-Agent remains efficient and stable under large-scale concurrent simulation. Let $N_{\mathrm{submitted}}$ denote the number of tasks placed in the queue, $N_{\mathrm{finished}}$ the number of tasks that terminate with either success or failure, and $N_{\mathrm{success}}$ the number of tasks that complete successfully. Let $T_{\mathrm{wall}}$ denote the elapsed time from queue initialization until the last worker terminates.

\begin{itemize}
    \item \textbf{Throughput} ($\uparrow$):
    aggregate terminal tasks per second across all workers,
    $$
    \mathrm{Throughput}
    =
    \frac{N_{\mathrm{finished}}}{T_{\mathrm{wall}}}.
    $$
    \item \textbf{p95 Latency/Decision} ($\downarrow$):
    the 95th percentile of per-decision latency across completed runs.
    \item \textbf{Success Rate} ($\uparrow$):
    $$
    \mathrm{SuccessRate}
    =
    \frac{N_{\mathrm{success}}}{N_{\mathrm{submitted}}}.
    $$
    \item \textbf{Total Token Cost} ($\downarrow$):
    total LLM token consumption across all decisions,
    $$
    \mathrm{TokenCost}
    =
    \sum_{q\in\mathcal{Q}} \mathrm{Tok}(q),
    $$
    where $\mathcal{Q}$ denotes all LLM calls made during the evaluation and $\mathrm{Tok}(q)$ is the total number of input and output tokens consumed by call $q$. For Table~\ref{tab:scalability_cost}, throughput is computed globally across all workers using the same wall-clock interval reported in the table.
\end{itemize}

\paragraph{Group Behavior.}
\label{app:metric_gro}
For controlled social-cue perturbations, we use the following diagnostic metrics to quantify emergent collective behavior.
\begin{itemize}
    \item \textbf{Herding Gain}:
    increase in interaction probability after positive social cues are displayed.
    \item \textbf{Comment Conformity} ($\uparrow$):
    fraction of generated feedback aligned with the displayed comment-board sentiment. This metric is used as a manipulation check for whether displayed comment-board sentiment is perceived and reflected in generated feedback. It is not used as standalone evidence of action-level herding.
    \item \textbf{Polarization Shift} ($\Delta r$):
    change in exposure-similarity assortativity induced by social-cue perturbations,
    $$
    \Delta r = r_{\mathrm{treat}} - r_{\mathrm{control}},
    $$
    where $r_{\mathrm{treat}}$ and $r_{\mathrm{control}}$ are the assortativity coefficients under perturbed and control social cues, respectively.
    \item \textbf{Contrast--Interest Correlation} ($\rho_{\mathrm{sp}}$):
    Spearman correlation between native item visual contrast and agent interest, used to quantify whether visual salience systematically affects collective interaction patterns.
\end{itemize}

\paragraph{Ablation Metrics.}
For component and information-channel ablations, we report simulation-fidelity metrics that capture behavior, exposure, and latent-state deviations.

Here, $\mathrm{CTR}_{\mathrm{sim}}$ and $\mathrm{CTR}_{\mathrm{exp}}$ denote the average click-through rates in the simulated and expert-reference environments, respectively. For Exp-JS, we compare the normalized item-exposure distributions induced by the simulator and the expert policy. For CES-MSE, $\mathcal{M}$ is the set of matched simulated--expert user pairs, $T_{u,u'}$ is the aligned interaction horizon for pair $(u,u')$, and $\mathbf{z}_{u,t}^{\mathrm{sim}}$ and $\mathbf{z}_{u',t}^{\mathrm{exp}}$ are the corresponding latent user-state vectors at step $t$; the factor $3$ averages over the three latent dimensions.

\begin{itemize}
    \item \textbf{CTRGap} ($\downarrow$):
    $$
    \mathrm{CTRGap}
    =
    \left|
    \mathrm{CTR}_{\mathrm{sim}}-\mathrm{CTR}_{\mathrm{exp}}
    \right|.
    $$
    \item \textbf{Exposure JSD (Exp-JS)} ($\downarrow$):
    Jensen--Shannon divergence between simulated and expert-reference item exposure distributions.
    \item \textbf{CES-MSE} ($\downarrow$):
    $$
    \mathrm{CES\text{-}MSE}
    =
    \frac{1}{3\sum_{(u,u')\in\mathcal{M}} T_{u,u'}}
    \sum_{(u,u')\in\mathcal{M}}
    \sum_{t=1}^{T_{u,u'}}
    \left\|
    \mathbf{z}_{u,t}^{\mathrm{sim}}-\mathbf{z}_{u',t}^{\mathrm{exp}}
    \right\|_2^2.
    $$
\end{itemize}

CES-MSE measures internal agreement between simulator-generated latent states and reference states reconstructed by replaying observed interactions through Psy-Engine. Because both trajectories use the same model-defined C/E/S space, a lower CES-MSE indicates greater internal state consistency but does not establish correspondence to independently measured human psychological states.

\paragraph{Robustness Metrics.}
For robustness analysis, we use stress-test-specific metrics for behavior, decision, and system robustness.

\begin{itemize}
    \item \textbf{Perturbation Exposure JSD} ($\downarrow$):
    Jensen--Shannon divergence between the perturbed and clean exposure distributions.
    \item \textbf{Relative Drop} ($\downarrow$):
    percentage degradation relative to the clean in-domain reference. For the long-horizon stress experiment, this metric is interpreted as a within-method degradation measure; claims of superior long-term stability require comparison with baselines evaluated at the same horizons.
    \item \textbf{Decision Agreement} ($\uparrow$):
    fraction of perturbed rollouts that preserve the same selected action as the clean condition.
    \item \textbf{Policy Shift} ($\downarrow$):
    total variation distance between clean and perturbed action distributions.
    \item \textbf{Latency/Decision} and \textbf{p95 Latency} ($\downarrow$):
    average per-decision latency and 95th percentile execution latency under perturbations.
    \item \textbf{Stress Success Rate} ($\uparrow$):
    fraction of simulation runs completed successfully under system-level perturbations.
\end{itemize}


\section{Additional Details}

\subsection{Details of the Evaluated Datasets}
\label{App:tested_datasets}

In this study, we evaluate PsyBer-Agent on a diverse set of multimodal recommendation datasets spanning movie, e-commerce, short-video, and cross-domain recommendation scenarios. The PsyBer Benchmark includes seven datasets in total: \textbf{MM-ML-1M}~\cite{zhang2026exploring}, four \textbf{Amazon Review} domains~\cite{hou2024bridging}, \textbf{MicroLens-100K}~\cite{ni2025content}, and \textbf{NineRec-Bili500K}~\cite{zhang2025ninerec}. The detailed statistics of each dataset are summarized in \autoref{tab:dataset_stats}.
\begin{table}[ht]
\centering
\caption{Statistics of datasets used in the PsyBer Benchmark.}
\label{tab:dataset_stats}
\setlength{\tabcolsep}{8pt}
\small
\begin{tabular}{lrrrrr}
\toprule
\textbf{Dataset} & \textbf{\#Users} & \textbf{\#Items} & \textbf{\#Inter.} & \textbf{Sparsity} & \textbf{Avg. Inter./User} \\
\midrule
\textbf{MM-ML-1M} & 6,040 & 3,952 & 1,000,209 & 95.81\% & 165.60 \\
\textbf{Amazon\_Magazine} & 380 & 865 & 2,679 & 99.18\% & 7.05 \\
\textbf{Amazon\_All\_Beauty} & 22,363 & 12,101 & 198,502 & 99.93\% & 8.88 \\
\textbf{Amazon\_Luxury\_Beauty} & 1,054 & 791 & 12,397 & 98.51\% & 11.76 \\
\textbf{Amazon\_Musical\_Instruments} & 25,402 & 10,611 & 210,646 & 99.92\% & 8.29 \\
\textbf{MicroLens-100K} & 100,000 & 19,738 & 719,405 & 99.96\% & 7.19 \\
\textbf{NineRec-Bili500K} & 500,000 & 138,033 & 7,845,805 & 99.99\% & 15.69 \\
\bottomrule
\end{tabular}
\end{table}

\textbf{MM-ML-1M}~\cite{zhang2026exploring} is a multimodal extension of the classic MovieLens benchmark, enriching user--item interactions with high-quality movie posters and detailed plot summaries. It supports recommendation evaluation with rich visual and textual item signals.

\textbf{Amazon Review domains}~\cite{hou2024bridging} are built from the Amazon Reviews corpus and include four representative domains: \textbf{Magazine\_Subscriptions}, \textbf{All\_Beauty}, \textbf{Luxury\_Beauty}, and \textbf{Musical\_Instruments}. Each domain is paired with multimodal product metadata, including item images and textual descriptions, providing a realistic testbed for e-commerce recommendation.

\textbf{MicroLens-100K}~\cite{ni2025content} is a large-scale short-video recommendation dataset. In addition to behavioral feedback, each item is associated with multimodal content signals, including titles, cover images, audio, and video, enabling evaluation beyond ID-based recommendation.

\textbf{NineRec-Bili500K}~\cite{zhang2025ninerec} is a large-scale dataset drawn from a cross-domain recommendation benchmark, with aligned multimodal item metadata such as descriptive text and cover images. It is used to evaluate the robustness of simulated-user behavior under transferable and cross-domain recommendation settings.

\subsection{Details of the Evaluated Recommender Systems}
\label{App:tested_recsys}

To evaluate PsyBer-Agent across diverse recommendation environments, we select seven representative recommender backbones covering five mainstream paradigms: heuristic methods, collaborative filtering, graph neural networks, deep generative models, and sequential recommendation. The details of these models are summarized in \autoref{tab:recsys_models}.

\begin{table}[htbp]
\centering
\caption{Recommender system models evaluated with PsyBer-Agent.}
\label{tab:recsys_models}
\renewcommand{\arraystretch}{1.0}
\footnotesize
\begin{tabular}{@{}p{4.5cm}p{3.8cm}p{5cm}@{}}
\toprule
\textbf{Model Name} & \textbf{Model Type} & \textbf{Core Mechanism} \\
\midrule
\textbf{PopRec}~\cite{cremonesi2010performance} & Heuristic Baseline & Item popularity ranking based on historical interaction frequency. \\
\addlinespace
\textbf{BPR-MF}~\cite{rendle2012bpr} & Collaborative Filtering & Matrix factorization optimized with pairwise ranking loss. \\
\addlinespace
\textbf{LightGCN}~\cite{he2020lightgcn} & Graph Neural Network & Simplified linear message passing on the user--item bipartite graph. \\
\addlinespace
\textbf{Multi-VAE}~\cite{liang2018variational} & Deep Generative Model & Variational autoencoder with multinomial likelihood for implicit-feedback recommendation. \\
\addlinespace
\textbf{GRU4Rec}~\cite{hidasi2015session} & Sequential Recommendation & Gated recurrent units for session-based sequence modeling. \\
\addlinespace
\textbf{SASRec}~\cite{kang2018self} & Sequential Recommendation & Unidirectional self-attention for sparse sequential recommendation. \\
\addlinespace
\textbf{BERT4Rec}~\cite{sun2019bert4rec} & Sequential Recommendation & Bidirectional self-attention trained with masked item prediction. \\
\bottomrule
\end{tabular}
\end{table}

\newpage

\subsection{Details of the User Simulator Baselines}
\label{App:sim_baseline}

\begin{wraptable}{r}{0.56\textwidth}
\centering
\small
\setlength{\tabcolsep}{3pt}
\renewcommand{\arraystretch}{1.08}

\caption{Categorization of user simulator baselines.}
\label{tab:user_simulator_baseline_categories}

\begin{tabular}{>{\raggedright\arraybackslash}p{4.5cm}
                >{\raggedright\arraybackslash}p{2.5cm}}
\toprule
\textbf{Methods} & \textbf{Category} \\
\midrule
RecAgent~\cite{10.1145/3708985} & Social simulation \\
\midrule
Agent4Rec~\cite{zhang2024generative} & Generative-agent simulation \\
\midrule
UserSim~\cite{zhang2025llm} & RL-oriented simulation \\
\midrule
DyTA4Rec~\cite{10.1145/3746252.3760878} & Temporal-aware simulation \\
\bottomrule
\end{tabular}
\end{wraptable}

We compare PsyBer-Agent with representative agent-based user simulators for recommender-system evaluation, covering social simulation, generative-agent simulation, RL-oriented evaluation, and temporal-aware modeling. RecAgent~\cite{10.1145/3708985} simulates recommendation and social behaviors with profile, memory, and action modules; Agent4Rec~\cite{zhang2024generative} constructs generative users from recommendation data with persona/profile, memory, and reasoning modules; UserSim~\cite{zhang2025llm} provides LLM-based engagement inference for RL recommender evaluation; and DyTA4Rec~\cite{10.1145/3746252.3760878} captures temporal preference dynamics. Compared with these baselines, PsyBer-Agent jointly combines explicit psychological state modeling, real-log psycho-behavioral alignment, WebSim-based interaction, and million-scale simulation.

\paragraph{Baseline inclusion criterion.}
We include a method in the quantitative comparison only when an official or
author-provided implementation, required checkpoints, and a protocol-compatible
action interface are available at the experimental freeze date. The method must
also accept the available user histories, item metadata, and fixed candidate
pages, and its outputs must be evaluable under the same metrics as PsyBer-Agent.

We considered PUB, RecInter, AlignUSER, ContextSim, RecUserSim, and CSHI as
relevant recent methods. However, at the experimental freeze date, we could
not obtain a complete executable artifact satisfying the above requirements
for these methods. Our reproducibility records document the artifact source, version or commit when available, date checked, and the specific protocol mismatch observed for each method. These observations describe the materials we could verify at the freeze date; they do not assert that the methods are unavailable, unreleased, or inferior. We therefore discuss
these methods as related work but do not report reimplementations based only
on paper descriptions, since such reimplementations would introduce
additional implementation and tuning choices. At the experimental freeze date, we excluded methods only when their artifacts or interfaces could not be verified under the fixed protocol; our per-method records include the source, version, date, and observed limitation. This exclusion reflects a
reproducibility and protocol-compatibility constraint, rather than a claim
about the relative quality of these methods.

\subsection{Details of the Prompt Design}
\label{App:prompts}

This section presents the prompt used to drive the LLM for final action selection. The prompt template is shown below:
\begin{lstlisting}[style=promptstyle]
Prompt:
    "You are an online user. Please select your preferred option based on the screenshot and click the corresponding image. You must output only JSON with no additional text.\n"
    "You will receive the following decision-making information simultaneously: current psychological state (psy_state), potential choices inferred from your psychological state (fast_policy_candidates, top-k), recent click history (recent_clicks), and available items on the current page (candidates).\n"
    "Make your final decision through comprehensive judgment.\n"
    "Available Actions:\n"
    '1) {"action":"click","item_id":"<item_id from candidates>","reason":"..."}\n'
    "Rules:\n"
    "- Only perform click actions using item_ids from the candidate list.\n"
    "- If next_enabled=true and Next is not included in fast_policy_candidates, you must select click. In addition, you cannot consecutively select Next by referring to recent_clicks.\n"
    "- If all candidates on the current page are contained in clicked_item_ids and next_enabled=true, you must select next.\n"
    "- If next_enabled=false, you must select click.\n\n"
    "Notes:\n"
    "- Your current psychological state and fast_policy_candidates (choices predicted from your psychological state) serve only as reference signals, not hard constraints.\n"
    "- In case of conflicts among multiple reference signals, select your preferred option based on the screenshot and behavioral coherence as the final judgment.\n\n"
    f"dataset={memory.selected_dataset_key}, model={memory.selected_model_key}\n"
    f"next_enabled={str(next_enabled).lower()}\n"
    f"status_text={status_text}\n"
    f"psy_state={psy_state_view}\n"
    f"psy_pareto_front={psy_pareto_front}\n"
    f"psy_candidates={psy_candidate_text}\n"
    f"fast_policy_candidates={fast_policy_text}\n"
    f"recent_clicks={recent_lines}\n"
    f"clicked_item_ids={clicked_item_text}\n"
    f"candidates={card_lines}\n"
\end{lstlisting}

\begin{figure}[!ht]
    \centering
    
    \includegraphics[width=0.8\linewidth]{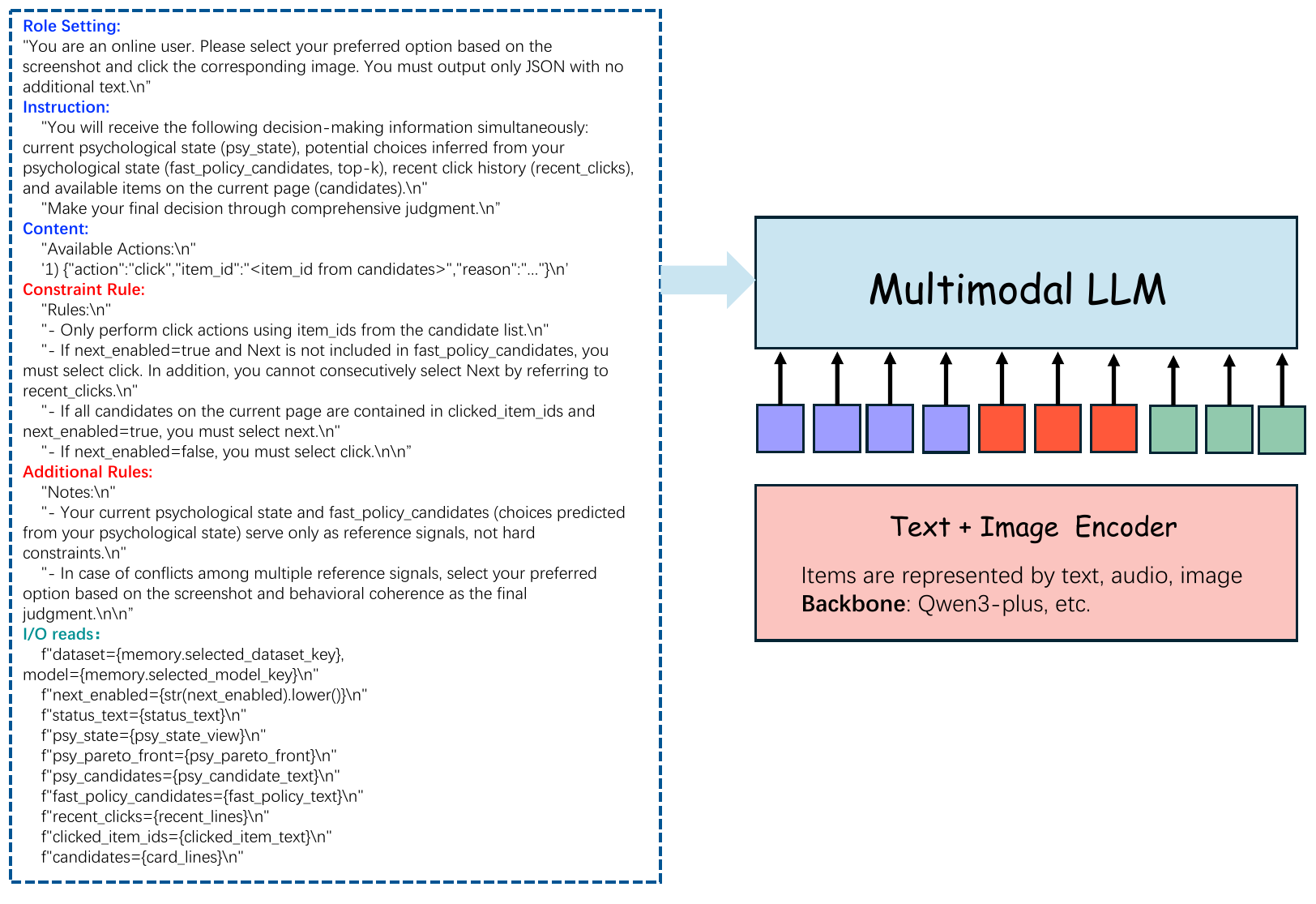}
    \caption{The schematic diagram of the specific scheme for reasoning modeling using a multimodal large model.}
    \label{App:fig_prompt}
    
\end{figure}

Here, \texttt{psy\_state} denotes the current psychological state; \texttt{psy\_pareto\_front} denotes the Pareto-optimal action subset under Psy-Engine; \texttt{psy\_candidates} provides psycho-behavioral annotations of the current-page candidates; \texttt{fast\_policy\_candidates} gives the top-$k$ suggestions from the learned fast policy; \texttt{recent\_clicks} records recent click history; and \texttt{status\_text} summarizes the current interaction context. Along with this Prompt, a live WebSim Platform Page screenshot is passed to drive the LLM. Below is the decomposition ~\autoref{App:fig_prompt} of our prompt.

\subsection{Details of Experimental Hardware Environment and Parallel Simulation Settings}
\label{App:exp_hard}

\textbf{Experimental Hardware Environment} All experiments are conducted on a dedicated server running Ubuntu 20.04.6 LTS, equipped with an Intel Xeon Gold 5218R CPU @ 2.10GHz (80 cores), 251 GB of memory, and 787G of root disk space. The graphics card is 6 $\times$ NVIDIA RTX 3090 GPUs. The inference of the LLM driving the Agent is mainly implemented by calling the vendor API.

\textbf{Large-scale Concurrent Simulation Setup} PsyBer-Agent uses a queue-based rollout launcher for large-scale simulation. Given a target number of simulated users, the launcher first creates a task queue and starts a bounded number of asynchronous workers. Each worker takes one task from the queue and runs an independent WebSim agent process, so each simulated user maintains its own persona, Psy-Engine state, browser trace, and memory file. The launcher controls the active parallelism with a maximum-concurrency parameter and additionally applies launch-rate limiting to avoid bursty process creation. For LLM/VLM-based decisions, an API-pool mechanism assigns each task an available API slot and releases it after completion, preventing all agents from competing for the same endpoint. Failed or timed-out runs are isolated at the process level, while batch-level logs and summaries are collected for throughput, success rate, timeout rate, and decision-call statistics.

\paragraph{Cost Model and Unit Cost.}
Under a fixed configuration, the resource cost approximately satisfies Cost(N,R) $\propto NR$, where $N$ is the total number of simulated tasks and $R$ is the number of interaction rounds per task, so that $NR$ denotes the total number of interaction rounds generated. In Table 5, increasing the workload from 100K to 1M and from 1M to 5M increases token consumption by approximately 10.45$\times$ and 5.09$\times$, while total execution time increases by approximately 10.19$\times$ and 5.02$\times$, respectively, indicating approximately linear rather than exponential cost growth. The unit cost is approximately \$4.45 per 1K tasks; doubling the number of interaction rounds would also approximately double the API cost. We therefore frame Table 5 as stable execution under million-scale stress rather than evidence that 5M tasks are suitable for routine evaluation. The reported token total includes offline trajectory construction, UGW alignment and policy training, browser/environment calls, and online LLM/VLM fallbacks; it therefore cannot be inferred from the small fallback percentage alone.

\paragraph{Fallback Usage.}
When the fast policy cannot make a reliable decision, the system invokes an LLM/VLM fallback; when all decision routes fail, a random fallback selects a valid action. Across large-scale runs, the LLM/VLM fallback is triggered in only $\approx$0.6\% of decisions and the random fallback in $\approx$0.04\%, so the fast policy handles the overwhelming majority of decisions and fidelity is not dominated by fallback behavior.

\subsection{Details of Fixed-Backend Human--Simulator Agreement Protocol}
\label{App:ab_test}
The fixed-backend human--simulator agreement study consists of an experimental group of 500 real human users and 500 simulated PsyBer-Agent users, each completing 10 consecutive browsing rounds. Both groups interact with the same recommendation interface and backend models in every round, with the decision-maker (human vs.\ agent) as the only independent variable. Participants were genuine anonymous users of the website, gave informed consent, and were informed about the purpose, procedure, and duration of the study. The specific sensitive information of users will also not be involved during the experiment. HR@10, NDCG@10, and CTR are computed using the same aggregation protocol for both groups. The relevance set for HR@10/NDCG@10 follows the held-out interaction target defined in Appendix C.6, and candidate slates are generated by the fixed backend and replayed identically for both decision-makers. The agreement is measured by the absolute difference between the corresponding metrics. Interaction logs are stored in anonymized form. The simulated users are initialized from the corresponding pre-window persona and history, and the 10 repeated decisions are treated as clustered within user. We use a prespecified user-level paired-bootstrap procedure. The tables report point estimates; when user-level outcomes are available, 95\% intervals are obtained from the 2.5th and 97.5th percentiles of the paired bootstrap distribution.

\section{Additional Experiments \& Analysis}
\label{app:add_exp}

\subsection{PsyBer-Agent Group Behavior Exploration.} 
\label{App:Group_be}
We probe collective dynamics in WebSim by fixing the candidate pool and base ranking while perturbing only social cues, including rating priors, hotness badges, and comment sentiment. We report herding gain, Comment Conformity (Conf.), polarization shift $\Delta r$, and contrast--interest correlation. As shown in ~\autoref{tab:group_behavior_perturb}, PsyBer-Agent reliably reflects local comment sentiment, as evidenced by high Sent. Align., but this cue uptake does not translate into strong action-level herding, polarization, or visual-salience-driven behavior.

\begin{table}[!ht]
\centering
\setlength{\tabcolsep}{4.2pt}
\caption{Group behavior under social-cue perturbations in WebSim.}
\label{tab:group_behavior_perturb}
\begin{threeparttable}
\small
\begin{tabular}{llcccc}
\toprule
\textbf{Scenario} & \textbf{Backbone} &
\textbf{Herding} &
\textbf{Conf.} &
\textbf{$\Delta r$} &
\textbf{Contrast Corr.} \\
\midrule
\multirow{4}{*}{\textbf{MM-ML-1M}}
& BPR-MF     & 0.005  & 0.996 & 0.006  & -0.005 \\
& LightGCN   & 0.005  & 0.992 & 0.009  & -0.034 \\
& Multi-VAE  & 0.016  & 0.995 & 0.007  &  0.018 \\
& SASRec     & -0.069 & 0.998 & -0.002 &  0.029 \\
\midrule
\multirow{4}{*}{\textbf{Amazon Magazine }}
& BPR-MF     & 0.002  & 1.000 & 0.012  &  0.014 \\
& LightGCN   & 0.000  & 0.000 & 0.000  &  0.000 \\
& Multi-VAE  & -0.011 & 1.000 & -0.012 & -0.038 \\
& SASRec     & 0.003  & 1.000 & 0.006  & -0.082 \\
\bottomrule
\end{tabular}
\end{threeparttable}
\end{table}
\begin{checklistbox}{Findings 4}
PsyBer-Agent responds to social cues mainly through high sentiment conformity, while herding, polarization, and visual-salience effects remain limited in most tested settings.
\end{checklistbox}

\subsection{Psy-Engine and PsyBer-Agent Parameter Sensitivity Analysis}
\label{App:Exp_Sens}

This section examines the sensitivity of PsyBer-Agent to its update and training hyperparameters. We additionally specify an aggregation-weight sensitivity analysis to assess the dependence of behavioral fidelity on the equal-weight rules used in Psy-Engine.

\paragraph{Update and Training Hyperparameters.}
We study two hyperparameters: the Psy-Engine update rate $\alpha$ and the alignment learning rate \texttt{lr}. This analysis evaluates model-defined latent-state consistency and runtime efficiency. CES-MSE is an internal consistency metric and does not establish
correspondence to independently measured human psychological states.

\textbf{Evaluation Protocol.}
We adopt a one-factor-at-a-time protocol around a fixed base configuration. For each hyperparameter, we sweep 4--5 values while keeping all other settings unchanged. Each setting is repeated five times with random seeds, and \autoref{tab:app_tab_hyp} reports the mean CES-MSE and Latency/Decision.
%

\begin{table}[ht]
\centering
\caption{Sensitivity results for $\alpha$ and \texttt{lr}. Bold entries in the Values column indicate the default settings. Results are means over five runs.}
\label{tab:app_tab_hyp}
\footnotesize
\begin{tabular}{cccc}
\toprule
\textbf{Parameter}
& \textbf{Values}
& \textbf{CES-MSE $\downarrow$}
& \textbf{Latency (s) $\downarrow$} \\
\midrule
\multirow{5}{*}{$\alpha$}
& 0.10          & 0.0156 & 4.11 \\
& 0.20          & 0.0070 & 4.08 \\
& \textbf{0.35} & 0.0023 & 4.37 \\
& 0.50          & 0.0077 & 4.03 \\
& 0.70          & 0.0121 & 4.04 \\
\midrule
\multirow{4}{*}{\texttt{lr}}
& $10^{-3}$                & 0.0038 & 4.04 \\
& $\mathbf{3\times10^{-3}}$ & 0.0027 & 4.02 \\
& $5\times10^{-3}$         & 0.0800 & 5.78 \\
& $10^{-2}$                & 0.0026 & 7.51 \\
\bottomrule
\end{tabular}
\end{table}

Within the respective sweeps in \autoref{tab:app_tab_hyp}, $\alpha=0.35$ achieves the lowest mean CES-MSE, with higher latency than the other tested values. For the learning rate, $\texttt{lr}=10^{-2}$ yields a slightly lower mean CES-MSE than $3\times10^{-3}$ but substantially higher latency. We use $\alpha=0.35$, and $\texttt{lr}=3\times10^{-3}$ as the default configuration. These results characterize internal consistency and efficiency under the evaluated settings.

\paragraph{Aggregation-Weight Sensitivity}
\label{App:Aggregation_Sens}

To examine whether behavioral fidelity depends on the precise equal-weight setting, we parameterize the attractiveness, fatigue, and affective-target functions as
\begin{equation}
\begin{aligned}
A_t &= w_A\nu_t+(1-w_A)n_t, \\
F_t &= w_Fc_t+(1-w_F)\sigma_t, \\
E_t^\star &= w_\beta\beta_t+w_\rho\rho_t+w_pp_t,
\end{aligned}
\label{eq:aggregation_weights}
\end{equation}
where $w_A,w_F\in[0,1]$, $w_\beta,w_\rho,w_p\geq0$, and $w_\beta+w_\rho+w_p=1$. These constraints preserve bounded outputs for inputs in $[0,1]$. The original formulation corresponds to $w_A=w_F=0.5$ and $w_\beta=w_\rho=w_p=1/3$.

\textbf{Prespecified Protocol and Scope.}
The analysis comprises eight prespecified configurations, listed in \autoref{tab:aggregation_configs}, with one aggregation function varied at a time. This protocol quantifies sensitivity to the tested aggregation choices; it does not claim that equal-weight aggregation is optimal or psychologically validated. The self-congruence target, state-update equations, and remaining hyperparameters are held fixed.

For each configuration, the protocol requires regenerating the training reference trajectories and retraining the regression prior and fast policy, using identical data splits, training budgets, and seed sets. This ensures that each variant is evaluated with a policy trained under its own state dynamics, rather than introducing a mismatch between training and inference.

\begin{table}[ht]
\centering
\caption{Prespecified configurations for aggregation-weight sensitivity analysis. Each variant changes one aggregation function relative to the equal-weight default.}
\label{tab:aggregation_configs}
\small
\begin{tabular}{lccccc}
\toprule
\textbf{Configuration}
& $w_A$ & $w_F$ & $w_\beta$ & $w_\rho$ & $w_p$ \\
\midrule
Default & 0.50 & 0.50 & $1/3$ & $1/3$ & $1/3$ \\
A1      & 0.25 & 0.50 & $1/3$ & $1/3$ & $1/3$ \\
A2      & 0.75 & 0.50 & $1/3$ & $1/3$ & $1/3$ \\
F1      & 0.50 & 0.25 & $1/3$ & $1/3$ & $1/3$ \\
F2      & 0.50 & 0.75 & $1/3$ & $1/3$ & $1/3$ \\
E1      & 0.50 & 0.50 & 0.60 & 0.20 & 0.20 \\
E2      & 0.50 & 0.50 & 0.20 & 0.60 & 0.20 \\
E3      & 0.50 & 0.50 & 0.20 & 0.20 & 0.60 \\
\bottomrule
\end{tabular}
\end{table}
The experiment evaluation uses MM-ML-1M dataset, with a fixed recommender backbone within each scenario and five random seeds per configuration. Behavioral fidelity will be assessed against held-out real interactions using CTRGap, Interaction JSD, and Decision Accuracy. The variant without Psy-Engine will be evaluated under the same protocol to contextualize the magnitude of weight-induced changes. The results shows in ~\autoref{tab:aggregation_results}.

\begin{table}[!ht]
\centering
\scriptsize
\caption{Aggregation-weight sensitivity results in a single scenario.}
\label{tab:aggregation_results}
\begin{tabular}{lcccc}
\toprule
\textbf{Configuration}
& \textbf{Varied term}
& \textbf{CTRGap $\downarrow$}
& \textbf{Interaction JSD $\downarrow$}
& \textbf{Decision Acc. $\uparrow$} \\
\midrule
Default & None & $0.400\pm0.012$ & $0.290\pm0.010$ & $0.280\pm0.009$ \\
A1 & $w_A=0.25$ & $0.407\pm0.011$ & $0.294\pm0.012$ & $0.277\pm0.010$ \\
A2 & $w_A=0.75$ & $0.396\pm0.014$ & $0.298\pm0.011$ & $0.278\pm0.008$ \\
F1 & $w_F=0.25$ & $0.403\pm0.013$ & $0.286\pm0.010$ & $0.281\pm0.010$ \\
F2 & $w_F=0.75$ & $0.410\pm0.012$ & $0.296\pm0.013$ & $0.275\pm0.009$ \\
E1 & $(w_\beta,w_\rho,w_p)=(0.60,0.20,0.20)$ & $0.402\pm0.011$ & $0.287\pm0.012$ & $0.285\pm0.009$ \\
E2 & $(w_\beta,w_\rho,w_p)=(0.20,0.60,0.20)$ & $0.408\pm0.014$ & $0.301\pm0.011$ & $0.276\pm0.011$ \\
E3 & $(w_\beta,w_\rho,w_p)=(0.20,0.20,0.60)$ & $0.394\pm0.013$ & $0.305\pm0.014$ & $0.273\pm0.010$ \\
\midrule
w/o Psy-Engine & Reference baseline & $0.450\pm0.016$ & $0.340\pm0.015$ & $0.245\pm0.012$ \\
\bottomrule
\end{tabular}

\end{table}

Table~\ref{tab:aggregation_results} shows limited sensitivity to the tested aggregation weights, with maximum absolute changes from the default of 0.010 in CTRGap, 0.015 in Interaction JSD, and 0.007 in Decision Accuracy. No configuration performs best across all metrics: E1 achieves the highest accuracy, F1 the lowest JSD, and E3 the lowest CTRGap but a higher JSD, indicating trade-offs among behavioral fidelity measures. All weighting configurations outperform the variant without Psy-Engine on the reported mean metrics; the default reduces CTRGap and JSD by 0.050 each and improves accuracy by 0.035.

Overall, the differences associated with removing Psy-Engine exceed those induced by weight adjustments, suggesting that retaining the engine matters more than the precise weighting choice under the tested setting. These results support equal weighting as a simple and effective default, without establishing its optimality or the psychological validity of the latent states.

\subsection{PsyBer-Agent Robustness Stress Analysis}
\label{App:Exp_Rob}

\begin{figure}[ht]
    \centering
    \includegraphics[width=1.0\linewidth]{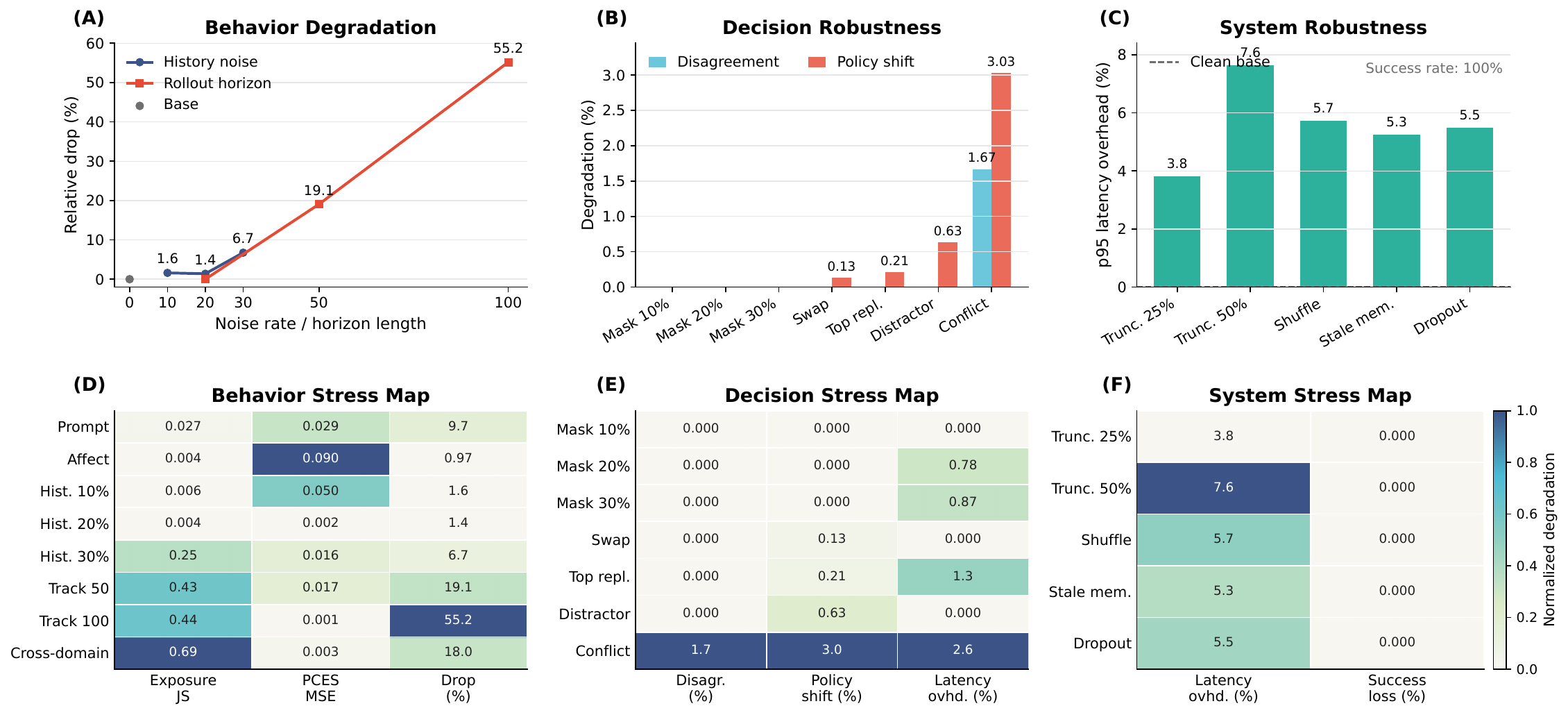}
    \caption{Robustness evaluation of PsyBer-Agent. \textbf{(A)} behavior degradation, \textbf{(B)} decision robustness, and \textbf{(C)} system robustness, together with the corresponding normalized stress maps for \textbf{(D)} behavior, \textbf{(E)} decision, and \textbf{(F)} system robustness, where darker colors indicate larger degradation.}
    \label{app:rob_fig}
\end{figure}

Robustness is a prerequisite for a user agent to serve as a reliable simulation proxy. We therefore stress-test \textsc{PsyBer-Agent} from three complementary perspectives: behavior robustness, decision robustness, and system robustness. For all robustness experiments, we define a clean base condition and compare each perturbation against it using the same evaluation protocol. We report metrics tailored to each stress test, including exposure-distribution deviation, CES sequence consistency, decision agreement, policy shift, success rate, and latency. See \autoref{app:rob_fig} for the robustness results visualization.

\textbf{Robustness Evaluation Protocol:} For behavior robustness, we report \texttt{Exposure JSD} and \texttt{CES-MSE} as absolute fidelity metrics against expert-reference logs, and \texttt{Relative Drop} as the degradation of the primary in-domain evaluation score with respect to the clean short-horizon reference. For decision robustness, we use \texttt{Decision Agreement}, \texttt{Policy Shift}, and \texttt{Latency/Decision} to quantify action stability, distributional drift, and decision efficiency. For system robustness, we report \texttt{Success Rate} and \texttt{p95 Latency} to measure execution reliability and latency under memory perturbations.

\textbf{Behavior robustness.} We evaluate whether PsyBer-Agent preserves stable and interpretable behavior under contextual perturbations and environment shift. For user-profile consistency, we fix the same underlying profile and perturb the interaction context with prompt paraphrases, transient affective shifts, and noisy history observations. We quantify robustness using Exposure JSD and CES-MSE, which measure absolute exposure-distribution deviation and psychological-sequence deviation relative to expert-reference logs. For long-range and cross-domain stability, we vary the rollout horizon (track $=20/50/100$) and evaluate transfer to an unseen target domain. We report both absolute fidelity metrics and Relative Drop (\%) with respect to the in-domain short-horizon clean reference.

\begin{table}[ht]
\caption{Behavior robustness of PsyBer-Agent under contextual perturbations and domain shifts.}
\label{tab:behavior_robustness}
\centering
\small
\begin{tabular}{llccc}
\toprule
\textbf{Setting} & \textbf{Severity}
& \textbf{Exposure JSD $\downarrow$} & \textbf{CES-MSE $\downarrow$} & \textbf{Relative Drop (\%) $\downarrow$} \\
\midrule
\textbf{Base}               & \textbf{clean} & 0.5950 & 0.0636 & -- \\
\midrule
\textbf{Perturbation}      & mild          & 0.0266 & 0.0294 & 9.6608 \\
\textbf{Affective shift}   & mild          & 0.0035 & 0.0900 & 0.9705 \\
\midrule
\multirow{3}{*}{\textbf{History noise}}
                           & 10\%          & 0.0058 & 0.0500 & 1.6055 \\
                           & 20\%          & 0.0035 & 0.0016 & 1.3801 \\
                           & 30\%          & 0.2542 & 0.0157 & 6.7454 \\
\midrule
\multirow{2}{*}{\textbf{Long horizon}}
                           & track=50      & 0.4279 & 0.0170 & 19.0528 \\
                           & track=100     & 0.4363 & 0.0014 & 55.1798 \\
\midrule
\textbf{Cross-domain}      & target domain & 0.6931 & 0.0032 & 18.0066 \\
\bottomrule
\end{tabular}
\end{table}

As shown in \autoref{tab:behavior_robustness}, PsyBer-Agent remains robust under local in-domain perturbations but degrades under longer horizons and domain shift. Because Exposure JSD and CES-MSE are measured against expert-reference logs rather than against the clean rollout, mild perturbations can occasionally improve one absolute fidelity metric; Relative Drop therefore serves as the primary summary of robustness degradation. Under this criterion, mild perturbation and affective shift yield limited degradation (Relative Drop $=9.66\%$ and $0.97\%$), and history noise remains stable at $10\%$--$20\%$ before a clearer decline at $30\%$ noise. In contrast, long-range rollout shows substantial accumulation effects, where Relative Drop increases from $19.05\%$ (track$=50$) to $55.18\%$ (track$=100$), and cross-domain transfer exhibits the largest exposure mismatch (Exposure JSD $=0.6931$). These results characterize PsyBer-Agent’s degradation profile under the tested horizons; because comparable long-horizon curves for all baselines were not available under the same protocol, they do not establish superiority in long-term stability.

\textbf{Decision robustness.} We evaluate decision robustness by testing whether the decision module preserves stable choices under imperfect observations and mild perturbations to the recommendation slate. To assess noise tolerance, we randomly mask click/history signals at rates of 10\%, 20\%, and 30\% and compare the resulting decisions with those under the clean setting. To assess sensitivity to ranking changes, we introduce small but structured slate perturbations, including adjacent swaps, top-rank replacements, distractor insertion, and conflicting signals.

\begin{table}[ht]
\caption{Decision robustness of PsyBer-Agent under noisy observations and ranking perturbations.}
\label{tab:decision_robustness}
\centering
\small
\begin{tabular}{lcccc}
\toprule
\textbf{Setting} & \textbf{Severity}
& \textbf{Decision Agreement $\uparrow$} & \textbf{Policy Shift $\downarrow$} & \textbf{Latency/Decision(s) $\downarrow$} \\
\midrule
\textbf{Base} & \textbf{Clean} & 1.0000 & 0.0000 & 4.3935 \\
\midrule
\multirow{3}{*}{\textbf{Click/history mask}} & 10\% & 1.0000 & 0.0000 & 4.3566 \\
                                             & 20\% & 1.0000 & 0.0000 & 4.4279 \\
                                             & 30\% & 1.0000 & 0.0000 & 4.4319 \\
\midrule
\textbf{Adjacent swap} & 1 pair & 1.0000 & 0.0013 & 4.2522 \\
\textbf{Top-rank replacement} & top-1 & 1.0000 & 0.0021 & 4.4487 \\
\textbf{Distractor insertion} & 1 item & 1.0000 & 0.0063 & 4.3444 \\
\textbf{Conflicting signals} & mixed & 0.9833 & 0.0303 & 4.5094 \\
\bottomrule
\end{tabular}
\end{table}

Because this experiment targets decision stability rather than oracle-based A/B correctness, we report Decision Agreement, Policy Shift (measured as total variation distance between action distributions), and Latency/Decision. As shown in \autoref{tab:decision_robustness}, click/history masking and mild ranking perturbations yield nearly identical decisions with negligible policy drift, indicating that light observation loss does not alter the top action. Conflicting signals produce the largest degradation in both agreement and policy shift, showing that PsyBer-Agent remains stable under modest perturbations while remaining sensitive to stronger contradictory evidence.

\textbf{System robustness.} Finally, we stress-test the full system under internal-state corruption. For memory perturbation, we modify the memory module through recent-window truncation, temporal shuffling, stale-memory injection, and random memory dropout, and then evaluate whether the agent can still maintain stable execution. We report Success Rate and p95 Latency to measure execution reliability and efficiency under memory noise.

\begin{table*}[ht]
\caption{System robustness of PsyBer-Agent under memory perturbations.}
\label{tab:system_robustness}
\centering
\small
\begin{tabular}{lccc}
\toprule
\textbf{Setting} & \textbf{Severity}
& \textbf{Success Rate $\uparrow$} & \textbf{p95 Latency (s) $\downarrow$} \\
\midrule
\textbf{Base Agent} & \textbf{clean} & 1.00 & 4.19 \\
\midrule
\multirow{2}{*}{\textbf{Memory truncation}} & 25\% & 1.00 & 4.35 \\
                                            & 50\% & 1.00 & 4.51 \\
\midrule
\textbf{Memory shuffle} & full & 1.00 & 4.43 \\
\textbf{Stale memory injection} & full & 1.00 & 4.41 \\
\textbf{Memory dropout} & 30\% & 1.00 & 4.42 \\
\bottomrule
\end{tabular}
\end{table*}

\autoref{tab:system_robustness} shows that PsyBer-Agent remains stable under memory perturbations: all settings maintain a perfect success rate (1.00), and the latency overhead is modest (4.35s--4.51s vs.\ 4.19s in the clean setting; worst case: +7.6\%). Overall, the results suggest graceful degradation rather than brittle failure, supporting PsyBer-Agent as a reliable simulation proxy under realistic memory noise.

\subsection{Representative Benchmark Performance of Recommendation Models on the Test Platform}
\label{App:Res_Per}

\autoref{tab:rec-benchmark} reports the performance of the recommendation models used in the backend of the WebSim platform on a representative subset of three datasets: MovieLens-1M, Amazon All Beauty, and Amazon Magazine Subscriptions. These three datasets cover dense movie recommendation, sparse beauty recommendation, and a small-catalog shopping scenario, and are the settings most frequently used in the closed-loop experiments. The table should therefore be viewed as a backend sanity check rather than the primary benchmark result.

\begin{table}[ht]
\centering
\small
\caption{Representative benchmark performance of base recommendation models on MovieLens-1M, Amazon All Beauty, and Amazon Magazine Subscriptions.}
\label{tab:rec-benchmark}
\begin{tabular}{lcccccc}
\toprule
\multirow{2}{*}{\textbf{Models}} &
\multicolumn{2}{c}{\textbf{MovieLens-1M}} &
\multicolumn{2}{c}{\textbf{All Beauty}} &
\multicolumn{2}{c}{\textbf{Magazine Subscriptions}} \\
\cmidrule(lr){2-3} \cmidrule(lr){4-5} \cmidrule(lr){6-7}
& \textbf{HR@10} & \textbf{NDCG@10} & \textbf{HR@10} & \textbf{NDCG@10} & \textbf{HR@10} & \textbf{NDCG@10} \\
\midrule
\textbf{SASRec}    & 0.3305 & 0.3170 & 0.4854 & 0.3219 & 0.2782 & 0.1102 \\
\textbf{LightGCN}  & 0.0354 & 0.0176 & 0.4768 & 0.3782 & 0.2380 & 0.1166 \\
\textbf{Multi-VAE} & 0.0626 & 0.0299 & 0.5430 & 0.3726 & 0.2493 & 0.1182 \\
\textbf{PopRec}    & 0.0368 & 0.0179 & 0.7736 & 0.5348 & 0.1406 & 0.0594 \\
\textbf{BPR-MF}    & 0.0402 & 0.0188 & 0.2781 & 0.2363 & 0.1530 & 0.0946 \\
\textbf{GRU4Rec}   & 0.1328 & 0.0656 & 0.5364 & 0.3831 & 0.1303 & 0.0578 \\
\textbf{BERT4Rec}  & 0.2232 & 0.1159 & 0.5232 & 0.4767 & 0.2238 & 0.1217 \\
\bottomrule
\end{tabular}
\end{table}

The near-ceiling results on some sparse shopping subsets indicate that these settings are comparatively easy for strong sequential recommenders and mainly serve to verify the backend recommendation layer used by WebSim.

\subsection{Profile-Conditioned Group Fidelity Diagnostics}
\label{App:Group_Fid}

We further examine whether PsyBer-Agent produces structured behavioral variation across heterogeneous simulator-side user profiles. This analysis is intended as a diagnostic study rather than a strict real-user group benchmark: the offline logs provide aggregate user--item interactions but do not contain matched annotations for personality, consumption value, or identity-gap groups. Therefore, each simulated group is compared against the overall real interaction distribution within the same scenario. The goal is to test whether profile-conditioned agents remain anchored to real aggregate behavior while exhibiting interpretable group-level differences.

Each agent's memory contains a structured user profile with three components: personality type, consumption-value type, and latent real-self/ideal-self preference embeddings. We construct three profile groupings. The \texttt{Personality} grouping contains social-driven, novelty-seeking, risk-averse, and goal-oriented users. The \texttt{Consumption Value} grouping contains brand-loyal, value-for-money, experience-first, pragmatic, trend-sensitive, and status-oriented users. The \texttt{Identity Gap} grouping is computed as
$$
g_u = 1 - \cos(\mathbf{u}^{\mathrm{real}}_u, \mathbf{u}^{\mathrm{ideal}}_u),
$$
where $\mathbf{u}^{\mathrm{real}}_u$ and $\mathbf{u}^{\mathrm{ideal}}_u$ denote the agent's real-self and ideal-self preference embeddings. We split agents into low-, medium-, and high-gap groups using scenario-specific tertiles. For each group, we aggregate simulated click sequences and compare the resulting item distribution with the real aggregate interaction distribution using JSD, Pearson correlation ($R$), and EMD.
\begin{table}[ht]
\caption{Profile-conditioned group fidelity diagnostics across user-profile groups.}
\label{tab:profile_group_fidelity}
\centering
\footnotesize
\setlength{\tabcolsep}{3pt}
\renewcommand{\arraystretch}{1.05}
\begin{tabular}{l|l|l|rrrccc}
\toprule
\textbf{Scenario} & \textbf{Group Type} & \textbf{User Group}
& \textbf{Runs} & \textbf{Clicks} & \textbf{Unique}
& \textbf{JSD$\downarrow$} & \textbf{$R\uparrow$} & \textbf{EMD$\downarrow$} \\
\midrule
\multirow{13}{*}{\textbf{MM-ML-1M}}
& \multirow{4}{*}{Personality}
& social-driven & 188 & 3741 & 1946 & 0.3310 & 0.2789 & 0.2776 \\
& & novelty-seeking & 170 & 3400 & 1682 & 0.3712 & 0.2485 & 0.2802 \\
& & risk-averse & 178 & 3534 & 1849 & 0.3393 & 0.2738 & 0.2579 \\
& & goal-oriented & 207 & 4150 & 2051 & 0.3319 & 0.2286 & 0.2807 \\
\cmidrule(lr){2-9}
\addlinespace[2pt]
& \multirow{6}{*}{\makecell[l]{Consumption\\ Value}}
& brand-loyal & 222 & 4422 & 2195 & 0.3127 & 0.3306 & 0.2774 \\
& & value-for-money & 244 & 4880 & 2280 & 0.3042 & 0.3347 & 0.2607 \\
& & experience-first & 200 & 3981 & 2105 & 0.3266 & 0.1419 & 0.2976 \\
& & pragmatic & 202 & 4050 & 1834 & 0.3503 & 0.2749 & 0.2550 \\
& & trend-sensitive & 177 & 3539 & 1650 & 0.3828 & 0.2233 & 0.2759 \\
& & status-oriented & 163 & 3253 & 1622 & 0.3671 & 0.2670 & 0.2611 \\
\cmidrule(lr){2-9}
\addlinespace[2pt]
& \multirow{3}{*}{Identity Gap}
& low & 403 & 8016 & 2939 & 0.2737 & 0.2893 & 0.2675 \\
& & medium & 405 & 8110 & 2910 & 0.2762 & 0.2771 & 0.2787 \\
& & high & 400 & 7999 & 2853 & 0.2861 & 0.2602 & 0.2657 \\
\midrule
\multirow{13}{*}{\makecell[l]{\textbf{Amazon Magazine}\\\textbf{Subscriptions}}}
& \multirow{4}{*}{Personality}
& social-driven & 193 & 3828 & 392 & 0.2076 & 0.3429 & 0.1976 \\
& & novelty-seeking & 156 & 3093 & 392 & 0.2153 & 0.3160 & 0.1954 \\
& & risk-averse & 152 & 3020 & 392 & 0.2137 & 0.3028 & 0.1947 \\
& & goal-oriented & 210 & 4146 & 392 & 0.2093 & 0.3299 & 0.1949 \\
\cmidrule(lr){2-9}
\addlinespace[2pt]
& \multirow{6}{*}{\makecell[l]{Consumption\\ Value}}
& brand-loyal & 257 & 5108 & 392 & 0.2058 & 0.3414 & 0.1928 \\
& & value-for-money & 234 & 4612 & 463 & 0.2083 & 0.3558 & 0.1898 \\
& & experience-first & 210 & 4178 & 392 & 0.2076 & 0.3463 & 0.1931 \\
& & pragmatic & 187 & 3685 & 392 & 0.2180 & 0.2915 & 0.2031 \\
& & trend-sensitive & 181 & 3610 & 392 & 0.2139 & 0.3132 & 0.1984 \\
& & status-oriented & 155 & 3058 & 392 & 0.2090 & 0.3305 & 0.1919 \\
\cmidrule(lr){2-9}
\addlinespace[2pt]
& \multirow{3}{*}{Identity Gap}
& low & 409 & 8116 & 392 & 0.2038 & 0.3518 & 0.1942 \\
& & medium & 408 & 8059 & 392 & 0.2081 & 0.3288 & 0.1994 \\
& & high & 407 & 8076 & 463 & 0.2052 & 0.3658 & 0.1842 \\
\bottomrule
\end{tabular}

\end{table}

As shown in \autoref{tab:profile_group_fidelity}, PsyBer-Agent exhibits non-trivial profile-conditioned variation while remaining close to the aggregate real interaction distribution. Across nearly all profile groups, Amazon Magazine Subscriptions obtains lower JSD and EMD than MM-ML-1M, indicating that behavior in the narrower e-commerce domain is easier to align with real aggregate interaction patterns. In contrast, MM-ML-1M shows larger variation across profile groups, consistent with the higher semantic diversity and broader preference space of movie consumption.

The profile-group results provide additional diagnostic signals. Across personality groups, socially driven users achieve strong or near-strong alignment in both scenarios, whereas novelty-seeking users tend to have higher JSD, suggesting that exploratory behavior induces greater long-tail exposure. For consumption-value groups, value-for-money users show the strongest alignment on MM-ML-1M, with the lowest JSD and highest Pearson $R$, whereas trend-sensitive and status-oriented users are harder to match. On Amazon Magazine Subscriptions, value-for-money and brand-loyal users remain well aligned, while pragmatic users show the largest distributional gap. For identity-gap groups, the three tertiles remain relatively stable, but low-gap users achieve the best JSD on MM-ML-1M, suggesting that agents whose real-self and ideal-self preference embeddings are closer generate more predictable behavior.

Overall, this diagnostic indicates that PsyBer-Agent does not collapse different personas into a single average behavior pattern. Instead, it preserves structured profile-conditioned differences while maintaining reasonable agreement with real aggregate logs. Since the offline data lack matched real-user profile labels, these results should be interpreted as evidence of simulator-side profile consistency and aggregate-log anchoring, rather than direct proof of real-world persona-level fidelity.

The following diagnostic studies complement the benchmark and use the same data processing, candidate-generation protocol, and metric definitions unless otherwise specified.

\subsection{Offline Backbone Sanity Check}

Using the publicly released WebSim checkpoints and MovieLens-1M logs, we sampled 500 users with at least 20 pre-window interactions (seed 42), held out each user's last 10 chronological interactions, and evaluated the three available backbones. The intervals below are user-level percentile bootstrap 95\% intervals with 2,000 resamples. This is an offline backend sanity check; it does not provide human outcomes or human--simulator ranking validation.

\begin{table}[ht]
\centering
\caption{Offline MovieLens-1M backbone sanity check on 500 users. Values are means with user-level 95\% bootstrap intervals.}
\label{tab:offline_backbone_sanity}
\scriptsize
\setlength{\tabcolsep}{3pt}
\begin{tabular}{lcccccc}
\toprule
Backbone & HR@10 & 95\% CI & NDCG@10 & 95\% CI & Recall@10 & 95\% CI\\
\midrule
BPR-MF & 0.2900 & [0.2350, 0.3501] & 0.2101 & [0.1531, 0.2731] & 0.0465 & [0.0350, 0.0600]\\
LightGCN & 0.2600 & [0.2050, 0.3200] & 0.1874 & [0.1393, 0.2431] & 0.0400 & [0.0295, 0.0515]\\
SASRec & 0.2550 & [0.1950, 0.3150] & 0.1767 & [0.1306, 0.2264] & 0.0340 & [0.0255, 0.0430]\\
\bottomrule
\end{tabular}
\end{table}

On this offline split, BPR-MF has the highest point estimates, followed by LightGCN and SASRec. We use this ordering only to verify the backend and held-out evaluation pipeline; it is not evidence of online recommender ranking validity.

\subsection{Aggregate Baseline Comparison}
\label{App:Aggregate_Baseline}

The main-text baseline comparison in Table~\ref{tab:baseline_simulator_basic} is the primary quantitative result. It reports results for 500 users over 10 interaction steps per user, averaged across five random seeds. This appendix does not provide separate estimates for individual backbones; the main-text values should not be interpreted as backbone-specific measurements.

\subsection{Fallback Mechanisms and Usage Statistics}
\label{app:subsec:fallback}

PsyBer-Agent uses a hierarchical decision pipeline in which a distilled fast policy handles most decision steps. When the fast policy cannot produce a valid decision, the system invokes an LLM/VLM fallback; if all decision routes fail, a valid action is selected by random fallback. These mechanisms prevent the closed-loop simulation from stalling.

We monitor their activation rates as an operational diagnostic. Table~\ref{tab:fallback_usage} reports the rates over all decision steps in the evaluation workload.
\begin{table}[htbp]
\centering
\caption{Fallback activation frequency over all decision steps.}
\label{tab:fallback_usage}
\small
\begin{tabular}{lcc}
\toprule
\textbf{Method}
& \textbf{LLM/VLM Fallback} $\downarrow$
& \textbf{Random Fallback} $\downarrow$ \\
\midrule
PsyBer-Agent & $\approx 0.6\%$ & $\approx 0.04\%$ \\
\bottomrule
\end{tabular}

\end{table}

The low activation rates indicate that most observed decisions are handled by the fast policy, while random fallback contributes only a small fraction of actions in the reported workload.

\subsection{Large-Scale Cost and Execution Analysis}
\label{App:Rebuttal_Cost}

The large-scale evaluation uses a queue-based launcher with bounded asynchronous workers. Under a fixed configuration, the resource cost approximately follows

\begin{equation}
    \mathrm{Cost}(N,R)\propto NR,
\end{equation}

where $N$ is the total number of queued simulation tasks and $R$ is the number of interaction rounds per task.

Increasing the workload from 100K to 1M tasks increases token consumption and execution time by approximately $10.45\times$ and $10.19\times$, respectively. Increasing the workload from 1M to 5M tasks increases token consumption and execution time by approximately $5.09\times$ and $5.02\times$, respectively. Under the reported configuration, the estimated API cost is approximately \$4.45 per 1K tasks.

The experiments are conducted on a server equipped with an Intel Xeon Gold 5218R CPU with 80 cores, 251 GB RAM, 787 GB disk space, and six NVIDIA RTX 3090 GPUs. The 5M-task setting is therefore a stress test rather than a recommended routine evaluation size; practical deployment should select the workload and rollout length according to the required fidelity--cost trade-off. Each workload scale is evaluated over three runs. The 5M setting denotes the total number of queued tasks rather than 5M simultaneously active tasks. Each task corresponds to a complete WebSim trajectory with $R$ interaction rounds.

\begin{table}[ht]
\centering
\caption{Cost and execution statistics supplemented in the rebuttal.}
\label{tab:app_rebuttal_cost}
\small
\begin{tabular}{ll}
\toprule
\textbf{Quantity} & \textbf{Value} \\
\midrule
Cost model & $\mathrm{Cost}(N,R)\propto NR$ \\
Token increase, 100K $\rightarrow$ 1M & $\approx 10.45\times$ \\
Token increase, 1M $\rightarrow$ 5M & $\approx 5.09\times$ \\
Time increase, 100K $\rightarrow$ 1M & $\approx 10.19\times$ \\
Time increase, 1M $\rightarrow$ 5M & $\approx 5.02\times$ \\
Estimated API cost & $\approx \$4.45$ per 1K tasks \\
Runs per workload scale & 3 \\
CPU & Intel Xeon Gold 5218R, 80 cores \\
Memory & 251 GB \\
Disk & 787 GB \\
GPUs & 6 $\times$ NVIDIA RTX 3090 \\
\bottomrule
\end{tabular}
\end{table}

These measurements characterize execution scaling under the reported configuration and are intended as a stress-test analysis rather than a claim of universal cost-effectiveness.

\paragraph{Alignment Ablation Protocol.}

To evaluate the contribution of trajectory alignment and prior-guided initialization, we compare four variants under a matched experimental protocol: direct behavioral cloning without trajectory alignment, deterministic relative-position matching, UGW with uniform initialization, and the proposed prior-guided UGW. For the direct behavioral cloning and relative-position variants, the expert position assigned to simulator step $j$ is determined by
\begin{equation}
    r(j) = 1+\operatorname{round}\left(\frac{(j-1)(n-1)}{m-1}\right).
\end{equation}

If the corresponding expert item is absent from the current candidate slate, its target action is mapped to \texttt{skip}. For the UGW variants, all methods use the same state representation, expert-reference trajectories, candidate replay, policy architecture, cross-entropy objective, optimization budget, data split, and random-seed set. The only differences are the trajectory-matching procedure and the initialization of the transport matrix. We evaluate all variants on identical test rollouts using Decision Accuracy, CTRGap, Interaction JSD, and Latency/Decision. CES-MSE is reported as an auxiliary measure of consistency in the model-defined C/E/S state space. The comparison between UGW with uniform initialization and prior-guided UGW isolates the contribution of the regression-based initialization prior.

\begin{table}[t]
\centering
\caption{Alignment ablation results.}
\label{tab:alignment_ablation}
\small
\setlength{\tabcolsep}{3pt}
\begin{tabularx}{\columnwidth}{@{}l X c c c@{}}
\toprule
Variant
& Alignment
& Acc. $\uparrow$
& CTRGap $\downarrow$
& Lat./Dec. $\downarrow$ \\
\midrule
Direct BC
& None
& 0.242 & 0.458 & 2.48 \\

Relative-pos.
& Deterministic matching
& 0.253 & 0.441 & 2.51 \\

UGW-Uniform
& Uniform initialization
& 0.267 & 0.418 & 2.51 \\

Prior-guided UGW
& Regression prior
& 0.278 & 0.406 & 2.53 \\
\bottomrule
\end{tabularx}
\end{table}

According to the experimental results, prior guided UGW has the best effect: accuracy 0.278 and CTRGap 0.406. Compared with relative position matching, uniform initialization improves the accuracy by 1.4pp and decreases the CTRGap by 0.023, which reflects the advantages of structure-aware trajectory alignment. With the introduction of prior guidance, the accuracy is further +1.1pp and CTRGap -0.012. The latency of each variant is similar (2.48 -- 2.53s/ decision), which is consistent with offline alignment.

\newpage

\section{PsyBer-Agent Case Study}
\label{App:case}

\subsection{MovieLens-based Web Browsing Scenario}
\begin{figure}[ht]
    \centering
    \includegraphics[width=1.0\linewidth]{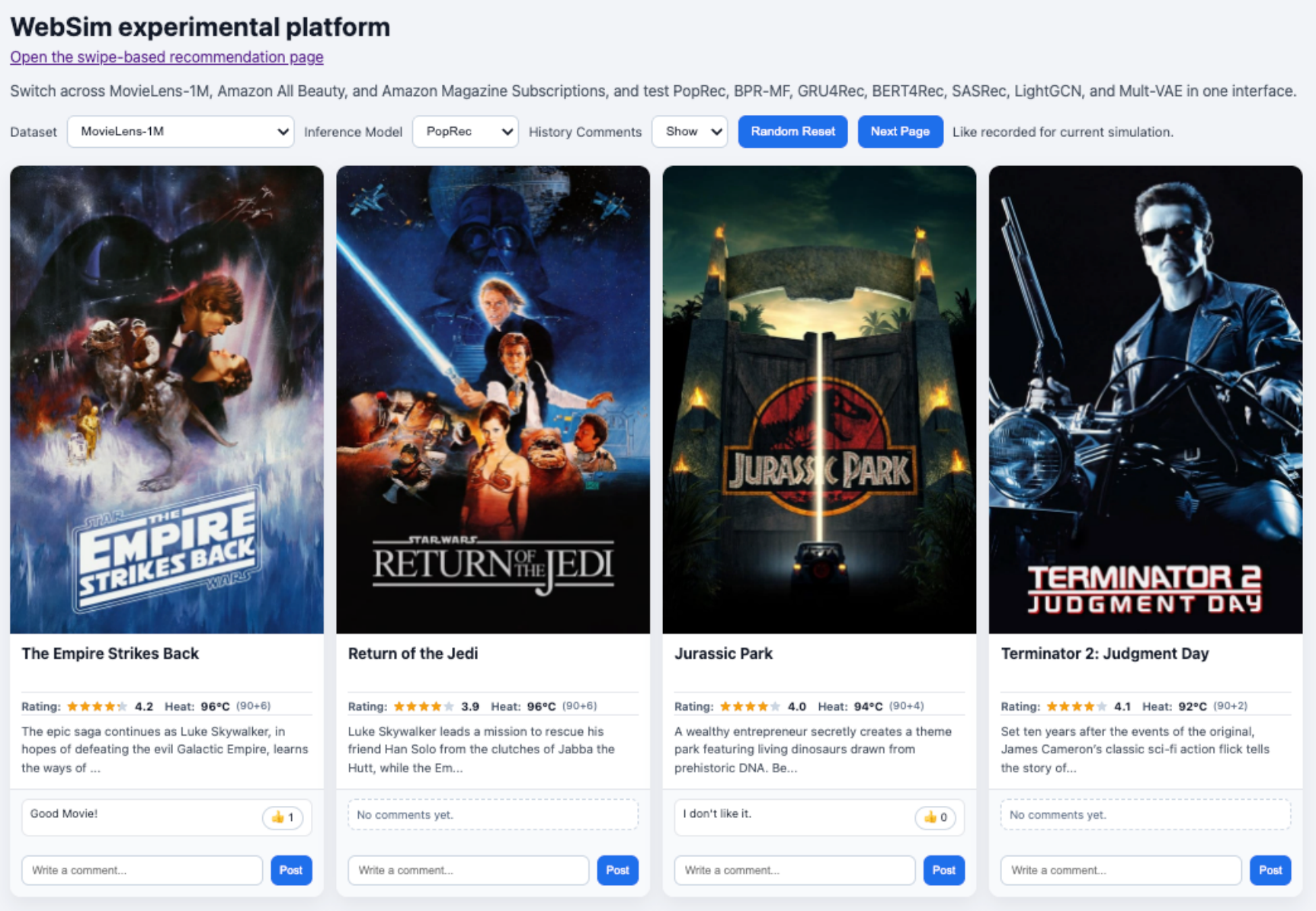}
    \caption{WebSim platform for the multimodal MovieLens browsing interface. Each movie item consists of a cover, title, rating, and popularity score. The task can switch the dataset and the recommender under test, and the agent may navigate to the next page through psychology-informed decision-making.}
    \label{fig:app_ml1m_sim}
\end{figure}

We instantiate WebSim on the Multi-Modal MovieLens dataset as a movie-browsing interface for agent evaluation. As shown in~\autoref{fig:app_ml1m_sim}, the default page presents recommendation items in a one-row, four-column grid, while two drop-down menus allow switching the dataset and recommender under test. In addition to item clicking, the interface supports random reset and next-page navigation. In this case study, we use LightGCN as the driving recommendation system and MM-ML-1M as the dataset.

\begin{figure}[ht]
    \centering
    \begin{subfigure}[t]{0.49\linewidth}
        \centering
        \includegraphics[width=\linewidth]{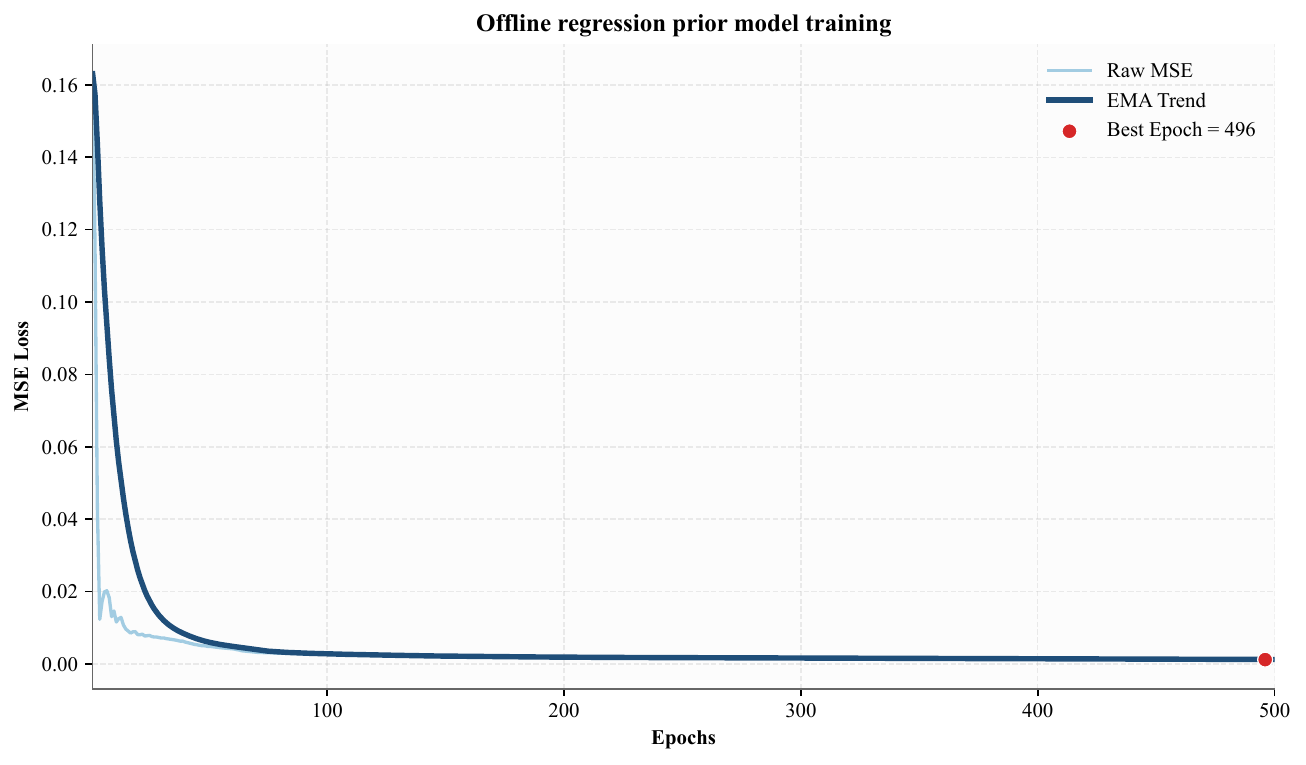}
        \caption{Offline regression prior model training curve}
        \label{fig:app_case_train_C}
    \end{subfigure}
    \hfill
    \begin{subfigure}[t]{0.49\linewidth}
        \centering
        \includegraphics[width=\linewidth]{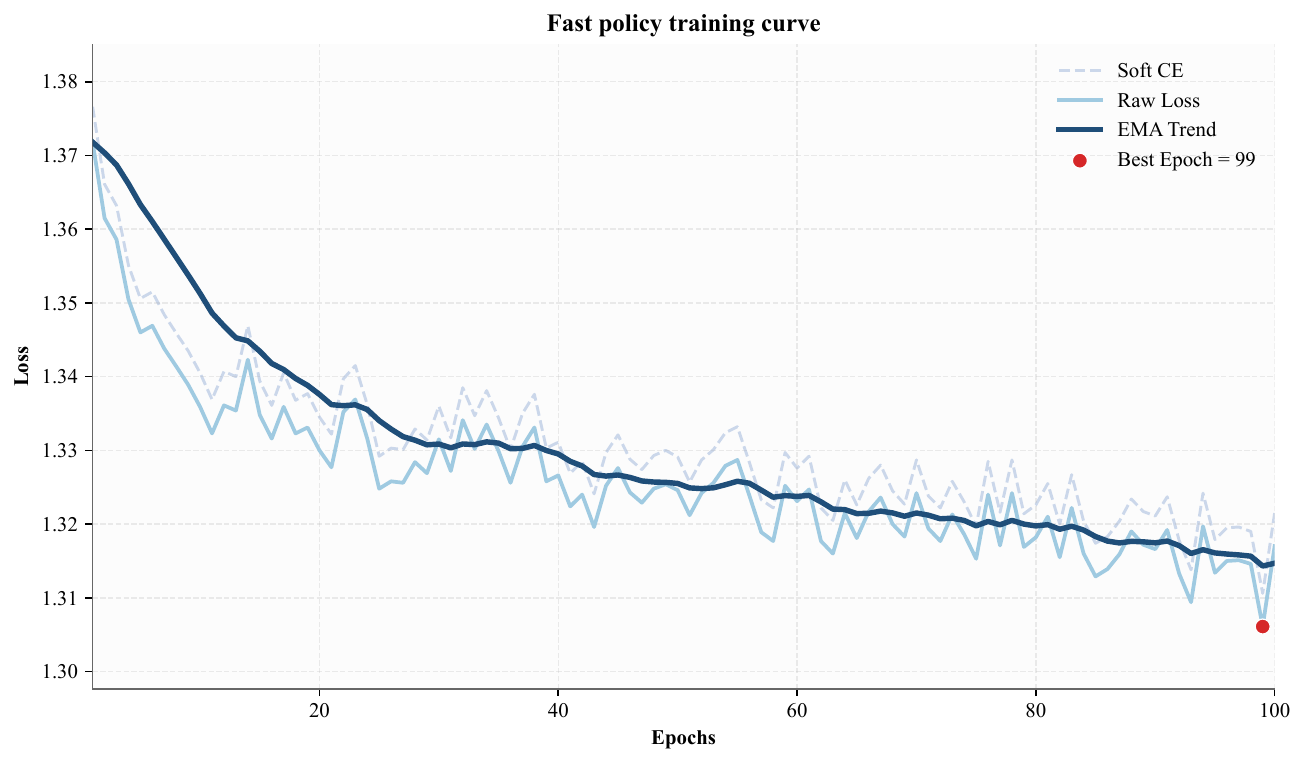}
        \caption{Fast policy training curve after UGW-based alignment}
        \label{fig:app_case_train_uwg}
    \end{subfigure}
\end{figure}

We first train an offline regression prior model to obtain an initialization prior for psycho-behavioral alignment. \autoref{fig:app_case_train_C} shows the training curve of the prior model. Based on this initialization, we further optimize the fast policy using UGW-based alignment to better link latent psychological changes to specific browsing behaviors, as shown in~\autoref{fig:app_case_train_uwg}.

\begin{figure}[ht]
    \centering
    \includegraphics[width=0.6\linewidth]{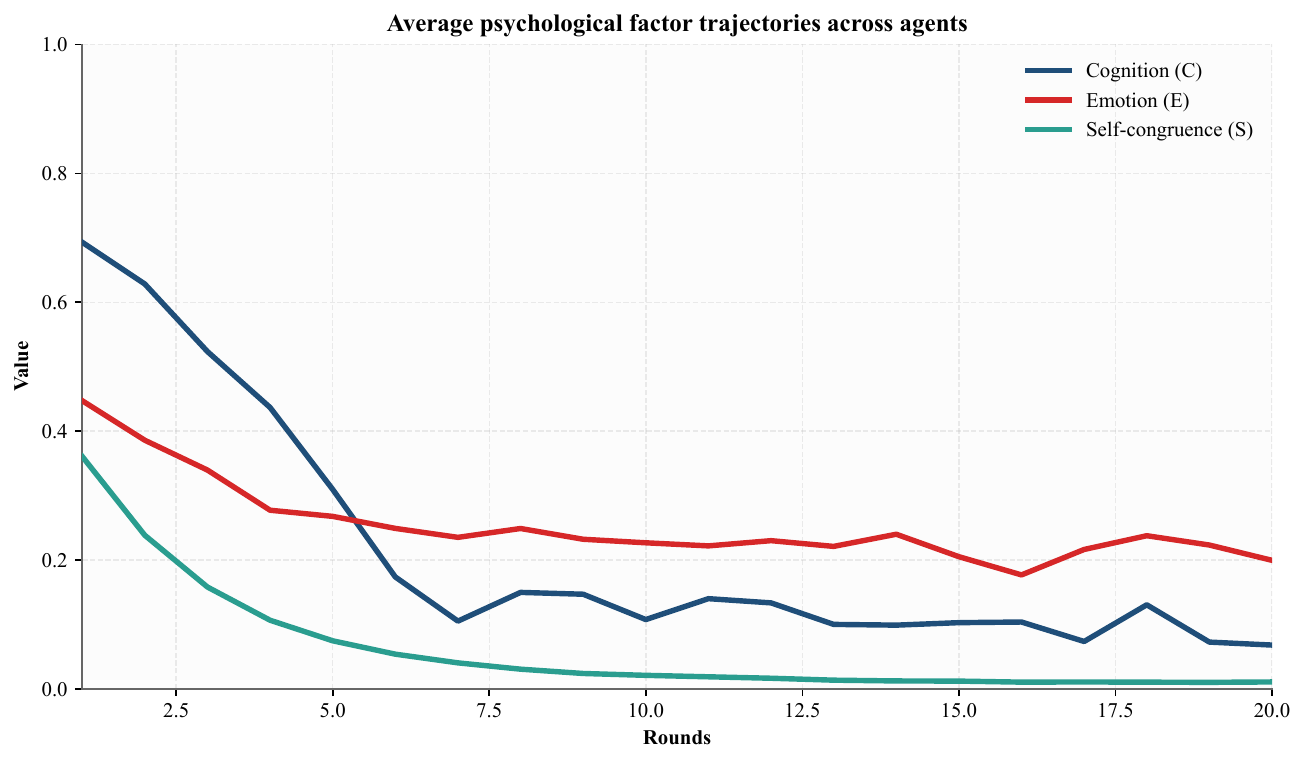}
    \caption{Average psychological factor trajectories across agents in the MovieLens scenario}
    \label{fig:app_case_train_psyfac}
\end{figure}

\autoref{fig:app_case_train_psyfac} shows the average psychological factor trajectories across browsing steps from a simulation of 10,000 users. The three variables $C$, $E$, and $S$ generally decrease as agents browse for longer horizons. Among them, self-congruence declines the fastest, which is consistent with relatively weaker profile--item alignment in this scenario.

\begin{figure}[ht]
    \centering
    \includegraphics[width=0.75\linewidth]{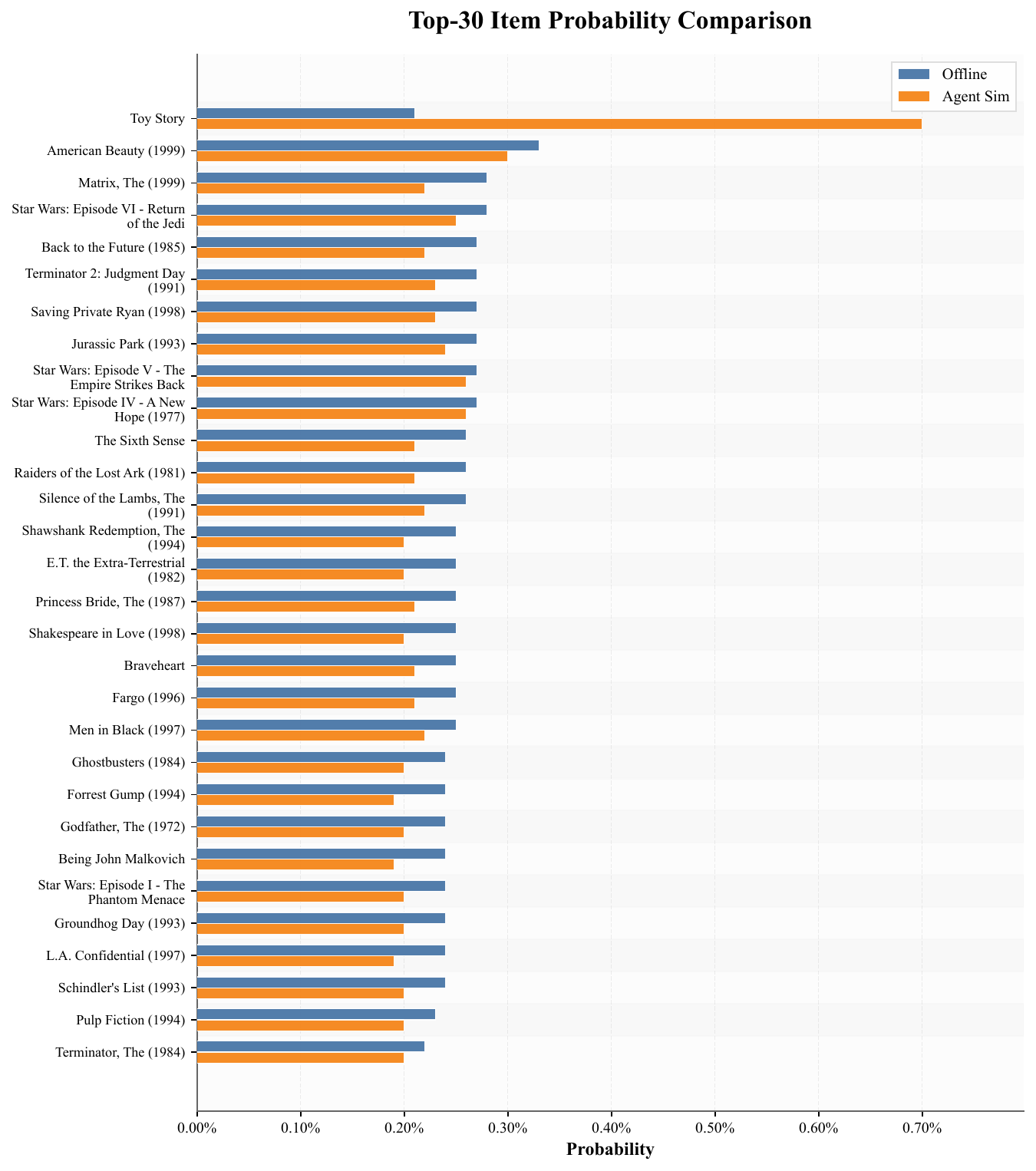}
    \caption{Top-30 item popularity comparison between the offline dataset and PsyBer-Agent selections in the MovieLens scenario}
    \label{fig:app_case_hotitem_1}
\end{figure}

\autoref{fig:app_case_hotitem_1} compares the item popularity distribution in the offline dataset with the popularity of items selected by PsyBer-Agent. We report the Top-30 items for comparison. Except for a higher-than-reference selection probability for \textit{Toy Story}, the remaining head-item probabilities broadly follow the offline popularity distribution.

\subsection{Amazon Magazine Subscriptions Shopping Scenario}
\begin{figure}[!ht]
    \centering
    \includegraphics[width=0.95\linewidth]{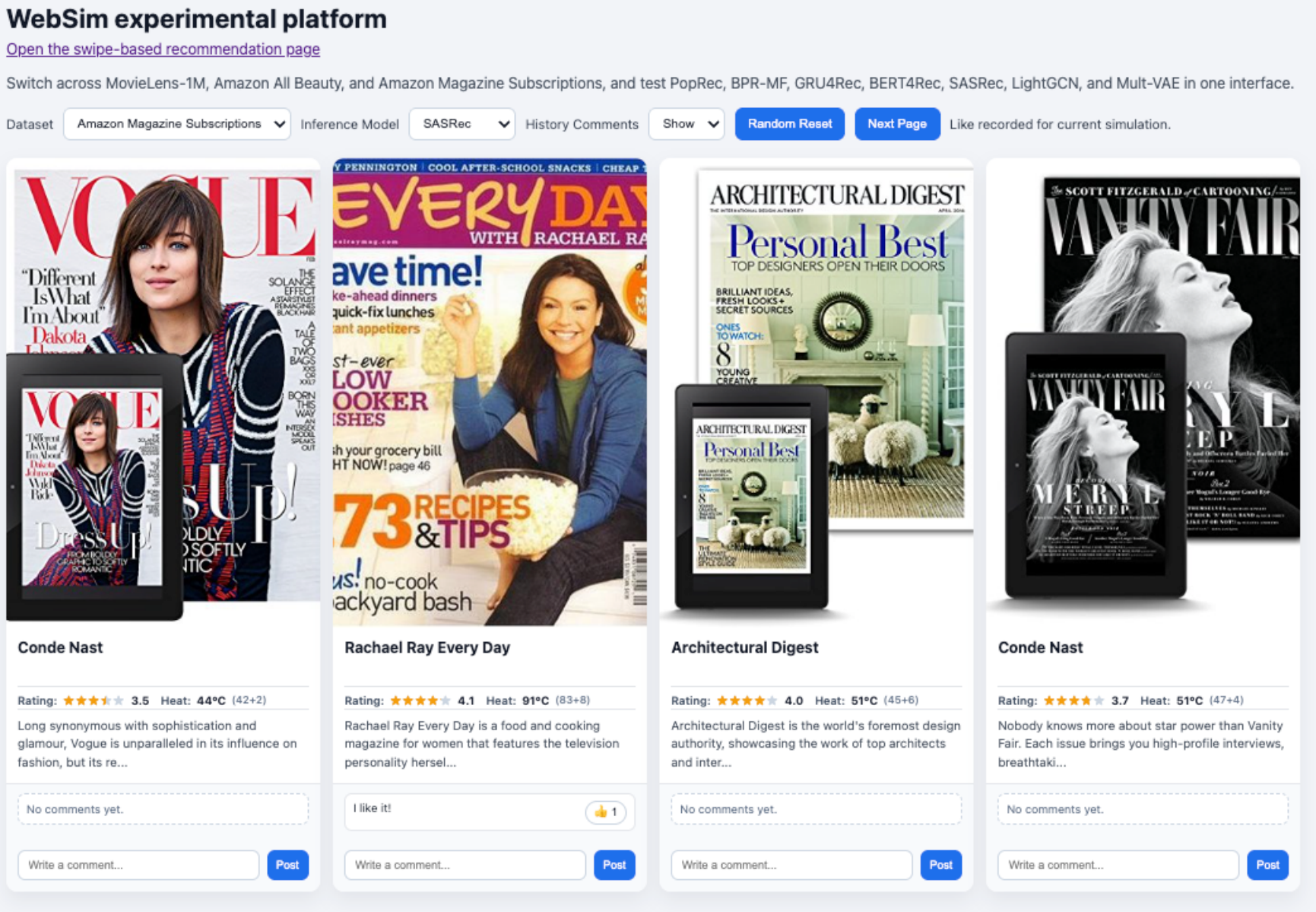}
    \caption{WebSim platform for the Amazon Magazine Subscriptions shopping interface. Each item consists of a cover, title, rating, and popularity score. The task can switch the dataset and the recommender under test, and the agent may navigate to the next page through psychology-informed decision making. Most Amazon Review subsets are supported in this mode.}
    \label{fig:app_ama_sim}
\end{figure}

We instantiate WebSim on the Amazon Magazine Subscriptions subset of the Amazon Review dataset as a shopping interface for agent evaluation. As shown in~\autoref{fig:app_ama_sim}, the default page displays recommended items in a one-row, four-column grid, while two drop-down menus allow switching between datasets and the recommender under test. In addition to item clicking, the interface supports random reset and next-page navigation. In this case study, we use GRU4Rec as the driving recommendation model.

\begin{figure}[!ht]
    \centering
    \begin{subfigure}[t]{0.49\linewidth}
        \centering
        \includegraphics[width=\linewidth]{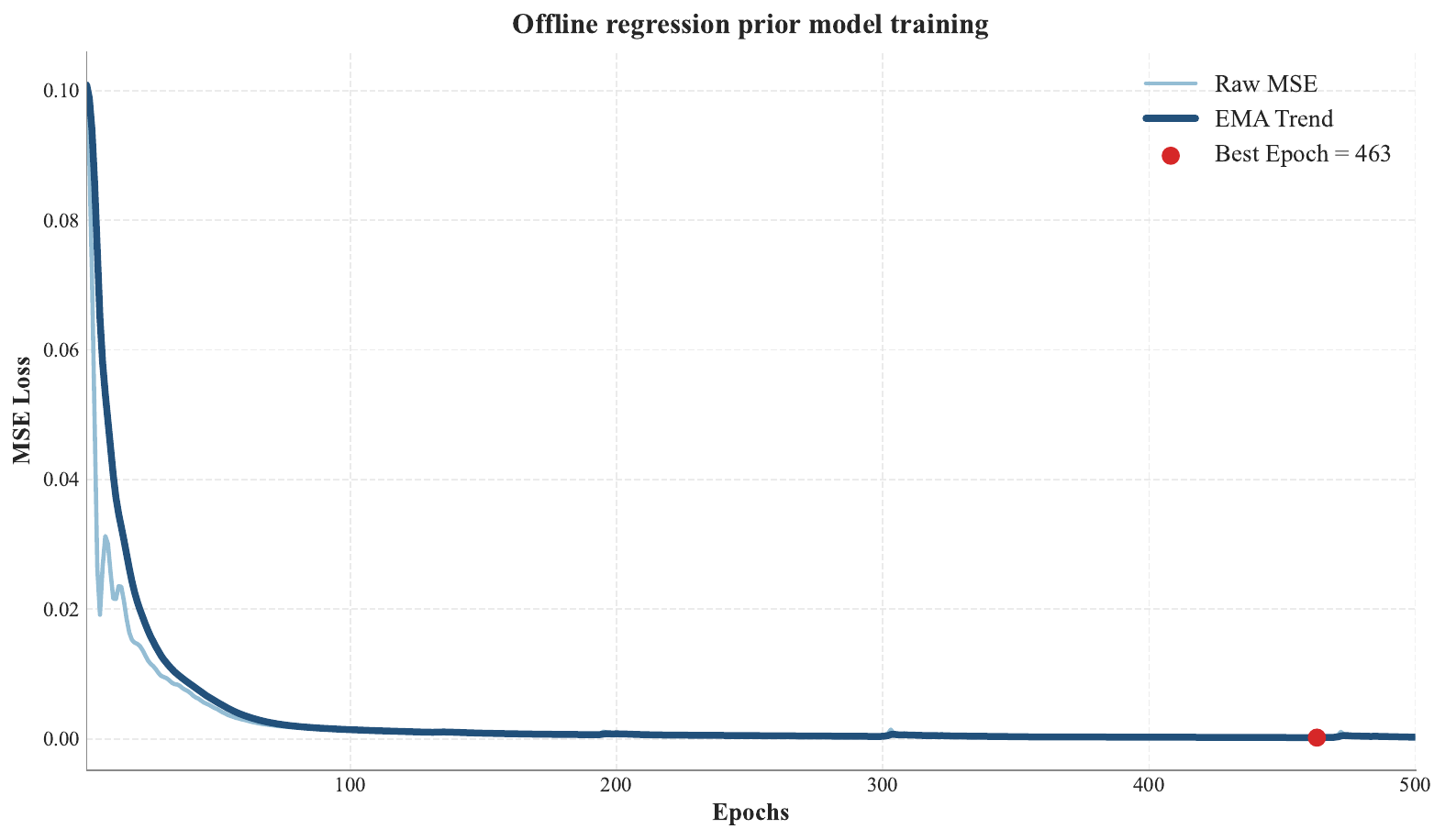}
        \caption{Offline regression prior model training curve}
        \label{fig:app_case_train_C_2}
    \end{subfigure}
    \hfill
    \begin{subfigure}[t]{0.49\linewidth}
        \centering
        \includegraphics[width=\linewidth]{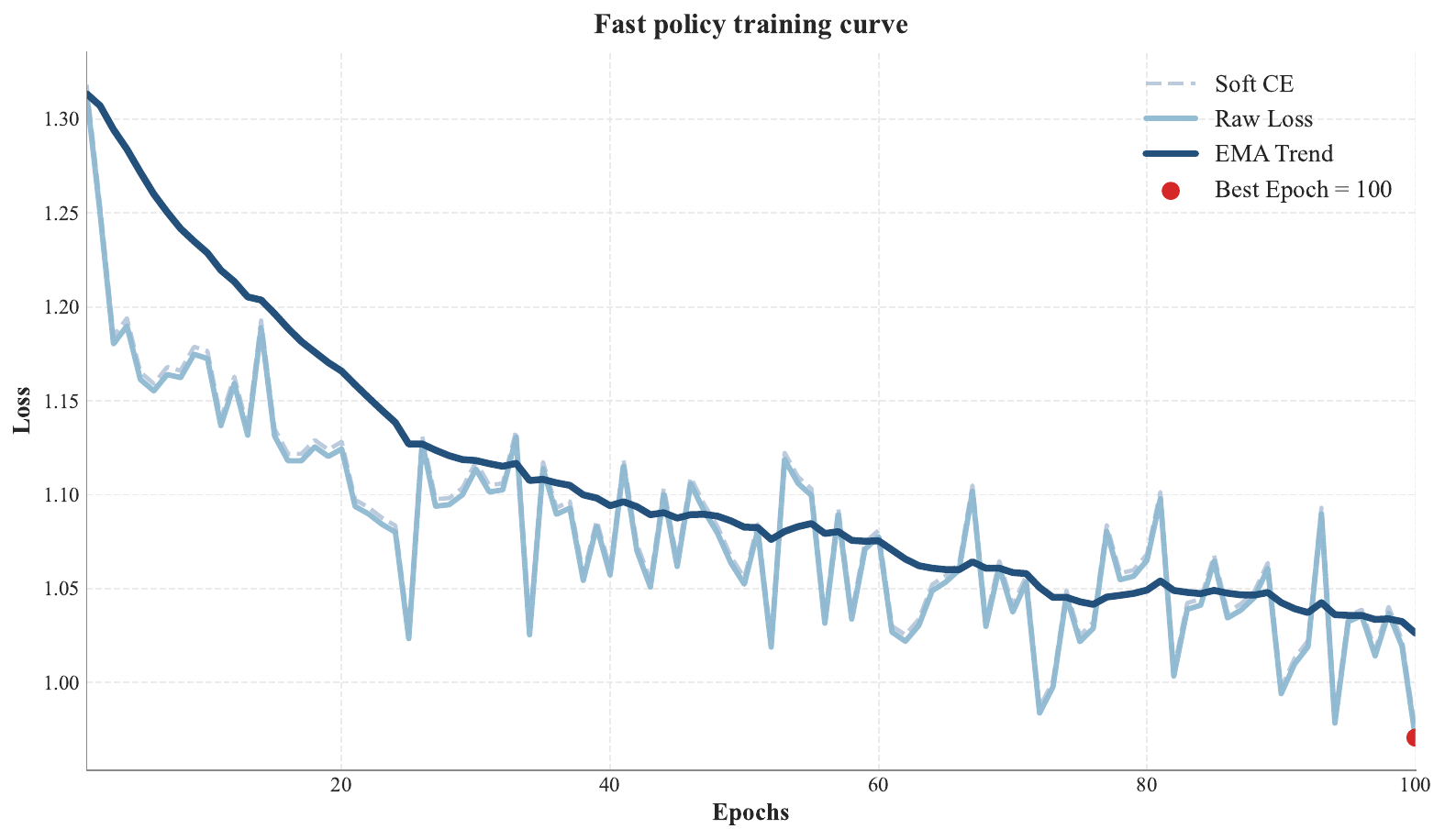}
        \caption{Fast policy training curve after UGW-based alignment}
        \label{fig:app_case_train_uwg_2}
    \end{subfigure}
\end{figure}

We first train an offline regression prior model on Amazon interaction data to obtain an initialization prior for psycho-behavioral alignment. \autoref{fig:app_case_train_C_2} shows the training curve of the prior model, where the loss decreases rapidly in early epochs and then converges stably. Based on this initialization, we further optimize the fast policy using UGW-based alignment to better link latent psychological changes to shopping behaviors such as clicks, skips, and pagination, as shown in~\autoref{fig:app_case_train_uwg_2}.

\begin{figure}[!ht]
    \centering
    \includegraphics[width=0.6\linewidth]{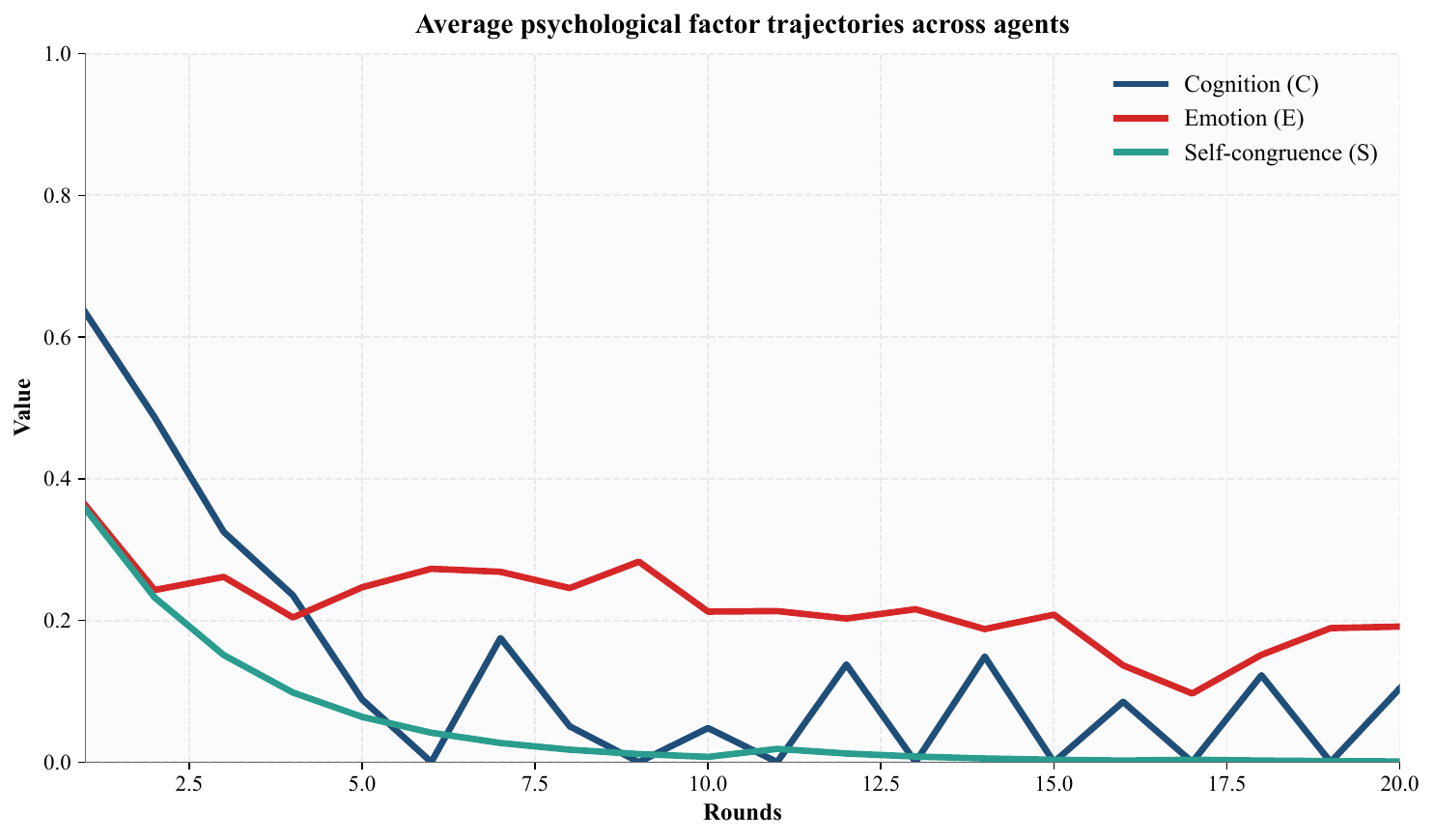}
    \caption{Average psychological factor trajectories across agents in the Amazon Magazine Subscriptions scenario}
    \label{fig:app_case_psy2}
\end{figure}

\autoref{fig:app_case_psy2} reports the average psychological factor trajectories in a 10,000-user simulation. We observe consistent temporal dynamics across browsing steps, with gradual attenuation under prolonged exposure and noticeable sensitivity when recommendation relevance drops. This trend suggests that the Amazon-based simulator preserves interpretable latent-state evolution rather than producing purely random shifts in behavior.

\begin{figure}[!ht]
    \centering
    \includegraphics[width=0.75\linewidth]{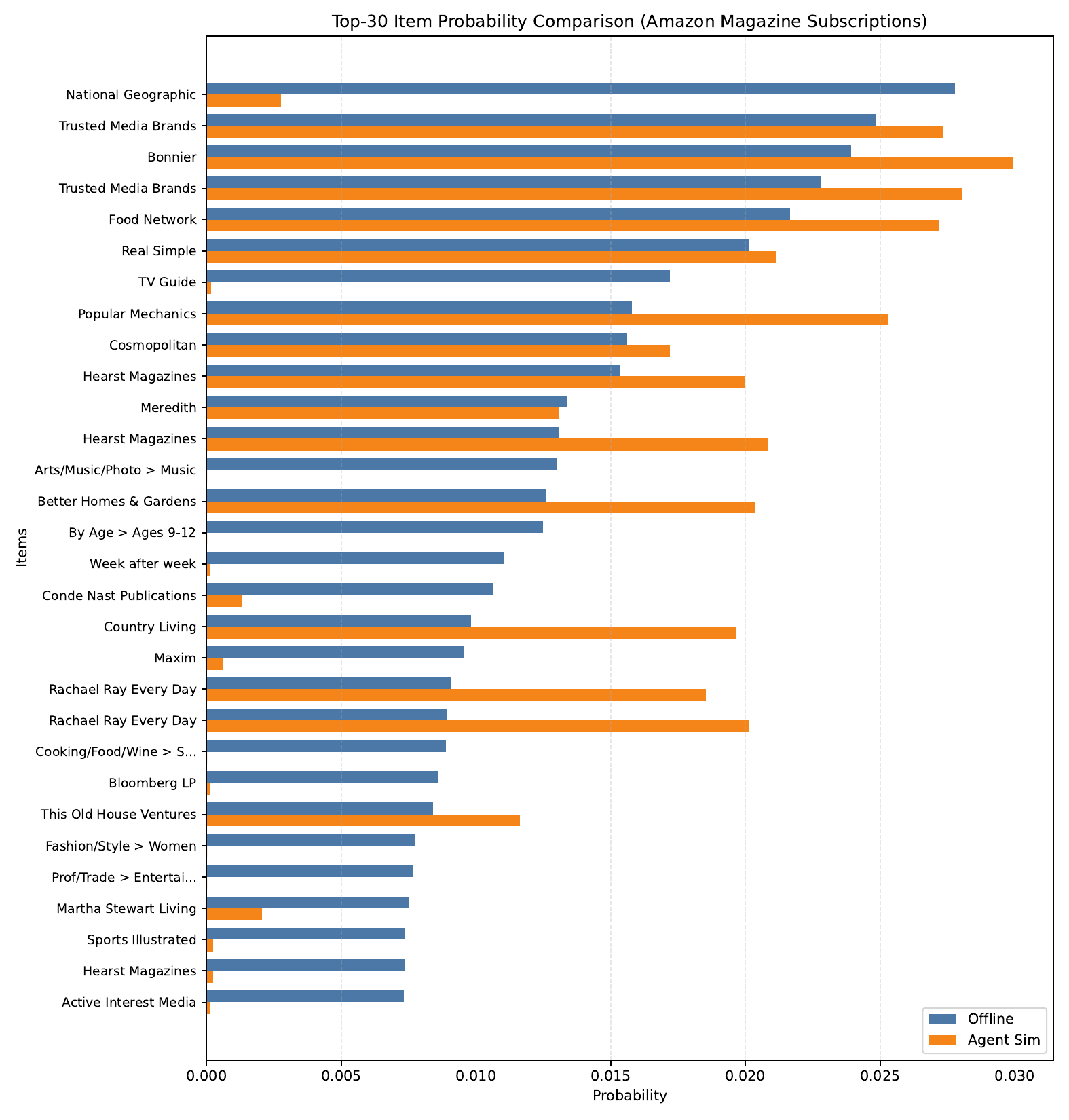}
    \caption{Top-30 item popularity comparison between the offline dataset and PsyBer-Agent selections in the Amazon Magazine Subscriptions scenario}
    \label{fig:app_case_hotitem_2}
\end{figure}

Finally, \autoref{fig:app_case_hotitem_2} compares item popularity in the offline dataset with the popularity of items selected by PsyBer-Agent. We report the Top-30 items for comparison. The two distributions are broadly aligned in the head, indicating that the simulator preserves the main demand pattern of real interactions while still showing selective deviations on several items due to psychology-driven preference dynamics.

\newpage


\section{Declarations}
\label{Declarations}

\subsection{Limitations}
\label{App:limitations}

PsyBer-Agent provides a controllable proxy for pre-deployment recommender evaluation, but it still has several limitations. First, its psychological state dynamics are designed around a compact three-dimensional representation of cognition, emotion, and self-congruence, which improves interpretability but may not capture all factors that shape real user behavior, such as social context, long-term habit formation, or domain-specific motivations. Second, the psycho-behavioral alignment module constructs teacher-forced pseudo-expert latent trajectories using the same Psy-Engine, because offline interaction logs do not contain ground-truth psychological annotations. As a result, these latent trajectories should be interpreted as log-conditioned reference states rather than human-validated psychological ground truth. If the inductive bias of Psy-Engine is misspecified, the resulting alignment signal may partially reinforce that misspecification, since the current framework does not directly anchor latent states to independently validated human psychological measurements. Third, although the simulator is aligned with offline interaction logs and evaluated across movie and e-commerce scenarios, its fidelity may degrade under substantially longer horizons, in unseen domains, or in recommendation interfaces whose interaction patterns differ from those in WebSim. Fourth, simulated users cannot fully replace real-user studies or online A/B testing, because LLM-based decisions and psycho-behavioral alignment may introduce systematic biases or overestimate deployment readiness. Finally, large-scale simulation depends partly on external LLM APIs and public interaction logs, so cost, latency, reproducibility across API versions, privacy protection, and fairness across user groups remain important practical considerations for future work.

\subsection{Broader impacts}
\label{App:broad}

This work supports safer and more cost-effective pre-deployment evaluation of recommender systems by using PsyBer-Agent and WebSim to simulate closed-loop user interactions before online release, helping identify behavioral mismatch, exposure concentration, robustness failures, and unintended group-level effects without immediately exposing real users to untested policies. At the same time, psychology-driven user simulation may be misused to optimize manipulative recommendation strategies, amplify engagement-driven feedback loops, or probe group responses to social cues in ways that increase polarization or unfair exposure; careless use of interaction logs may also raise privacy concerns. Therefore, PsyBer-Agent should be used as a diagnostic tool rather than the sole basis for deployment decisions, with responsible data handling, documented dataset provenance, monitoring for distributional bias and excessive herding, fairness checks across user groups, human oversight, and staged online validation in high-impact recommendation domains.

\stopcontents[app]
\end{appendices}
\end{document}